\documentclass[lettersize,journal]{IEEEtran}
\usepackage{amsmath}
\usepackage{wasysym}
\usepackage{footnote}
\usepackage{multirow}
\usepackage{pifont}
\usepackage{subfigure}
\usepackage[lined,ruled,commentsnumbered]{algorithm2e}
\usepackage{algpseudocode}
\usepackage{pdfpages}
\usepackage{subfigure}
\usepackage{tabularx}
\usepackage{booktabs}
\usepackage{amssymb}
\usepackage[font={small}]{caption}
\usepackage{makecell}
\usepackage{pifont}
\usepackage{float}
\usepackage{stfloats}

\begin{document}

\title{TrustGuard: GNN-based Robust and Explainable Trust Evaluation with Dynamicity Support}

\author{Jie Wang,
        Zheng Yan,~\IEEEmembership{Fellow,~IEEE,}
        Jiahe Lan,
        Elisa Bertino,~\IEEEmembership{Fellow,~IEEE,} \\
        and
        Witold Pedrycz,~\IEEEmembership{Life Fellow,~IEEE}

\thanks{Jie Wang, Zheng Yan (corresponding author), and Jiahe Lan are with the State Key Laboratory on Integrated Services Networks, School of Cyber Engineering, Xidian University, Xi’an 710071, China (e-mail: jwang1997@stu.xidian.edu.cn; zyan@xidian.edu.cn; jhlan16@stu.xidian.edu.cn).}
\thanks{Elisa Bertino is with the Department of Computer Science, Purdue University, West Lafayette, IN 47907 USA (e-mail:
bertino@purdue.edu).}
\thanks{Witold Pedrycz is with the Department of Electrical and Computer Engineering, University of Alberta, Edmonton, AB T6G 2R3, Canada, also with the Systems Research Institute, Polish Academy of Sciences, 00-901 Warsaw, Poland, also with the Department of Electrical and Computer Engineering, King Abdulaziz University, Jeddah 21589, Saudi Arabia, and also with the Department of Computer Engineering, Istinye University, 34010 Sariyer/Istanbul, Turkey (e-mail: wpedrycz@ualberta.ca).}
}

\markboth{Journal of \LaTeX\ Class Files,~Vol.~14, No.~8, August~2021}%
{Shell \MakeLowercase{\textit{et al.}}: A Sample Article Using IEEEtran.cls for IEEE Journals}


\maketitle

\begin{abstract}
Trust evaluation assesses trust relationships between entities and facilitates decision-making. Machine Learning (ML) shows great potential for trust evaluation owing to its learning capabilities. In recent years, Graph Neural Networks (GNNs), as a new ML paradigm, have demonstrated superiority in dealing with graph data. This has motivated researchers to explore their use in trust evaluation, as trust relationships among entities can be modeled as a graph. However, current trust evaluation methods that employ GNNs fail to fully satisfy the dynamic nature of trust, overlook the adverse effects of trust-related attacks, and cannot provide convincing explanations on evaluation results. To address these problems, we propose TrustGuard, a GNN-based accurate trust evaluation model that supports trust dynamicity, is robust against typical attacks, and provides explanations through visualization. Specifically, TrustGuard is designed with a layered architecture that contains a snapshot input layer, a spatial aggregation layer, a temporal aggregation layer, and a prediction layer. Among them, the spatial aggregation layer adopts a defense mechanism to robustly aggregate local trust, and the temporal aggregation layer applies an attention mechanism for effective learning of temporal patterns. Extensive experiments on two real-world datasets show that TrustGuard outperforms state-of-the-art GNN-based trust evaluation models with respect to trust prediction across single-timeslot and multi-timeslot, even in the presence of attacks. In addition, TrustGuard can explain its evaluation results by visualizing both spatial and temporal views.
\end{abstract}



\section{Introduction}
\IEEEPARstart{T}{rust} plays a crucial role in cyber security as it can effectively mitigate potential risks. Trust is characterized by such properties as subjectivity, dynamicity, context-awareness, asymmetry, and conditionally transferability~\cite{yan2008trust,yan2010autonomic}. Additionally, it is influenced by a range of factors, including context, as well as the subjective and objective properties of both the trustor and the trustee - two entities involved in a trust relationship. To quantify trust, trust evaluation has been investigated by taking the factors that affect trust into digital trust assessment and expression~\cite{wang2020survey}. Trust evaluation has been widely used in various fields to achieve diverse goals, such as intrusion detection~\cite{bao2012hierarchical}, access control~\cite{yan2015flexible}, and service management~\cite{chen2015trust,chen2014trust}. It serves as a basis to realize trustworthy networking by evaluating network node trust and network domain trust, which is crucial for constructing secure next-generation 6G communication networks~\cite{ylianttila20206g}. In the Bitcoin-OTC market~\cite{snapnets}, which is an Over-The-Counter (OTC) market for Bitcoin currency, trust evaluation has emerged as a means to reduce the probability of being defrauded due to the anonymous nature of Bitcoin users and the inherent counterparty risks associated with transactions. All in all, trust evaluation has become one of the essential components to identify internal and external attackers, facilitate decision-making, and enhance system security.

Several trust evaluation models have been proposed, which can be generally classified into statistical models, reasoning models, and Machine Learning (ML) models~\cite{wang2022survey}. Statistical models use simple computations to reflect the impacts or weights of different pieces of trust evidence. This type of models is intuitive and can involve various trust evidence into evaluation. However, it is particularly sensitive to the weights, which may vary for different applications. Reasoning models first characterize trust with respect to multiple dimensions (e.g., belief, disbelief and uncertainty) and then infer trust via pre-defined rules (e.g., Subjective Logic~\cite{josang2016subjective} and Bayes theorem~\cite{josang2002beta}). Although they can well represent real-world situations, these models rely on prior knowledge and suffer from improper configuration of evaluation rules. Notably, both statistical models and reasoning models heavily rely on previous knowledge about the application domain and some system parameter settings, hindering the generality of trust evaluation. ML models typically evaluate trust through supervised learning classification~\cite{wang2020survey}, which is often referred to as trust prediction\footnote{Herein, we use ``trust evaluation" and ``trust prediction" interchangeably, as GNN is one type of ML.}. By analyzing the patterns and trends in trust-related data, such models can predict trust. They can learn weights and rules automatically even in different contexts, as well as achieve high accuracy when sufficient data are available~\cite{liu2019neuralwalk}. As a result, the ML models hold extraordinary advantages for achieving intelligent and precise trust evaluation.

Graph Neural Networks (GNNs) are a relatively new type of ML models, which have shown superiority over a variety of network analysis tasks, including node classification, link prediction, and graph classification~\cite{WTAGRAPH,stgcn}. The strength of GNNs lies in their strong capability to learn meaningful representations (i.e., embeddings) of nodes/edges through information propagation and aggregation guided by a graph structure~\cite{huo2023trustgnn}. This makes GNNs particularly suitable for predicting trust relationships, since trust relationships among entities in different networks (e.g., social networks and Internet of Things (IoT)) can be naturally modeled as graphs, with nodes representing entities and edges between nodes representing their trust relationships~\cite{KGTrust}. In addition, unlike traditional ML models that often require feature engineering, GNNs offer an end-to-end trust evaluation, where raw graph data can be directly fed into GNN models to obtain an evaluation result. The simplicity and effectiveness of GNNs have motivated researchers to explore their potential for trust evaluation.

Despite the attractive advantages of GNNs, research on GNN-based trust evaluation is still in its infancy. Existing models~\cite{huo2023trustgnn,lin2020guardian,jiang2022gatrust,lin2021medley} suffer from several shortcomings. First, dynamicity, the inherent nature of trust, is not well considered in those models~\cite{huo2023trustgnn,lin2020guardian,jiang2022gatrust}. Trust relationships always evolve over time or change with events. This implies that the trust relationships of a same trustor-trustee pair at different timeslots are different. By considering temporal information, it is possible to make reasonable predictions on future trust relationships from their previous patterns. Conversely, ignoring temporal information about trust leads to questionable inference~\cite{lin2021medley,Xu2020Inductive}, which restricts the effectiveness of trust evaluation. Second, trust-related attacks significantly undermine the trust evaluation process, resulting in meaningless evaluation results. Specifically, malicious nodes can manipulate ratings to destroy the trustworthiness of a well-behaved node or boost the trustworthiness of a malicious node~\cite{wang2019graphsecurity}, which are known as bad-mouthing and good-mouthing attacks, respectively. These types of attacks are prevalent in the real world~\cite{Liu2017efficiently}. For instance, Yelp has identified around 16\% of its restaurant ratings as dishonest~\cite{luca2016fake}. Such dishonest ratings can mislead an evaluation model, but none of the existing models make efforts to enhance their robustness. Third, trust evaluation results derived from GNN models are not human-friendly due to the lack of explanations. Specifically, it is unclear whether the results are reasonable and user-acceptable. The lack of explanations reduces the trustworthiness of the evaluation model from a human perspective and further hinders its adoption~\cite{wang2022survey,Anomaly}. Therefore, it is critical to investigate an accurate, robust, and explainable GNN-based trust evaluation model that fully incorporates intrinsic trust properties, offers accurate evaluation even in a malicious environment, and provides convincing explanations on evaluation results.

Unfortunately, achieving those goals requires addressing several challenges. First, it is difficult to automatically capture temporal patterns of trust and encode them into node representations. Time is a continuous variable with infinite values, making it difficult to accurately represent and process timing information. Additionally, trust interactions at different timeslots should be assigned different weights in trust evaluation. For example, recent interactions may have higher weights than historical ones. But how to automatically model the temporal patterns has not yet been explored in the context of trust evaluation. Second, it is not easy to differentiate between benign and malicious interactions, and assign low weights to malicious ones during trust propagation and aggregation. The reason is the lack of sufficient information about nodes apart from their trust relationships to determine their benign or malicious nature. Third, it is challenging to explicitly present the importance of different trust interactions formed over time in order to explain a model output, i.e., a trust evaluation result. This is because the spatial and temporal importance learned by ML models are always implicit. For example, recurrent methods~\cite{pareja2020evolvegcn,Euler}, extensively employed in dynamic graph modeling for learning temporal importance, are difficult for humans to understand due to the intricate structure of Recurrent Neural Networks (RNNs).

Although a lot of dynamic graph models~\cite{pareja2020evolvegcn,seo2018structured,sankar2020dysat} have been proposed to capture both spatial and temporal patterns, known as spatial-temporal GNN models, they cannot be directly applied to trust evaluation. On one hand, most of them focus on node-level tasks and link prediction, whereas we aim to tackle an edge classification problem, thus requiring a great focus on edge weights, i.e., trust levels. On the other hand, existing models do not incorporate fundamental trust properties such as asymmetry and conditional transferability, potentially resulting in inaccurate trust evaluation. Therefore, despite the success of existing spatial-temporal GNN models in various fields, how to apply them into trust evaluation requires further extensive investigation.

To this end, this paper proposes TrustGuard, the first trust evaluation model based on GNN that addresses the above challenges. TrustGuard meets basic trust properties, defends against typical trust-related attacks, and provides convincing explanations on evaluation results. It consists of four layers, namely a snapshot input layer, a spatial aggregation layer, a temporal aggregation layer, and a prediction layer. In the snapshot input layer, a dynamic graph is segmented into several ordered snapshots based on regular intervals. For each snapshot at a timeslot, the spatial aggregation layer employs a robust aggregator to defend against attacks (e.g., bad/good-mouthing attacks) and learns spatial node embeddings that contain local structural information (i.e., trust relationships). These embeddings are treated as inputs to the temporal aggregation layer, which is responsible for learning temporal patterns from a sequence of snapshots based on a position-aware attention mechanism. Lastly, final node embeddings, which incorporate both structural and temporal information, are fed to the prediction layer for trust relationship prediction between any two nodes in the next timeslot (i.e., single-timeslot prediction) or in the multiple future timeslots (i.e., multi-timeslot prediction). We explain trust evaluation results obtained from TrustGuard by visualizing both spatial and temporal views. A user study is further conducted to validate user acceptance on such an explanation. To summarize, the main contributions of this paper include:
\begin{itemize}
    \item We propose TrustGuard, the first GNN model for trust evaluation, which supports the dynamic nature of trust, performs robustly against typical trust-related attacks, and offers explanations.
    
    \item We design a defense mechanism in the spatial aggregation layer of TrustGuard to counter trust-related attacks, which, to the best of our knowledge, is the first study on robustness in GNN-based trust evaluation.
    
    \item We adopt a position-aware attention mechanism in the temporal aggregation layer, which enables TrustGuard to effectively capture temporal patterns to improve both evaluation accuracy and robustness.
    
    \item We conduct extensive experiments based on two real-world dynamic datasets to demonstrate the superiority of TrustGuard over state-of-the-art methods in terms of trust prediction across single-timeslot and multi-timeslot, regardless of the presence of attacks. The scalability of TrustGuard is also validated through tests on two large-scale datasets.
\end{itemize}

The remainder of this paper is organized as follows. In Section~\ref{section2}, we briefly review existing GNN-based trust evaluation models and compare them with TrustGuard in a qualitative way. Section~\ref{section3} defines the problem we aim to solve, specifies our security model, and introduces preliminary knowledge. We describe the design of TrustGuard in detail in Section~\ref{section4}, followed by extensive experiments and performance analysis in Section~\ref{section5}. The strengths and limitations of TrustGuard are discussed in Section~\ref{section6}. Finally, we draw a conclusion in the last section.

\newcommand{\minitab}[2][l]{\begin{tabular}{#1}#2\end{tabular}}

\begin{table*}[]
\footnotesize
\centering
\caption{Comparison of TrustGuard with existing models.}
\label{related_work}
{\vspace{-2mm}\CIRCLE: satisfy a criterion; \Circle: do not satisfy a criterion; "-": not available.}
\\[2mm]

\begin{tabular}{c|c|c|c|cc|cc}
\toprule[1.5pt]
\multirow{2.5}{*}{Model} & \multirow{2.5}{*}{Asymmetry} & \multirow{2.5}{*}{Dynamicity} & \multirow{2.5}{*}{\minitab[c]{No need fine-\\grained timestamps}} & \multicolumn{2}{c|}{Robustness} & \multicolumn{2}{c}{Explainability} \\
\cmidrule{5-8}
                       &               &             &                             & Bad-mouthing           & Good-mouthing           & \makebox[0.08\textwidth][c]{Spatial}         & \makebox[0.08\textwidth][c]{Temporal}   \\ \midrule
Guardian \cite{lin2020guardian} & \CIRCLE   & \Circle & -  & \Circle   & \Circle  & \Circle  & -\\
GATrust \cite{jiang2022gatrust} & \CIRCLE   & \Circle & -  & \Circle   & \Circle  & \Circle  & -\\
TrustGNN \cite{huo2023trustgnn} & \CIRCLE   & \Circle & -  & \Circle   & \Circle  & \CIRCLE  & -\\
Medley \cite{lin2021medley} & \Circle   & \CIRCLE & \Circle  & \Circle   & \Circle  & \Circle  & \Circle\\

\midrule

TrustGuard & \CIRCLE   & \CIRCLE   & \CIRCLE & \CIRCLE  & \CIRCLE  & \CIRCLE  & \CIRCLE\\

\bottomrule[1.5pt]
\end{tabular}

\vspace{-2mm}
\end{table*}

\section{Related Work} \label{section2}
In this section, we review four existing methods that apply GNNs to trust evaluation. We also compare them with TrustGuard in terms of asymmetry, dynamicity, fine-grained timestamp requirement, robustness against attacks (e.g., bad/good-mouthing attacks), and explainability, as shown in Table~\ref{related_work}.

Guardian~\cite{lin2020guardian} is the first work that applies GNNs to trust evaluation. It divides neighboring nodes into in-degree neighbors and out-degree neighbors to model the roles of trustee and trustor, respectively. Then it utilizes a Graph Convolution Network (GCN)~\cite{kipf2017semisupervised} to aggregate information from these nodes and their corresponding trust interactions. Specifically, Guardian utilizes a mean aggregator to aggregate information from neighbors at different hops to achieve trust propagation. It outperforms traditional trust evaluation models and neural network-based models~\cite{liu2019neuralwalk} in terms of accuracy and efficiency. However, it fails to capture the dynamic nature of trust. Moreover, the mean aggregator cannot differentiate the importance of different neighbors. Intuitively, different nodes should be assigned different weights when evaluating the trustworthiness of a target node, since they hold different trust values and have different numbers of interactions with the target node. Such issues lead to inaccurate trust evaluation. We also note that neither robustness nor explainability is considered in Guardian.

GATrust~\cite{jiang2022gatrust} was designed to address the problem of Guardian that does not consider node features (e.g., personal hobbies), which are also crucial in trust evaluation. Like Guardian, GATrust establishes trust asymmetry using the concepts of out-degree and in-degree, and propagates trust via the message-passing mechanism of GNN. GATrust integrates multi-facet properties of nodes, including node contextual features, network structural information, and trust relationships. It leverages the key concept of Graph Attention Network (GAT)~\cite{gat2018} to assign varying weights to each node's different properties. By fusing and aggregating these properties, node embeddings learned by GATrust contain rich information, facilitating accurate prediction of trust relationships. However, like Guardian, GATrust also ignores the evolving nature of trust, which hinders its efficacy and generality. Robustness and explainability are not explored, either.  

Differently, TrustGNN~\cite{huo2023trustgnn} achieves trust asymmetry and propagation by constructing different types of trust chains based on a RotatE-like approach~\cite{sun2018rotate}. These trust chains are aggregated to form comprehensive representations of nodes, i.e., embeddings, for trust relationship prediction. TrustGNN employs learnable attention scores to distinguish the contributions of different trust chains. The visualization of attention scores with respect to each type of trust chain offers spatial explainability. However, TrustGNN focuses on a specific snapshot, omitting the fact that trust relationships dynamically change. As such, TrustGNN lacks temporal explainability. Trust-related attacks are unfortunately ignored. 

Those three models consistently neglect the dynamicity of trust, which results in low efficacy and questionable predictions. To be specific, GNN models based on static analysis may mistakenly utilize later interactions to predict previous trust relationships in model construction due to the lack of temporal information in training data~\cite{Xu2020Inductive, Euler}. 

To tackle the issue caused by static trust evaluation, Medley~\cite{lin2021medley} encodes temporal features along with other essential features for discovering hidden and time-aware trust relationships. Specifically, it employs a functional time encoding module~\cite{Xu2020Inductive} to capture temporal features and assigns different weights to timestamped trust interactions with the help of an attention mechanism. With the incorporation of temporal information, Medley shows a significant improvement compared with Guardian. Nonetheless, such a design requires fine-grained timestamps, which may be hard to acquire in practice~\cite{sankar2020dysat}. Also, it incurs high storage consumption and high computational overhead when nodes and edges are frequently updated~\cite{zhu2022learnable}. Moreover, Medley only covers the role of trustors when learning node representations, leading to a lack of effective information about trustees. The rationality of evaluation results derived from Medley is not well clarified, causing poor explainability. Model robustness is also missing in this work.

\textbf{\emph{Discussion.}} From Table~\ref{related_work}, we obtain three observations. First, none of the existing GNN-based trust evaluation models consider attack resiliency. They all assume that nodes in a graph are honest and trust ratings provided by them are fair. However, this is not realistic. 
Second, existing models lack full support of basic trust properties, particularly, the dynamicity of trust. Although Medley supports dynamicity, its generality is limited due to the reliance on fine-grained timestamps. Asymmetry, another intrinsic property of trust, is unfortunately missed in Medley.
Third, explainability is seldom addressed in these models. GNN models look like black boxes to human beings, which probably affects the trustworthiness of the model and further hinders its practical deployment. Therefore, a robust and explainable GNN-based trust evaluation model with full consideration of the intrinsic trust properties should be explored to achieve high evaluation accuracy in a malicious environment with an easy understanding by users.

\section{Problem Statement} \label{section3}
In this section, we first formulate the problem we aim to solve. Then, we introduce the security model of TrustGuard, followed by preliminaries about GNNs and a self-attention mechanism. Table~\ref{notations} lists the notations used in this paper.

\subsection{Problem Definition}
We consider the problem of trust evaluation in a dynamic graph $\mathcal{G}(T)$ observed within the time interval $[0,T]$. $\mathcal{G}(T) = (\mathcal{V}(T),\mathcal{E}(T))$, where $\mathcal{V}(T)$ is a time-varying set of nodes and $\mathcal{E}(T)$ is a time-varying set of edges. 
$\mathcal{G}(T)$ can be split into a series of ordered snapshots based on a specific strategy, e.g., time-driven strategy or event-driven strategy. Applying the time-driven strategy, the graph is segmented into several snapshots based on a fixed time interval. This implies that all events (i.e., interactions) that occur within an interval are analyzed as a whole. Conversely, if adopting the event-driven strategy, the graph is segmented based on a predefined number of events, with each snapshot containing the same number of events. This results in varying time intervals for each snapshot, depending on the occurring rate of events. Both strategies have their pros and cons, and the choice between them depends on the specific requirements of the application.

In this paper, we adopt the time-driven strategy since it is more suitable for GNN models to capture temporal patterns from a series of ordered snapshots than the event-driven strategy. A detailed discussion and experimental comparison of the two strategies is presented in Appendix~\ref{appdendix_events}. As such, $\mathcal{G}(T) \to \{\mathcal{G}^{t_1},\mathcal{G}^{t_2},\cdots,\mathcal{G}^{t_n}\}$, where each snapshot $\mathcal{G}^{t_i}$ is a weighted directed graph observed at timeslot $t_i$, with a node set $\mathcal{V}^{t_i}$ and an edge set $\mathcal{E}^{t_i}$. Note that a timeslot refers to a specific time interval, during which all interactions among nodes are collected to form a snapshot. For $\mathcal{G}^{t_i}=(\mathcal{V}^{t_i},\mathcal{E}^{t_i})$, any node $u,v \in \mathcal{V}^{t_i}$, and $e^{w,t_i}_{u \to v} \in \mathcal{E}^{t_i}$ denotes that $u$ trusts $v$ with level $w$ at timeslot $t_i$. $w$ belongs to $\mathcal{W}$, which denotes the set of trust levels and varies across different application scenarios. For example, $\mathcal{W}=\{Trust, Distrust\}$ in Bitcoin-OTC and Bitcoin-Alpha~\cite{snapnets}, while $\mathcal{W}=\{Observer, Apprentice, Journeyer, Master\}$ in Advogato~\cite{massa2009bowling}. We define $\left | \mathcal{W} \right |$ as the number of trust levels.

With the aforementioned notations and definitions, we can formally define the trust evaluation problem as below. Given a dynamic graph $\mathcal{G}(T)$ that may contain malicious interactions, our goal is to learn/train an accurate and robust trust evaluation model $f(\cdot)$ using $\mathcal{G}(T)$, such that it can evaluate/predict the trust level $w$ between any two nodes $u,v$ at time $T+ \Delta T$, i.e., $f(e_{u \to v}(T+ \Delta T))=w$. $\Delta T$ can be a duration of one timeslot or multiple timeslots.

\begin{table}[]
\footnotesize
\centering
\caption{Summary of notations.}
\label{notations}
\begin{tabular}{c|c}
\toprule[1.5pt]
Notation       & Explanation \\ 
\midrule
$\mathcal{G}^{t_i}$ & the snapshot observed at timeslot $t_i$    \\
$\mathcal{V}^{t_i}$, $\mathcal{E}^{t_i}$ & the sets of nodes and edges of the snapshot $\mathcal{G}^{t_i}$    \\
$\mathcal{W}$ & the set of trust levels     \\
$\left | \mathcal{W} \right |$ & the number of trust levels     \\
$w_{u \gets v}$   & the trust level of $v$ towards $u$     \\
$\omega_{u \gets v}$   & the trust level embedding     \\
$d_e$ & the dimension of node embedding     \\
$Msg_{u \gets v}$ & the message from $v$ to $u$ containing trust information    \\
$N_{u}(te)$, $N_{u}(tr)$ & the neighbor sets of node $u$ as a trustee and a trustor    \\
$s_{u, v}$ & the cosine similarity between $u$ and $v$    \\
$r_{u \gets v}$ & the robust coefficient of the edge directed from $v$ to $u$    \\
$h_{u}(te)$, $h_{u}(tr)$ & the embeddings of node $u$ as a trustee and a trustor     \\
$h_u^{t_i}$ & the embedding of node $u$ at timeslot $t_i$    \\
$\alpha^{\langle t_i,t_n \rangle}_u$ & the attention score of timeslot $t_i$ to timeslot $t_n$    \\
$h_u^{t_{sum}}$ & the final embedding of node $u$    \\
$p_{u \gets v}$ & the probabilistic prediction vector of the trust level \\
& from $v$ towards $u$    \\
$\otimes$ & the concatenation operator    \\
$\sigma$ & the non-linear activation function    \\
$W$, $b$ & the trainable parameters in TrustGuard    \\
\bottomrule[1.5pt]
\end{tabular}
\vspace{-2mm}
\end{table}

\subsection{Security Model}
We focus on typical trust-related attacks, namely bad-mouthing, good-mouthing, and on-off attacks (i.e., conflict behavior attacks)~\cite{wang2020survey,chen2015trust,chen2014trust}, emphasizing the harmful behavior of nodes from a local view. These attacks are different from adversarial attacks, which exploit the inherent vulnerabilities of GNN models and emphasize the adversarial relationship between the adversary and the model from a global view. Specifically, within a graph, a malicious node can destroy the trustworthiness of a well-behaved node by providing bad ratings to launch a bad-mouthing attack. On the contrary, a malicious node can boost the trustworthiness of a node by providing good ratings to raise a good-mouthing attack (i.e., ballot-stuffing attack). These two attacks are the basic manifestations of various attacks in the graph since malicious nodes exhibit their malicious behaviors by adding positive/negative edges. For the on-off attack, a malicious node behaves well and badly alternatively to avoid detection. Note that on-off attackers occasionally perform badly by launching bad/good-mouthing attacks. Its difference from the bad/good-mouthing attacks is that on-off attacks take into account time factor, while bad/good-mouthing attacks do not require this consideration. Both bad-mouthing and good-mouthing attacks can be launched in a collaborative way by a number of attackers to form collusion attacks~\cite{chen2014trust}, i.e., collaborative bad/good-mouthing attacks. They are more disruptive than individual ones and thus are adopted for testing the robustness of TrustGuard in Section~\ref{rq2}. The presence of malicious nodes in the graph would generate a number of dishonest ratings, which mislead the training process of a trust evaluation model and further degrade its performance~\cite{jagielski2021subpopulation}. In this paper, we assume that most of the nodes in the graph are honest. This is reasonable as nodes are unlikely to join a system with predominantly malicious behaviors. Meanwhile, it is hard for attackers to monopolize most of system resources to carry out attacks.

\subsection{Preliminaries}
\subsubsection{Graph Neural Network} GNN is a type of neural network specifically designed to process graph-structural data. 
It can learn meaningful node representations, i.e., embeddings, which capture both topology and attribute information of nodes within their $L$-hop neighborhoods by iterating local information propagation and aggregation for $L$ times in a graph~\cite{huo2023trustgnn}. In this way, nodes that are similar in the graph are embedded close to each other in an embedding space. These embeddings can be then used for various downstream tasks, such as node classification and link prediction. To be specific, a GNN model contains three key functions: a message-passing function $MSG$, an aggregation function $AGG$, and an update function $UPD$ \cite{ying2019gnnexplainer}. Consider a node $u$ and its neighbor $v$, where $v$ belongs to $N_u$, the set of immediate neighbors of $u$. The definition of $N_u$ varies depending on whether the graph is directed or undirected. Let $h^{l-1}_u$ be the embedding of node $u$ in the $(l-1)$-th layer of a GNN model. In the $l$-th layer, firstly, $MSG$ specifies what messages need to be propagated from $v$ to $u$. The message is constructed as $m_{u,v}^l=MSG(h^{l-1}_u,h^{l-1}_v,r_{u,v})$, where $r_{u,v}$ refers to the relation between the two nodes. Secondly, $AGG$ aggregates all messages from $u$'s neighbors and obtains an aggregated message $\bar{m}^l_u=AGG(\{m^l_{u,v}, \forall v \in N_{u} \})$. Lastly, $UPD$ combines the aggregated message and $u$'s embedding $h^{l-1}_u$ from the previous layer to generate $u$'s embedding in the $l$-th layer, i.e., $h^l_u=UPD(\bar{m}^l_u,h^{l-1}_u)$.

\subsubsection{Self-attention Mechanism} The self-attention mechanism~\cite{vaswani2017attention} is widely used in natural language processing to enable a model to selectively focus on different parts of an input sequence while making predictions. Specifically, the input is firstly transformed into three vectors, namely key, query, and value. Among them, the query and the key vectors are then used to calculate an attention score for each element in the input sequence. Finally, the attention scores and the corresponding value vectors are summed to obtain an output. Formally, given an input sequence $\mathbf{x}=\langle x_1,x_2,\cdots,x_{\left | \mathbf{x} \right |} \rangle$ with $x_i \in \mathbb{R}^d$, the process is expressed as below:

\begin{equation}
    \setlength{\abovedisplayskip}{4pt}
    \setlength{\belowdisplayskip}{4pt}
    \label{eq_a1}
    e_{i,j} = \frac{(x_{i}W_Q)(x_{j}W_K)^\top}{\sqrt{d}},
\end{equation}

\begin{equation}
    \setlength{\abovedisplayskip}{4pt}
    \setlength{\belowdisplayskip}{4pt}
    \label{eq_a2}
    \alpha_{i,j} = \frac{\exp(e_{i,j})}{\textstyle \sum_{k=1}^{\left | \mathbf{x} \right |} \exp(e_{i,k})},
\end{equation}

\begin{equation}
    \setlength{\abovedisplayskip}{4pt}
    \setlength{\belowdisplayskip}{4pt}
    \label{eq_a3}
    z_i= \sum_{j=1}^{\left | \mathbf{x} \right |} \alpha_{i,j}(x_{j}W_V),
\end{equation}
where $e_{i,j}$ is an unnormalized attention score, and $W_Q$, $W_K$ and $W_V$ are learnable transformation matrices. The scaling coefficient $1/\sqrt{d}$ is used for stable computation. $\alpha_{i,j}$ denotes the attention score that indicates the importance of element $x_j$ to element $x_i$. $z_i$ is the output that combines value vectors with the corresponding attention scores.

\begin{figure}
	\centering
	\includegraphics[scale=0.61]{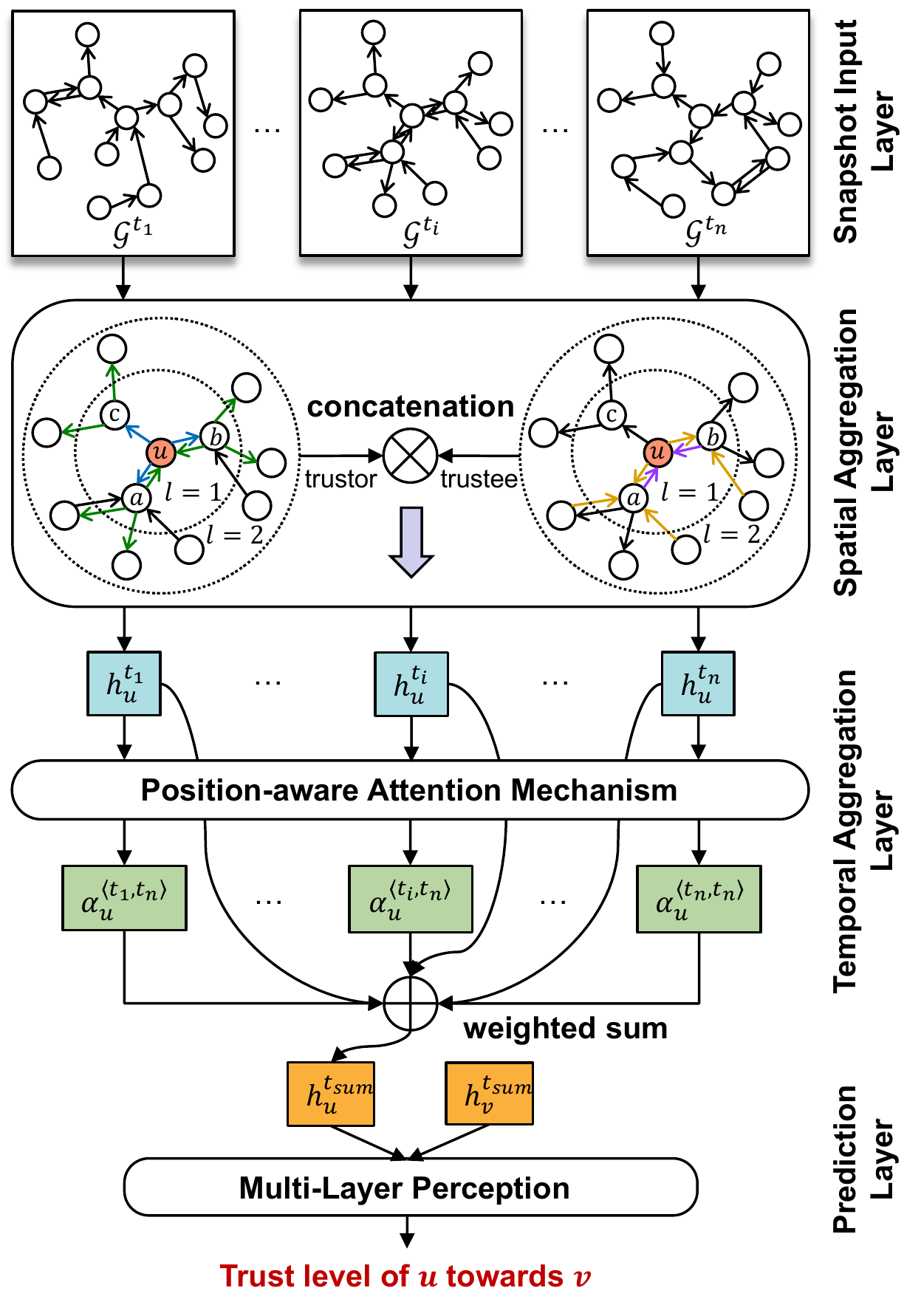}
	\caption{Overall architecture of TrustGuard. Lines of different colors in the spatial aggregation layer indicate the trust propagation and aggregation from the different-order neighbors of node $u$.}
	\label{scheme design}
        \vspace{-2mm}
\end{figure}

\section{TrustGuard Design} \label{section4}
In this section, we present the detailed design of TrustGuard. We first give a design overview and then introduce the key components of TrustGuard.

\subsection{Overview}
As illustrated in Fig.~\ref{scheme design}, TrustGuard consists of four layers, namely the snapshot input layer, the spatial aggregation layer, the temporal aggregation layer, and the prediction layer. In the snapshot input layer, a dynamic graph is segmented into a series of snapshots based on a time-driven strategy, and the snapshots are arranged in a chronological order for further analysis. The spatial aggregation layer focuses on the topological information in each given snapshot to generate the spatial embedding of each node. It achieves this by aggregating information from the first-order and high-order neighbors of the node with the help of a defense mechanism, while also considering the dual roles of the node. In addition, the temporal aggregation layer is designed to capture temporal patterns from a sequence of snapshots. A position-aware attention mechanism is employed herein to learn attention scores, which indicate the importance of each timeslot to the target timeslot. By employing a weighted sum to the spatial embeddings calculated across all snapshots, we can obtain the final embedding of each node that contains both spatial and temporal features. In the prediction layer, a Multi-Layer Perception (MLP) is constructed to transform the final embeddings of any two nodes into a directed trust relationship. In what follows, we present the key technical details of TrustGuard. We take node $u$ as an example to illustrate how its embedding is generated.

\subsection{Spatial Aggregation Layer}
This layer aims to learn node embeddings from a topological perspective, i.e., focusing on local trust relationships. Due to the asymmetry of trust, a node can play as either a trustor or a trustee. To fully consider the asymmetry nature, node embeddings should cover both roles. Specifically, a node’s out-degree can be considered as a trust interaction (i.e., edge) sent by that node as a trustor, while its in-degree can be regarded as a trust interaction it receives as a trustee. To aggregate these two types of information in a robust manner, robust coefficients are learned to prune potential malicious trust interactions. In the following, we first describe the process of first-order trust propagation, i.e., aggregating information from node $u$'s first-order neighbors. Then, we illustrate how this process can be generalized to high-order trust propagation. Finally, we discuss how to learn the robust coefficients to counter trust-related attacks.

\subsubsection{\textbf{First-order Trust Propagation}}
As shown in Fig. \ref{scheme design}, node $u$ can either be a trustor or a trustee. As a trustee, its first-order neighbors are node $a$ and node $b$. Note that node $c$ is not the first-order neighbor of node $u$ being a trustee since an arrow does not point to $u$ from $c$. For generating the embedding of $u$, we need to extract the information from the embeddings of $a$ and $b$ as well as their ratings (i.e., trust levels) towards $u$. These ratings are application-specific and can be modeled using several methods. For example, if $w_{u \gets v} \in \{Trust, Distrust\}$, we can model them as one-hot representations, i.e., $[1,0]^\top$ and $[0,1]^\top$. Since different applications have different levels of trust, it becomes essential to transform them into a vector space with the same dimension as the node embedding, as presented in Eq.~\ref{eq1}.
\begin{equation}
    \setlength{\abovedisplayskip}{4pt}
    \setlength{\belowdisplayskip}{4pt}
    \label{eq1}
    \omega_{u \gets v}=W_{te} \cdot w_{u \gets v},
\end{equation}
where $w_{u \gets v} \in \mathbb{R}^{\left | \mathcal{W} \right |}$ is the original rating sent from $v$ to $u$, $W_{te} \in \mathbb{R}^{d_e \times \left | \mathcal{W} \right |}$ is a learnable transformation matrix, and $\omega_{u \gets v}$ is a transformed rating. Based on this, a message referring to what information needs to be propagated from $u$'s neighbors to $u$ can be modeled as follows:
\begin{equation}
    \setlength{\abovedisplayskip}{4pt}
    \setlength{\belowdisplayskip}{4pt}
    \label{eq2}
    Msg_{u \gets v}=h_v \otimes \omega_{u \gets v},
\end{equation}
where $h_v \in \mathbb{R}^{d_e}$ denotes the embedding of a trustor $v$, $\omega_{u \gets v}$ stands for the rating from the trustor $v$ to the trustee $u$, $\otimes$ denotes a concatenation operation. Intuitively, the construction of this message can be understood as follows: $\omega_{u \gets v}$ looks like a recommendation from a trustor (i.e., $v$) to a trustee (i.e., $u$), while $h_v$ can be treated as the trustworthiness of the trustor. When aggregating the trustworthiness of a trustee, it is essential to consider trustors' trustworthiness and their recommendations towards the trustee. In other words, the features of node $u$ as a trustee, e.g., subjective and objective properties, can be extracted from the message $Msg_{u \gets v}$. Since $u$ may have several neighbors acting as trustors, we need to aggregate all relevant messages to fully capture its features as a trustee. The formal process is described as follows:
\begin{equation}
    \setlength{\abovedisplayskip}{4pt}
    \setlength{\belowdisplayskip}{4pt}
    \label{eq3}
    h_{u}(te)=\textbf{Aggregate}(\{Msg_{u \gets v}, \forall v \in N_{u}(te)\}),
\end{equation}
where $N_u(te)$ and $h_{u}(te)$ refer to the neighbor set and the embedding of node $u$ being a trustee, respectively.

There are several aggregators that can be used for aggregation in Eq.~\ref{eq3}. Guardian~\cite{lin2020guardian} applies a mean aggregator, which fails to distinguish the importance of different neighbors and their ratings. To address this problem, attention mechanisms are adopted in some approaches~\cite{jiang2022gatrust,lin2021medley}, which, however, do not take into account the presence of malicious ratings. As such, attention scores learned by the attention mechanism may not reflect reality. Inspired by previous work~\cite{wu2019adversarial,jin2020graph,zhang2020gnnguard}, we design a robust aggregator for defending against various trust-related attacks, which is introduced in Section~\ref{section_robustness}.

Correspondingly, $u$ can also be a trustor. The calculation process of its embedding when being a trustor is similar to the above processes, which can be expressed by the following equations:
\begin{equation}
    \setlength{\abovedisplayskip}{4pt}
    \setlength{\belowdisplayskip}{4pt}
    \label{eq4}
    \omega_{u \to v}=W_{tr} \cdot w_{u \to v},
\end{equation}
\begin{equation}
    \setlength{\abovedisplayskip}{4pt}
    \setlength{\belowdisplayskip}{4pt}
    \label{eq5}
    Msg_{u \to v}=h_v \otimes \omega_{u \to v},
\end{equation}
\begin{equation}
    \setlength{\abovedisplayskip}{4pt}
    \setlength{\belowdisplayskip}{4pt}
    \label{eq6}
    h_{u}(tr)=\textbf{Aggregate}(\{Msg_{u \to v}, \forall v \in N_{u}(tr)\}),
\end{equation}
where $W_{tr}$ is a learnable transformation matrix, and $N_u(tr)$ refers to the neighbor set of node $u$ as a trustor.

Since trustor and trustee are two roles of node $u$, $h_u(tr)$ and $h_u(te)$ should be considered jointly to form a complete spatial embedding of $u$. One approach is to concatenate $h_u(tr)$ and $h_u(te)$ directly, but this leads to high dimensions, expensive computational costs, as well as potentially duplicated information. Moreover, it also violates the definition of embedding, which is meant to capture representative features with a low-dimensional vector. Alternatively, we employ a fully-connected layer, a fundamental component in a neural network, to further mine key features from $h_u(tr)$ and $h_u(te)$. The fully-connected layer is capable of extracting high-level features from its input, and thus it is more effective than direct concatenation. By using the fully-connected layer, we can obtain a dense and informative spatial embedding of node $u$. The formal process is given in what follows:
\begin{equation}
    \setlength{\abovedisplayskip}{4pt}
    \setlength{\belowdisplayskip}{4pt}
    \label{eq7}
    h_u=\sigma (W_{both} \cdot (h_u(te) \otimes h_u(tr))+b_{both}),
\end{equation}
where $W_{both}$ and $b_{both}$ are learnable parameters, and $\sigma$ is a non-linear activation function (e.g., $ReLU$ \cite{eckle2019comparison}).

\subsubsection{\textbf{High-order Trust Propagation}}
In the first-order trust propagation, we generate node $u$'s embedding by aggregating messages from its first-order neighbors. However, this may not be sufficient since trust is conditionally transferable. This means information about trust can be propagated along a chain of recommendations. For capturing this characteristic, we stack $l$ trust propagation layers to enable node $u$ to receive diverse information about trust from its $l$-hop neighbors. As such, in the $l$-th layer, the embedding of node $u$ as a trustee is formally modeled as below:
\begin{equation}
    \setlength{\abovedisplayskip}{4pt}
    \setlength{\belowdisplayskip}{4pt}
    \label{eq8}
    \omega^l_{u \gets v}=W^l_{te} \cdot w_{u \gets v},
\end{equation}
\begin{equation}
    \setlength{\abovedisplayskip}{4pt}
    \setlength{\belowdisplayskip}{4pt}
    \label{eq9}
    Msg^l_{u \gets v}=h^{l-1}_{v} \otimes \omega^l_{u \gets v},
\end{equation}
\begin{equation}
    \setlength{\abovedisplayskip}{4pt}
    \setlength{\belowdisplayskip}{4pt}
    \label{eq10}
    h^l_{u}(te)=\textbf{Aggregate}(\{Msg^l_{u \gets v}, \forall v \in N_{u}(te)\}),
\end{equation}
where $W^l_{te}$ is a learnable parameter in the $l$-th layer for trustee nodes. Similarly, we can obtain the embedding of $u$ when it acts as a trustor using Eq. \ref{eq11}.
\begin{equation}
    \setlength{\abovedisplayskip}{4pt}
    \setlength{\belowdisplayskip}{4pt}
    \label{eq11}
    h^l_{u}(tr)=\textbf{Aggregate}(\{Msg^l_{u \to v}, \forall v \in N_{u}(tr) \}).
\end{equation}

Also, the complete spatial embedding of $u$ in the $l$-th layer is formulated by jointly considering $h^l_{u}(tr)$ and $h^l_{u}(te)$ using Eq. \ref{eq12}.
\begin{equation}
    \setlength{\abovedisplayskip}{4pt}
    \setlength{\belowdisplayskip}{4pt}
    \label{eq12}
    h^l_u=\sigma (W^l_{both} \cdot (h^l_u(te) \otimes h^l_u(tr))+b^l_{both}),
\end{equation}
where $W^{l}_{both}$ and $b^{l}_{both}$ are learnable parameters. Note that $h^{0}_{u}$ is the initial embedding of node $u$, which can be randomly initialized or pre-trained using walk-based methods such as node2vec \cite{grover2016node2vec}. 

To summarize, $h^{l}_u$ is the embedding that captures local topological information of node $u$ in the $l$-th layer. It fuses the information about trust from $u$'s $l$-hop neighbors with the consideration of $u$'s roles. By adjusting $l$, we can control the range of trust propagation. As suggested by Hamilton \textit{et al.}~\cite{hamilton2017inductive}, GNNs can  achieve the best performance when $l=2$ or 3. This can be explained by the fact that trust disappears when the propagation distance is too long~\cite{huo2023trustgnn}. We explore the impact of $l$ on evaluation accuracy in Section \ref{section_rq5}. 

We define $h_u^L$ as the embedding of $u$ in the last layer $L$. If the spatial aggregation is processed at timeslot $t_i$, $h_u^L$ can also be denoted as $h_u^{t_i}$, which serves as the output of the spatial aggregation layer for node $u$ at timeslot $t_i$.

\subsubsection{\textbf{Defense Mechanism against Trust-Related Attacks}}\label{section_robustness}
Malicious nodes in a graph can launch attacks such as bad-mouthing and good-mouthing attacks on targeted nodes, which could negatively impact the performance of a trust evaluation model. To mitigate these adverse effects, we design a robust aggregator based on the network theory of homophily~\cite{mcpherson2001birds}, insights about trust relationships~\cite{chen2014trust,wang2021c}, and analyses on adversarial attacks~\cite{wu2019adversarial,jin2020graph}. Our concrete justifications are presented as below. First, we calculate the edge homophily ratios of two trust networks, Bitcoin-OTC and Bitcoin-Alpha~\cite{snapnets}, following Zhu \textit{et al.}’s approach~\cite{Zhu2020nips}. The edge homophily ratios are 0.90 and 0.94, respectively, exceeding the 0.7 threshold set in~\cite{Zhu2020nips}. This indicates that these two trust networks are highly homophilous, thus connections are more likely to form between similar nodes~\cite{mcpherson2001birds}. Second, social similarity is often applied to filter trust feedback in trust evaluation approaches proposed in the literature~\cite{chen2014trust}. The intuition herein is that similar nodes are likely to have similar trust perspectives. Wang \textit{et al.}~\cite{wang2021c} conducted a significance test on the Epinions trust network~\cite{epinions} to study the correlation between trust relationships and node similarity. Their results reveal that trust relationships are more easily established between nodes with higher similarity, supported by a notably low $p$-value. Third, several studies~\cite{wu2019adversarial,jin2020graph} empirically demonstrated that adversarial attacks on GNNs tend to connect one target node to nodes with different features and labels (see Appendix~\ref{appendix_adversarial}), which has been regarded as the most powerful way to perform attacks. All these findings and observations underscore the importance of considering node similarity in trust propagation and aggregation. Built on those insights and previous analyses, the key idea of our defense mechanism is to assign high weights to edges between similar nodes, while low weights or pruning edges between unrelated ones. The concrete steps are described below.

\underline{\textit{Step 1.}} Calculate the similarity between any two nodes. There are a lot of similarity metrics available for calculating the similarity (i.e., distance) between two nodes, such as Jaccard similarity and Cosine similarity. Jaccard similarity is only designed for binary data, whereas Cosine similarity can be used for both continuous and binary data. Additionally, Cosine similarity can handle negative values and is useful for dealing with high-dimensional data. To this end, Cosine similarity is chosen in our design, which is calculated by
\begin{equation}
    \setlength{\abovedisplayskip}{4pt}
    \setlength{\belowdisplayskip}{4pt}
    \label{eq13}
    s^l_{u,v}=\frac{(h^{l-1}_u)^\top \cdot h^{l-1}_v}{\Vert h^{l-1}_u \Vert_2 \cdot \Vert h^{l-1}_v \Vert_2},
\end{equation}
where $\Vert \cdot \Vert_2$ calculates the $l_2$ norm of a vector. A high $s^l_{u,v}$ indicates that the edge between $u$ and $v$ in the $l$-th layer is robust and hard to destroy.

\underline{\textit{Step 2.}} Normalize $s^l_{u,v}$ within $u$'s neighbors. Note that $u$ can be either a trustor or a trustee, and the difference in normalization is that $u$ has different neighbor sets when playing as different roles. We use Eq. \ref{eq14} to calculate the normalized similarity between $u$ and its neighbors when $u$ is a trustee.
\begin{equation}
    \setlength{\abovedisplayskip}{4pt}
    \setlength{\belowdisplayskip}{4pt}
    \label{eq14}
    \hat{r}^l_{u \gets v} = \frac{s^l_{u,v}}{ {\textstyle \sum_{k \in N_u(te)}^{}} s^l_{u,k}}.
\end{equation}

\underline{\textit{Step 3.}} Prune edges with similarity below a pre-defined threshold $thr$, whose value setting is studied in Section \ref{section_rq5}.
\begin{equation}
    \setlength{\abovedisplayskip}{4pt}
    \setlength{\belowdisplayskip}{4pt}
\label{eq15}
\tilde{r}^l_{u \gets v}=
    \begin{cases}
        0 & \text{ if } \hat{r}^l_{u \gets v} < thr \\
        \hat{r}^l_{u \gets v} & \text{ otherwise }
    \end{cases}.
\end{equation}

\underline{\textit{Step 4.}} Re-normalize $\tilde{r}^l_{u \gets v}$ within $u$'s in-degree neighbors. Through this step, we can achieve the goal that edges connecting similar nodes are assigned high weights and vice versa.
\begin{equation}
    \setlength{\abovedisplayskip}{4pt}
    \setlength{\belowdisplayskip}{4pt}
    \label{eq16}
    r^l_{u \gets v} = \frac{\tilde{r}^l_{u \gets v}}{ {\textstyle \sum_{k \in N_u(te)}^{}} \tilde{r}^l_{u \gets k}}.
\end{equation}

The process of calculating $r^l_{u \to v}$ when $u$ is a trustor is similar to the process described above. We refer to $r^l_{u \gets v}$ ($r^l_{u \to v}$) as a robust coefficient for the edge sent from node $v$ ($u$) to node $u$ ($v$) in the $l$-th layer. In essence, it enhances the information aggregation from $u$’s high-coefficient neighbors, while weakens the information aggregation from $u$’s low-coefficient neighbors. As a consequence, Eq. \ref{eq10} and Eq. \ref{eq11} can be then revised as:
\begin{equation}
    \setlength{\abovedisplayskip}{4pt}
    \setlength{\belowdisplayskip}{4pt}
    \label{eq17}
    h^l_u(te)=\textstyle \sum_{v \in N_u(te)}^{} r^l_{u \gets v} \cdot Msg^l_{u \gets v},
\end{equation}
\begin{equation}
    \setlength{\abovedisplayskip}{4pt}
    \setlength{\belowdisplayskip}{4pt}
    \label{eq18}
    h^l_u(tr)=\textstyle \sum_{v \in N_u(tr)}^{} r^l_{u \to v} \cdot Msg^l_{u \to v}.
\end{equation}

\subsection{Temporal Aggregation Layer} This layer aims to learn temporal patterns from a series of chronologically sequenced snapshots. The input to this layer is a sequence of embeddings of a particular node $u$ at different timeslots. These embeddings are assumed to sufficiently capture topological information (i.e., local trust relationships) in each snapshot. Specifically, the input for each node $u$ is $\{h_u^{t_1},h_u^{t_2},\cdots,h_u^{t_n}\}$. Among them, $h_u^{t_i} \in \mathbb{R}^{d_e^{'}}$, where $t_i$ refers to the timeslot and $d^{'}_e$ refers to the dimension of the input embedding. The output of this layer is $h_u^{t_{sum}} \in \mathbb{R}^{d^{'}_e}$, which is assumed to cover both spatial and temporal features. In the following, we elaborate on the technical details.

The key component of this layer is a position-aware attention mechanism~\cite{zhang2017position}. In particular, a positional embedding $p^{t_i} \in \mathbb{R}^{d^{'}_e}$ is first encoded into an original input embedding $h^{t_i}_u$ using Eq. \ref{eq19}. This positional embedding represents the snapshot-specific information, which is randomly initialized and then learned during training. By incorporating this embedding, TrustGuard can effectively capture the context of each snapshot in the sequence and distinguish the importance of each snapshot/timeslot~\cite{Xu2020Inductive}. Then, the attention scores for the embeddings from each timeslot $t_i$ and the target timeslot $t_n$ are calculated following Eq. \ref{eq20}. Intuitively, these attention scores determine how much attention should TrustGuard pay to each timeslot when making predictions at the target timeslot, which also indicate temporal patterns learned by TrustGuard. To obtain the relative importance of each timeslot, Eq. \ref{eq21} is applied to normalize the attention scores. In the end, we fuse the embeddings across all timeslots by taking into account the importance of each timeslot via Eq. \ref{eq22}. Formally, the above process is represented as:
\begin{equation}
    \setlength{\abovedisplayskip}{4pt}
    \setlength{\belowdisplayskip}{4pt}
    \label{eq19}
    h^{t_i}_u \gets h^{t_i}_u + p^{t_i},
\end{equation}
\begin{equation}
    \setlength{\abovedisplayskip}{4pt}
    \setlength{\belowdisplayskip}{4pt}
    \label{eq20}
    \hat{\alpha}^{\langle t_i,t_n \rangle}_u = \frac{(W_Q h^{t_n}_u)^\top W_K h^{t_i}_u}{\sqrt{d^{'}_e}},
\end{equation}
\begin{equation}
    \setlength{\abovedisplayskip}{4pt}
    \setlength{\belowdisplayskip}{4pt}
    \label{eq21}
    \alpha^{\langle t_i,t_n \rangle}_u = \frac{\exp(\hat{\alpha}^{\langle t_i,t_n \rangle}_u)}{\sum_{k=1}^{n} \exp(\hat{\alpha}^{\langle t_k,t_n \rangle}_u)},
\end{equation}
\begin{equation}
    \setlength{\abovedisplayskip}{4pt}
    \setlength{\belowdisplayskip}{4pt}
    \label{eq22}
    h^{t_{sum}}_u=\sum_{i=1}^{n} \alpha^{\langle t_i,t_n \rangle}_u W_V h^{t_i}_u,
\end{equation}
where $W_Q, W_K, W_V \in \mathbb{R}^{d^{'}_e \times d^{'}_e}$ are learnable transformation matrices in the self-attention mechanism.

It is worth noting that Eq. \ref{eq19}-\ref{eq22} complete the definition of single-head attention, i.e., to learn a set of parameters $W_Q$, $W_K$ and $W_V$. As suggested by~\cite{gat2018}, we further use multi-head attention, i.e. to learn multiple sets of parameters. This enhances the learning ability and allows TrustGuard to jointly learn temporal patterns from different representation subspaces. In detail, the multi-head mechanism executes multiple single-head attentions in parallel. After that, the outputs from different heads are concatenated to create the final embedding of each node, formulated as:
\begin{equation}
\setlength{\abovedisplayskip}{4pt}
\setlength{\belowdisplayskip}{4pt}
    \label{eq23}
    h^{t_{sum}}_u \gets h^{t_{sum}}_u(1) \otimes h^{t_{sum}}_u(2) \otimes \cdots \otimes h^{t_{sum}}_u(S),
\end{equation}
where $S$ is the number of attention heads, and $h_u^{t_{sum}} \in \mathbb{R}^{d^{'}_e}$ is the final embedding of node $u$ with the consideration of its local trust relationships and the temporal patterns of ordered snapshots.

\subsection{Prediction Layer}
With the help of spatial and temporal aggregation, rich information is encoded into the node embedding $h^{t_{sum}}_u$. It can be used for various prediction tasks, including but not limited to node trust prediction and directed trust relationships prediction. In this paper, we focus on the second case as it is more complex. To predict the latent trust relationship between any two nodes, we adopt a Multi-Layer Perception (MLP) as the prediction model and feed their embeddings into it. The reason for choosing MLP is that it is simple and effective enough. Since the embeddings already cover representative features of the nodes, there is no need to employ other complex ML models to further extract features and perform prediction tasks. Mathematically, the process can be represented as:
\begin{equation}
    \setlength{\abovedisplayskip}{4pt}
    \setlength{\belowdisplayskip}{4pt}
    \label{eq24}
    p_{u \to v}=\sigma (W \cdot (h^{t_{sum}}_u \otimes h^{t_{sum}}_v) + b),
\end{equation}
where $W$ and $b$ are learnable parameters, $\sigma$ is the $softmax$ function, and $p_{u \to v}$ is the probabilistic prediction vector of the trust level from $u$ to $v$. Consequently, the predicted trust level is denoted as $argmax_j (p_{u \to v})$, where the index $j$ corresponds to the maximum value of $p_{u \to v}$. It is worth noting that $p_{u \to v}$ may be not equal to $p_{u \gets v}$ due to the asymmetry of trust. The procedure of TrustGuard is presented in \textbf{Algorithm \ref{TrustGuard_algorithm}}.

To train the TrustGuard model, a weighted cross-entropy loss function is employed to measure the difference between the predicted trust relationships and the ground truth. The reason for choosing this loss function is that it is good at dealing with imbalanced data and can speed up the model's convergence during training. Specifically, the loss function can be formulated as:
\begin{equation}
    \setlength{\abovedisplayskip}{4pt}
    \setlength{\belowdisplayskip}{4pt}
    \label{eq25}
    \mathcal{L} = - \sum_{e_{u \to v} \in \mathcal{E}(T)} \beta_{u \to v} \cdot \log_{}{p_{u \to v, w_{u \to v}}} + \lambda \cdot \Vert \Theta \Vert^2_2,
\end{equation}
where $\beta_{u \to v}$ denotes the weight of the edge $e_{u \to v}$, $w_{u \to v}$ denotes the ground truth, $\Theta$ denotes all the learnable parameters used in the model, and $\lambda$ controls the $L_2$ regularization to prevent overfitting \cite{lewkowycz2020training}. The Adam optimizer~\cite{adam} is used to update the model parameters.

\begin{algorithm}[!t]
\footnotesize
\SetCommentSty{small}
\LinesNumbered
\caption{TrustGuard Procedure}
\label{TrustGuard_algorithm}

\KwIn{A dynamic graph $\mathcal{G}(T)$,
initial node embedding $h^0_u$;
}
\KwOut{$argmax_j (p_{u \to v})$;}

\Comment{\textbf{Snapshots Formation}}

Segment $\mathcal{G}(T)$ into a sequence of ordered snapshots $\{\mathcal{G}^{t_1},\mathcal{G}^{t_2},\cdots,\mathcal{G}^{t_n}\}$ based on equal time intervals;

\Comment{\textbf{Spatial Aggregation}}

\For{$\mathcal{G}^{t_i} \in \{\mathcal{G}^{t_1},\mathcal{G}^{t_2},\cdots,\mathcal{G}^{t_n}\}$}
    {
    \For{$l \gets 1$ \KwTo $L$}
        {
        Calculate $s^l_{u,v}$ of each pair using Eq. \ref{eq13};
        
        \For{$u \in  \mathcal{V}^{t_i}$}
            {
            Calculate $r^l_{u \gets v}$ and $r^l_{u \to v}$ using Eq. \ref{eq14}-\ref{eq16};
            
            Form $u$'s two embeddings when being a trustor and a trustee using Eq. \ref{eq17}-\ref{eq18};

            Form $u$'s spatial embedding $h^l_u$ using Eq. \ref{eq12};
            }
        }
        $h^{t_i}_u \gets h^L_u$;
    }

\Comment{\textbf{Temporal Aggregation}}
    
\For{$u \in \mathcal{V}(T)$}
    {
    Obtain $\{h^{t_1}_u,h^{t_2}_u,\cdots,h^{t_n}_u\}$ through the spatial aggregation;

    Calculate $\alpha^{\langle t_i,t_n \rangle}_u$ using Eq. \ref{eq19}-\ref{eq21};

    Form $u$'s final embedding $h^{t_{sum}}_u$ using Eq. \ref{eq22};
    }

\Comment{\textbf{Trust Relationship Prediction}}

\For{$e_{u \to v} \in \mathcal{E}(T)$}
    {
    Calculate $p_{u \to v}$ using Eq. \ref{eq24};

    Obtain the trust level of $u$ towards $v$ via $argmax_j (p_{u \to v})$;
    }
\end{algorithm}

\begin{table*}[]
\footnotesize
\centering
\caption{Statistics of Bitcoin-OTC and Bitcoin-Alpha datasets.}
\label{dataset}
\begin{tabular}{c|c|c|c|c|c}
\toprule[1.5pt]
Dataset       & \# Nodes & \# Trust Edges & \# Distrust Edges & Avg. Degree & Temporal Information       \\ \midrule
Bitcoin-OTC   & 5881     & 32029          & 3563              & 12.1        & Nov 8, 2010 - Jan 24, 2016 \\
Bitcoin-Alpha & 3775     & 22650          & 1536              & 12.79       & Nov 7, 2010 - Jan 21, 2016 \\
\bottomrule[1.5pt]
\end{tabular}
\vspace{-2mm}
\end{table*}

\section{Experimental Evaluation} \label{section5}
In this section, we first introduce the experimental settings. Then, we answer the following research questions to demonstrate the effectiveness of TrustGuard.
\begin{itemize}
\item \textbf{RQ1:} How does TrustGuard perform compared with the state-of-the-art GNN-based trust evaluation models without experiencing trust-related attacks?
\item \textbf{RQ2:} How does TrustGuard perform compared with the state-of-the-art GNN-based trust evaluation models under different attacks?
\item \textbf{RQ3:} Are the results of TrustGuard explainable through visualization?
\item \textbf{RQ4:} How does each component contribute to the performance of TrustGuard? (Ablation study)
\item \textbf{RQ5:} How sensitive of TrustGuard to its hyperparameters?
\end{itemize}

\subsection{Experimental Settings}
\subsubsection{Datasets} Our experiments are conducted based on two real-world datasets with temporal information, i.e., Bitcoin-OTC and Bitcoin-Alpha~\cite{snapnets}. Note that the Advogato~\cite{massa2009bowling} dataset is not applicable in our context since it does not provide any temporal information. Both Bitcoin-OTC and Bitcoin-Alpha datasets are collected from an open market where users can make transactions using Bitcoins. As Bitcoin users are anonymous, it is important to keep track of users' trustworthiness to avoid transactions with malicious users. To achieve this, users in the open market can rate each other positively or negatively to express trust or distrust towards one another. These rating interactions have temporal information, allowing us to construct a dynamic graph that evolves over time. We segment each graph into 10 snapshots according to regular intervals and arrange them chronologically. We further investigate the sensitivity of TrustGuard to different observation frequencies in Appendix~\ref{appdendix_snapshots}. Table~\ref{dataset} presents the statistics of the two datasets.

Due to the lack of malicious edges in the above two datasets, we follow the approach proposed by Duan \textit{et al.}~\cite{duan2020aane} to manually inject some malicious edges for evaluation purposes. This approach is important and prevalent in graph anomaly detection research~\cite{Liu2023tbd,ma2023tkde}. We focus on collaborative bad-mouthing and good-mouthing attacks, where each target node is attacked by a group of nodes, since these types of attacks are more destructive than individual ones~\cite{chen2015trust,chen2014trust}. For collaborative bad-mouthing attacks, we first randomly select 10\% good nodes in the test set as target nodes. A node is regarded as good if it connects to more positively rated edges than negatively rated edges~\cite{mo2022motif}. Then, we use the approach described in~\cite{duan2020aane} to add negative ratings (i.e., edges with negative weights) to the target nodes. For collaborative good-mouthing attacks, we first select all the bad nodes in the test set as target nodes, since the number of bad nodes occupies a small percentage in both datasets. Then, using the same approach~\cite{duan2020aane}, we add positive ratings (i.e., edges with positive weights) to these nodes. For on-off attacks~\cite{chen2014trust}, we enable on-off attackers to perform bad-mouthing behavior at timeslot $t_i$ while behaving honestly at timeslot $t_{i+1}$, where $i=1,3,5,\cdots$. We pay more attention to the bad-mouthing attack due to its higher destructiveness compared to the good-mouthing attack. In addition, we set the number of malicious edges inserted to each target node as its degree to maximum adverse effects~\cite{yin20222}. This is reasonable as excessively inserting edges can easily destroy the original graph structure and make malicious edges easily detectable.

\subsubsection{Baselines} To demonstrate the effectiveness of TrustGuard, we compare it with state-of-the-art GNN-based trust evaluation models, including Guardian~\cite{lin2020guardian} and GATrust~\cite{jiang2022gatrust}. The reason for not choosing TrustGNN~\cite{huo2023trustgnn} is that it shows less improvement over Guardian compared with GATrust. The reason for not comparing Medley (a time-stamped model)~\cite{lin2021medley} with TrustGuard (a snapshot-based model) is that they hold different assumptions regarding the availability of fine-grained timestamps and the types of processable inputs. As such, they are considered orthogonal approaches for dynamic graph analysis, and the choice between them depends on a specific context and a goal. Therefore, they are not directly compared in experiments, as is commonly done in current studies~\cite{pareja2020evolvegcn,Euler,sankar2020dysat}. The details of the baselines are as follows:
\begin{itemize}
    \item \textbf{Guardian} \cite{lin2020guardian}: Guardian is the first approach that applies GNNs to trust evaluation. It achieves trust propagation using the notion of localized graph convolutions.
    \item \textbf{GATrust} \cite{jiang2022gatrust}: GATrust leverages multi-aspect properties of nodes and assigns different weights to each property using the key concept of GAT~\cite{gat2018}.
\end{itemize}

\subsubsection{Evaluation Metrics} As shown in Table~\ref{dataset}, both datasets exhibit high levels of class imbalance. To this end, it becomes necessary to select evaluation metrics that have no bias to any class. Guided by~\cite{luque2019impact}, we select Matthews Correlation Coefficient (MCC), AUC, Balanced Accuracy (BA), and F1-macro as our metrics to fairly assess the performance of trust evaluation models based on the imbalanced datasets.     
\begin{itemize}
    \item \textbf{Matthews Correlation Coefficient (MCC)}: MCC takes into account True Positive (TP), True Negative (TN), False Positive (FP), and False Negative (FN).
    
    \begin{footnotesize}
    \begin{equation}
    MCC = \frac{TP \cdot TN - FP \cdot FN}{\sqrt{(TP+FP)(TP+FN)(TN+FP)(TN+FN)}}.
    \nonumber
    \end{equation}
    \end{footnotesize}
    
    \item \textbf{AUC}: AUC is the area under the ROC curve, which shows the trade-off between True Positive Rate (TPR) and False Positive Rate (FPR) for different thresholds.

    \begin{footnotesize}
    \begin{equation}
    AUC = \frac{\textstyle \sum_{i \in P}^{} rank_i - |P|(|P|+1) / 2}{|P| \cdot |N|}.
    \nonumber
    \end{equation}
    \end{footnotesize}
    
    where $P$ and $N$ are a positive set and a negative set, respectively. $rank_i$ refers to the rank of sample $i$ with respect to its prediction score.
    
    \item \textbf{Balanced Accuracy (BA)}: BA is calculated as the average of True Positive Rate (TPR) and True Negative Rate (TNR). It assigns the same weight to both positive and negative classes.

    \begin{footnotesize}
    \begin{equation}
    BA = \frac{TP/(TP+FN) + TN/(TN+FP)}{2}.
    \nonumber
    \end{equation}
    \end{footnotesize}
    
    \item \textbf{F1-macro}: F1-macro is the unweighted mean of the F1 score of each class. The formula of the F1 score per class is given below:

    \begin{footnotesize}
    \begin{equation}
    F_1 = \frac{TP}{TP+(FP+FN)/2}.
    \nonumber
    \end{equation}
    \end{footnotesize}
\end{itemize}

Note that AUC, BA and F1-macro all range from [0,1], where 1 indicates perfect classification performance and 0 indicates poor classification performance. MCC ranges from [-1,1] with a higher value indicating better performance.

\subsubsection{Implementation Details} We implement our model using Pytorch\footnote{https://github.com/Jieerbobo/TrustGuard}, use the official version of Guardian~\cite{lin2020guardian}, and implement GATrust~\cite{jiang2022gatrust} by ourselves. Since both Bitcoin-OTC and Bitcoin-Alpha datasets do not have node features required in GATrust, we randomly initialize 8-dimensional features for each node to make GATrust applicable. To initialize node embeddings $h^0_u$, we adopt node2vec~\cite{grover2016node2vec} and fix the embedding dimension to 64 for both datasets. Following \cite{lin2020guardian,jiang2022gatrust}, all of the models have three propagation layers, where the first layer has an output dimension of 32, the second layer 64, and the third layer 32. To ensure a fair comparison, we also equip all models with a weighted cross-entropy loss function to handle imbalanced datasets. For training, the maximum epoch is set as 50, and an early stop strategy is adopted. We perform a grid search for hyperparameters and set the learning rate as 0.005, the structural dropout as 0, the temporal dropout as 0.5, and the $L_2$ normalization coefficient as $10^{-5}$. Other hyperparameters are analyzed in Section \ref{section_rq5}. Our experiments were conducted on a machine equipped with an Intel Xeon Processor (Skylake, IBRS) CPU, 56GB RAM, 80GB SSD, and NVIDIA Tesla T4 GPU. We report the average results obtained from 5 runs for each experiment.

\begin{table*}[t]
	\centering 
	\footnotesize
	\caption{Performance comparison of TrustGuard with different baselines on three types of tasks.}
        \label{without_attacks}
    {\vspace{-2mm}In each column, the best result is highlighted in \textbf{bold}. Improvement is calculated relative to the best baseline. \ding{172} denotes the single-timeslot prediction on observed nodes. \ding{173} denotes the multi-timeslot prediction on observed nodes. \ding{174} denotes the single-timeslot prediction on unobserved nodes.}
    \\[2mm] 

	\begin{tabular}{@{}c|c|cccc|cccc@{}}
		\toprule[1.5pt]
    \multirow{2.5}{*}{\makebox[0.05\textwidth][c]{Task}} &\multirow{2.5}{*}{Model} &\multicolumn{4}{c|}{Bitcoin-OTC} &\multicolumn{4}{c}{Bitcoin-Alpha}\\
    \cmidrule{3-10}
		& &MCC &AUC &BA &F1-macro &MCC &AUC &BA &F1-macro \\
   
   \midrule
		  \multirow{3}{*}{\ding{172}} & Guardian &0.351$\pm$0.007 &0.744$\pm$0.003 &  0.653$\pm$0.006 & 0.668$\pm$0.002 &0.328$\pm$0.012 &0.734$\pm$0.009& 0.671$\pm$0.004 & 0.649$\pm$0.004 \\
		& GATrust &0.346$\pm$0.005 &0.743$\pm$0.004 &0.654$\pm$0.008 &0.660$\pm$0.007 &0.329$\pm$0.009 &0.729$\pm$0.004& 0.663$\pm$0.004 &0.648$\pm$0.010 \\
		& \textbf{TrustGuard} & \textbf{0.389$\pm$0.007} &\textbf{0.765$\pm$0.008} & \textbf{0.693$\pm$0.008} &\textbf{0.687$\pm$0.007} &\textbf{0.362$\pm$0.004}&\textbf{0.756$\pm$0.009}& \textbf{0.692$\pm$0.004} &\textbf{0.669$\pm$0.002} \\
    \midrule
		 \multicolumn{2}{c|}{Improvement (\%)} &+10.83\% &+2.82\% &+5.96\% &+2.84\% &+10.03\% &+3.00\% &+3.13\% &+3.08\%\\

   \midrule
		  \multirow{3}{*}{\ding{173}} & Guardian &0.295$\pm$0.005 &0.715$\pm$0.003 &  0.622$\pm$0.001 & 0.639$\pm$0.001 &0.260$\pm$0.006 &0.673$\pm$0.003 & 0.618$\pm$0.005 & 0.624$\pm$0.003 \\
		& GATrust &0.290$\pm$0.002 &0.714$\pm$0.002 &0.622$\pm$0.002 &0.636$\pm$0.003 &0.257$\pm$0.004 &0.675$\pm$0.003 &0.615$\pm$0.005 &0.621$\pm$0.002 \\
		& \textbf{TrustGuard} & \textbf{0.330$\pm$0.005} &\textbf{0.725$\pm$0.004} & \textbf{0.642$\pm$0.003} &\textbf{0.658$\pm$0.003} &\textbf{0.288$\pm$0.002}&\textbf{0.692$\pm$0.003}& \textbf{0.632$\pm$0.006} &\textbf{0.639$\pm$0.001} \\
    \midrule
		 \multicolumn{2}{c|}{Improvement (\%)} &+11.86\% &+1.40\% &+3.22\% &+2.97\% &+10.77\% &+2.52\% &+2.27\% &+2.40\%\\

   \midrule
   \multirow{3}{*}{\ding{174}} & Guardian &0.447$\pm$0.019 &0.709$\pm$0.016 &  0.667$\pm$0.004 & 0.693$\pm$0.005 &0.325$\pm$0.012 &0.678$\pm$0.015 & 0.631$\pm$0.010 & 0.641$\pm$0.005 \\
		& GATrust &0.430$\pm$0.014 &0.712$\pm$0.011 &0.672$\pm$0.006 &0.691$\pm$0.006 &0.320$\pm$0.015 &0.681$\pm$0.004 &0.627$\pm$0.008 &0.636$\pm$0.004 \\
		& \textbf{TrustGuard} & \textbf{0.463$\pm$0.020} &\textbf{0.727$\pm$0.014} & \textbf{0.673$\pm$0.009} &\textbf{0.701$\pm$0.009} &\textbf{0.384$\pm$0.026}&\textbf{0.715$\pm$0.027}& \textbf{0.654$\pm$0.012} &\textbf{0.678$\pm$0.013} \\
    \midrule
		 \multicolumn{2}{c|}{Improvement (\%)} &+3.58\% &+2.11\% &+0.15\% &+1.15\% &+18.15\% &+4.99\% &+3.65\% &+5.77\%\\
	\bottomrule[1.5pt]
	\end{tabular}
	\vspace{-2mm}
\end{table*}

\subsection{Effectiveness of TrustGuard without Attacks (RQ1)}
In this subsection, we conduct experiments on three types of trust prediction tasks to fully test the efficacy of TrustGuard and baselines.

\subsubsection{Single-timeslot Prediction on Observed Nodes} This task is commonly conducted for evaluating the quality of embeddings learned by dynamic graph models~\cite{Euler}. Specifically, it involves using node embeddings trained on snapshots up to a given timeslot $t$, and then predicting trust relationships between previously observed nodes at the subsequent timeslot $t+1$. The observed nodes are defined as those that have edges (i.e., trust relationships) in any of the previous $t$ snapshots. As both datasets are segmented into 10 snapshots, there are 8 subtasks for evaluation, i.e., $\{1 \sim t\} \to t+1$ $(2 \le t \le 9)$. The reason for starting $t$ from 2 is to ensure that TrustGuard has at least 2 snapshots for training, enabling it to capture temporal patterns. We calculate macro metrics, such as macro MCC and macro AUC, by taking the average of the metrics obtained from each subtask. We report these results with their standard deviations in Table~\ref{without_attacks}. From the table, we find that TrustGuard consistently outperforms baseline models on both datasets. Specifically, TrustGuard achieves more than 10\% improvement on MCC on both datasets compared to the best baseline. This result highlights the significance of dynamic modeling in trust prediction. In addition, we observe that Guardian and GATrust exhibit similar performance. Although GATrust has better prediction performance than Guardian in static prediction tasks thanks to its attention mechanism, as demonstrated in~\cite{jiang2022gatrust}, it fails to maintain this advantage in dynamic prediction. This further indicates the difficulty of dynamic trust prediction.

\subsubsection{Multi-timeslot Prediction on Observed Nodes} This task is designed to assess the effectiveness of trust prediction models in predicting trust relationships across multiple future timeslots. It requires a model to be able to capture the underlying temporal patterns from a sequence of snapshots; otherwise it performs poorly. In this task, the model is first trained using $t$ previous snapshots, and then it is used to predict trust relationships between observed nodes in the next several timeslots $t+x$ $(1 \le x \le \triangle)$. We choose $\triangle = 3$ for evaluation. In this regard, there are 6 subtasks, i.e., $\{1 \sim t\} \to \{t+1,t+2,t+3\}$ $(2 \le t \le 7)$. As shown in Table~\ref{without_attacks}, TrustGuard still achieves significant improvements in comparison to baselines, with over 10\% improvements on MCC on both datasets. It is worth noting that all of the models have a decay in performance in this task than in the above single-timeslot prediction task. This is expected as this task involves more testing data than the first task, which requires the model to capture temporal patterns that may continue to emerge over multiple future timeslots. Despite the challenge of this task, TrustGuard exhibits the best performance, which demonstrates the strong capability of the attention mechanism in learning meaningful temporal patterns.

\begin{figure}[t]
    \centering
    \subfigcapskip=-3pt
    \hspace{4mm}
    \subfigure{\includegraphics[scale=0.24]{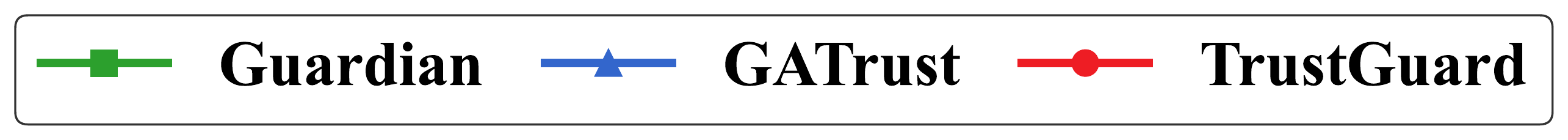}}
    \vspace{-3mm}

    \setcounter{subfigure}{0}
    \subfigure[Bitcoin-OTC]{\includegraphics[scale=0.24]{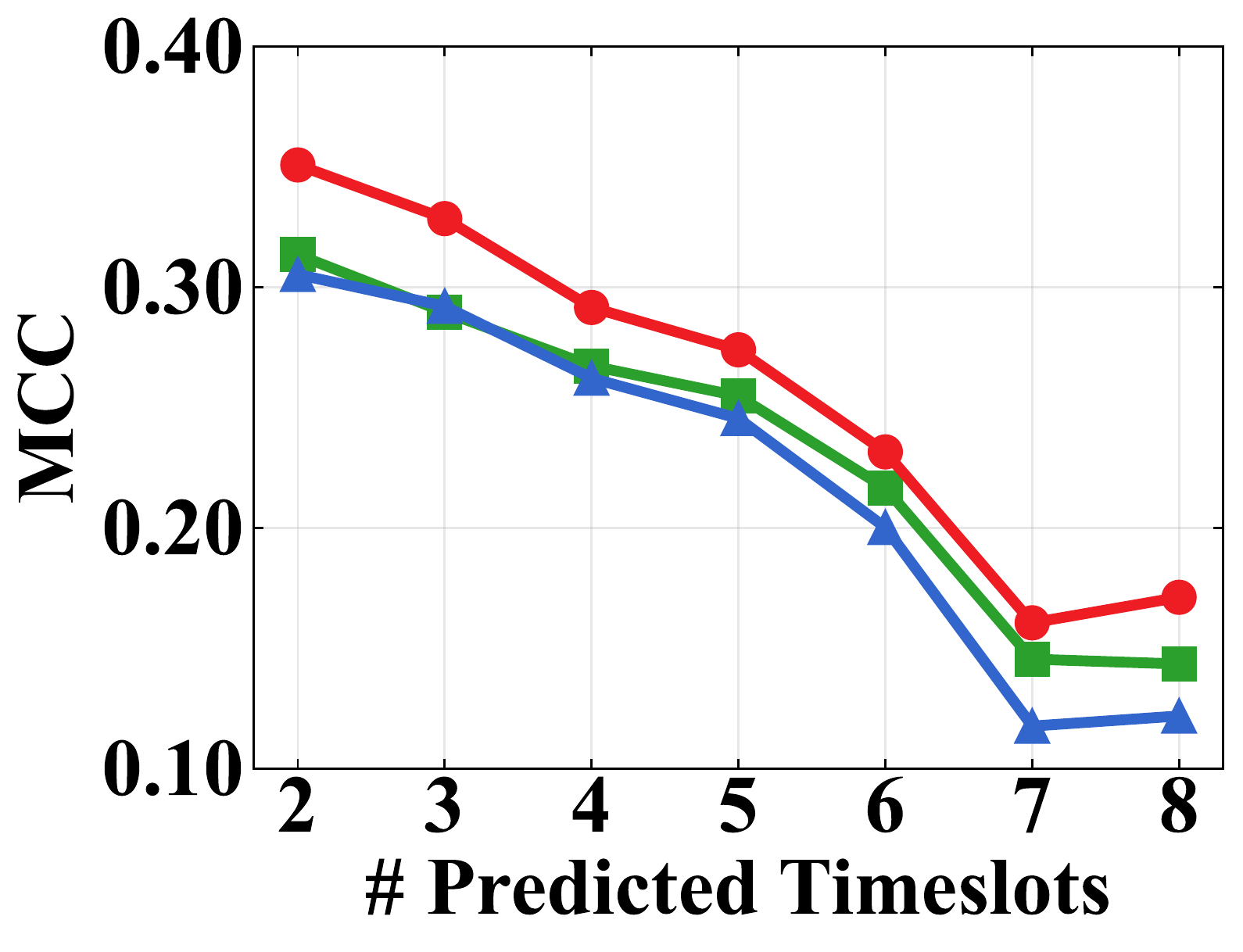}}
    \subfigure[Bitcoin-Alpha]{\includegraphics[scale=0.24]{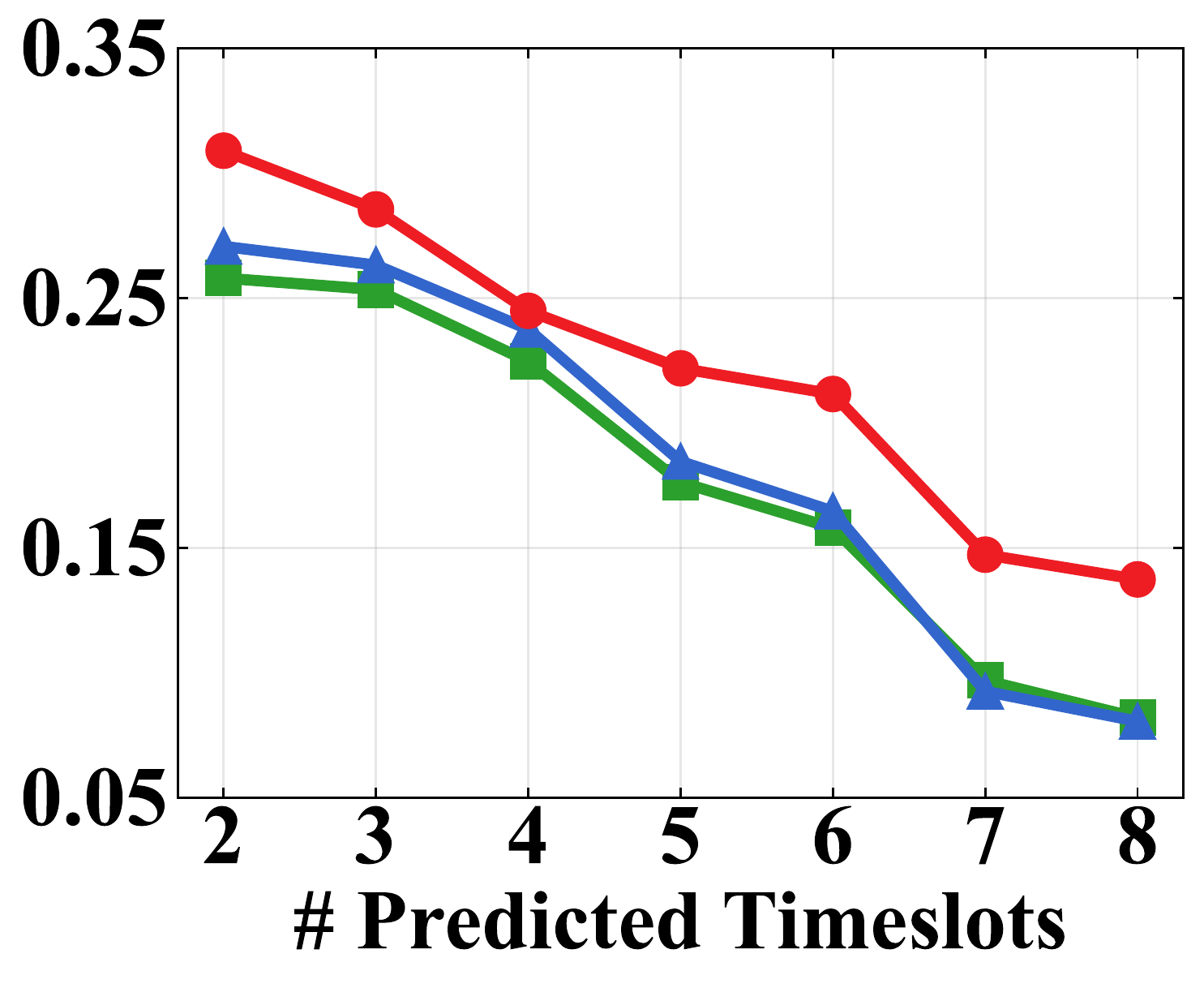}}

    \vspace{-3mm}
    \subfigure[Bitcoin-OTC]{\includegraphics[scale=0.24]{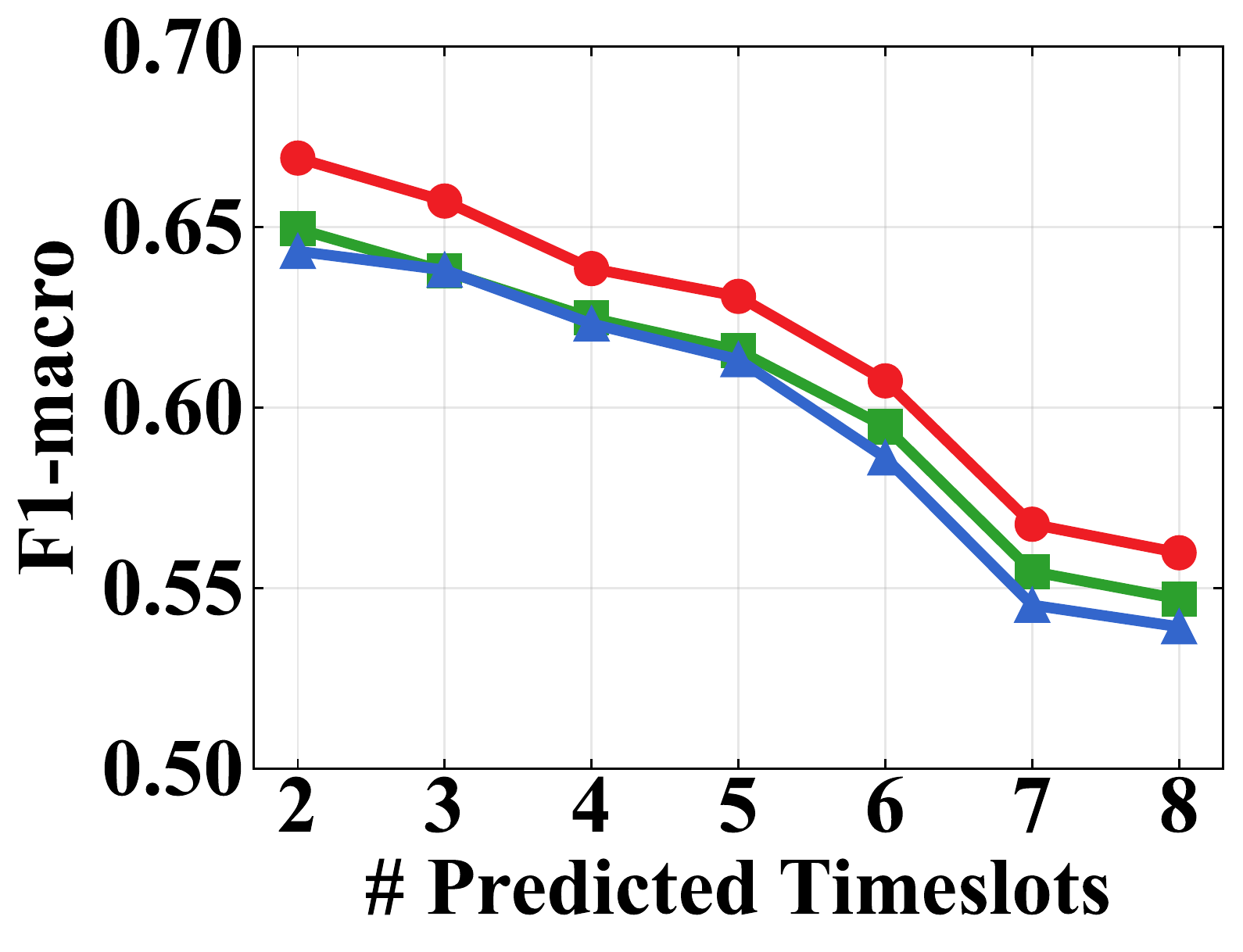}}
    \subfigure[Bitcoin-Alpha]{\includegraphics[scale=0.24]{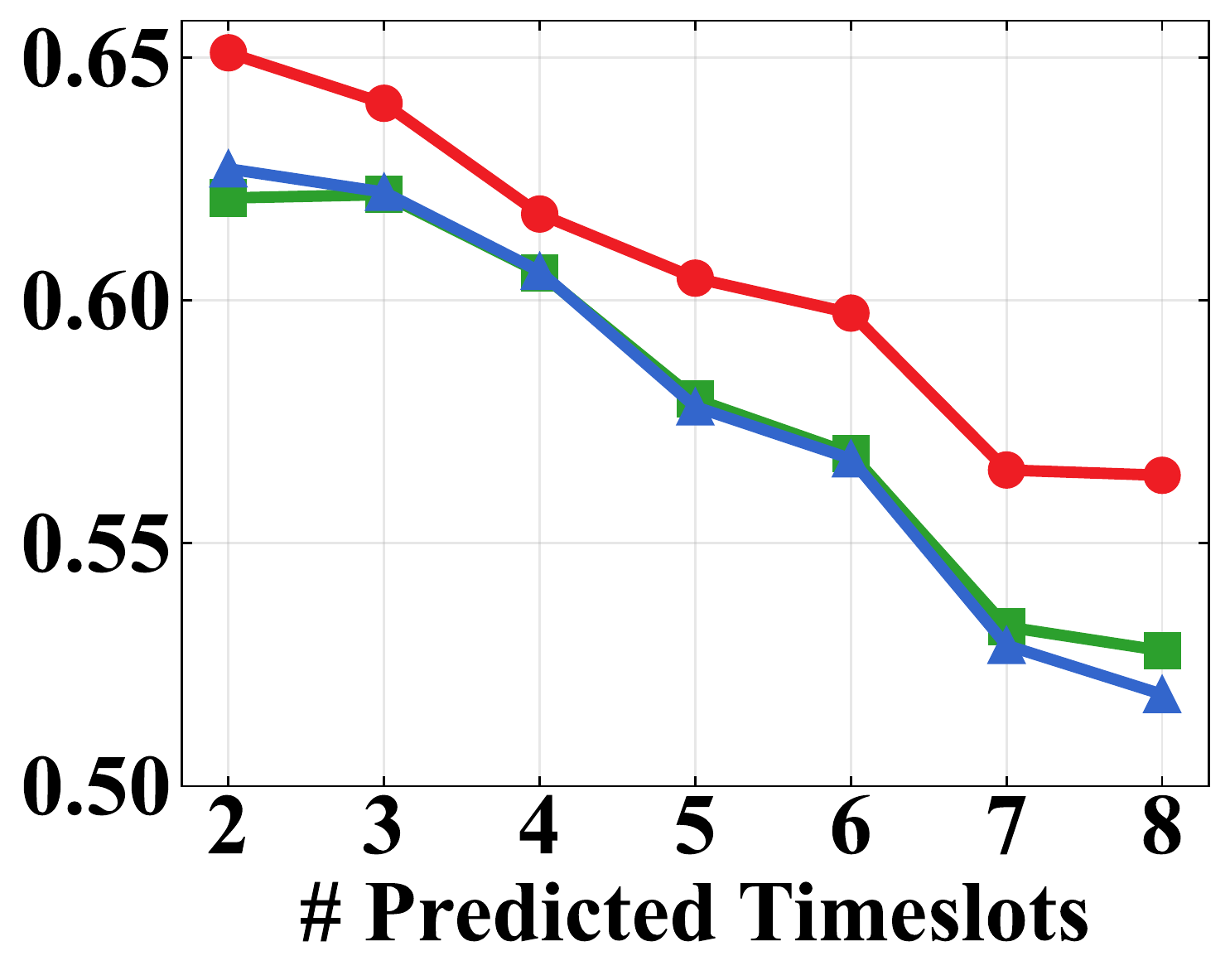}}

    \vspace{-2mm}
    \caption{Experimental results of trust prediction over multiple future timeslots.}
    \label{multi_slots}
    \vspace{-2mm}
\end{figure}

To further investigate the ability of TrustGuard in predicting future trust relationships, we conduct experiments with varying numbers of predicted timeslots, specifically, $\triangle$ starting from 2 to 8. From Fig.~\ref{multi_slots}, we observe a notable decay in performance as the number of predicted timeslots increases regarding all the models. The reasons are two-fold. Firstly, the increase of the number of predicted timeslots makes it more challenging to accurately predict future trust relationships. Secondly, temporal patterns may disappear in some future timeslots, which further causes performance decrease. Despite these challenges, TrustGuard has demonstrated exceptional performance across all settings. It achieves an average improvement of 11.09\% and 27.69\% on MCC regarding Bitcoin-OTC and Bitcoin-Alpha, respectively.

\subsubsection{Single-timeslot Prediction on Unobserved Nodes} The key characteristic of dynamic graphs is that nodes can join or exit at any time. This suggests that some nodes may be unobserved in the process of temporal evolution. To this end, we set this task to investigate whether TrustGuard can learn effective embeddings for unobserved nodes. Herein, a node is considered unobserved or new at timeslot $t$ if it has no edges (i.e., does not appear) in any of previous $t-1$ snapshots. Similar to the first task, this task also has 8 subtasks. The difference is that the first task focuses on observed nodes, while this task focuses on unobserved nodes. Thus, the testing sets used for the two tasks are different. The results in Table~\ref{without_attacks} show that TrustGuard consistently performs better than the baselines, particularly on the Bitcoin-Alpha dataset, where it improves MCC by almost 18\%. This indicates that TrustGuard can effectively characterize previously unobserved nodes. Although an unobserved node does not appear in the previous $t-1$ snapshots, its embedding is influenced by its neighboring observed nodes whose embeddings cover temporal patterns. We argue that this fact indirectly facilitates the encoding of the temporal patterns into the embeddings of unobserved nodes. Thus, TrustGuard can overcome the challenging cold start issue~\cite{wang2020survey,wang2022survey} of trust evaluation to some extent.

\begin{figure*}[!htb]
    \centering
    \subfigcapskip=-3pt
    \hspace{3mm}
    \subfigure{\includegraphics[scale=0.52]{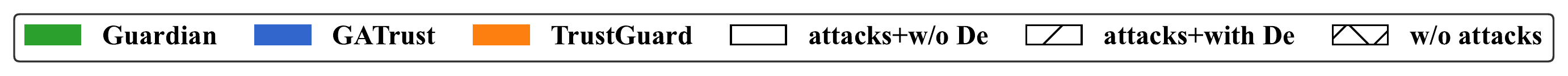}}
    \vspace{-3mm}

    \setcounter{subfigure}{0}
    \subfigure[Collaborative Bad-mouthing Attack]{\includegraphics[scale=0.26]{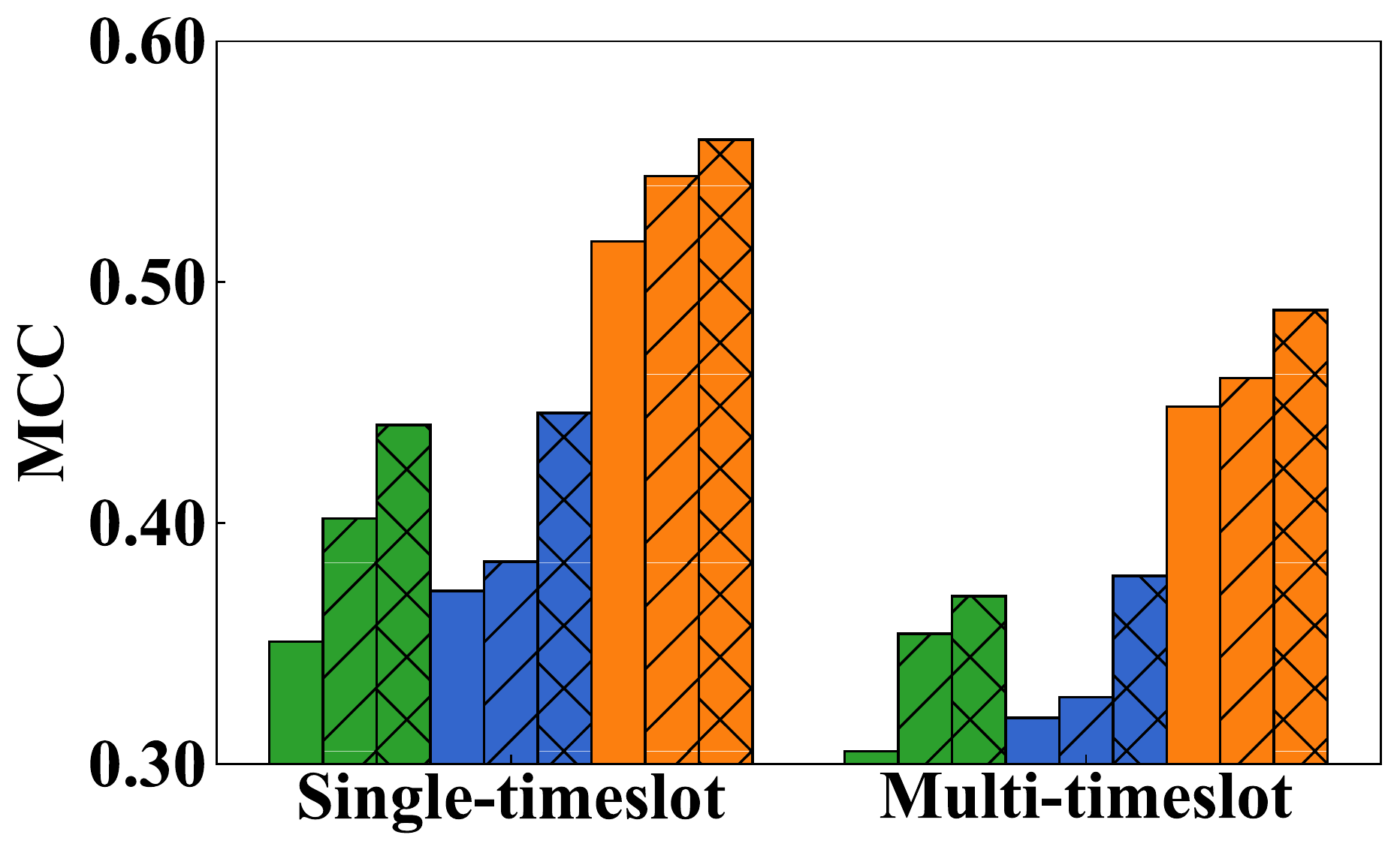}\label{rq2a}}
    \subfigure[Collaborative Good-mouthing Attack]{\includegraphics[scale=0.26]{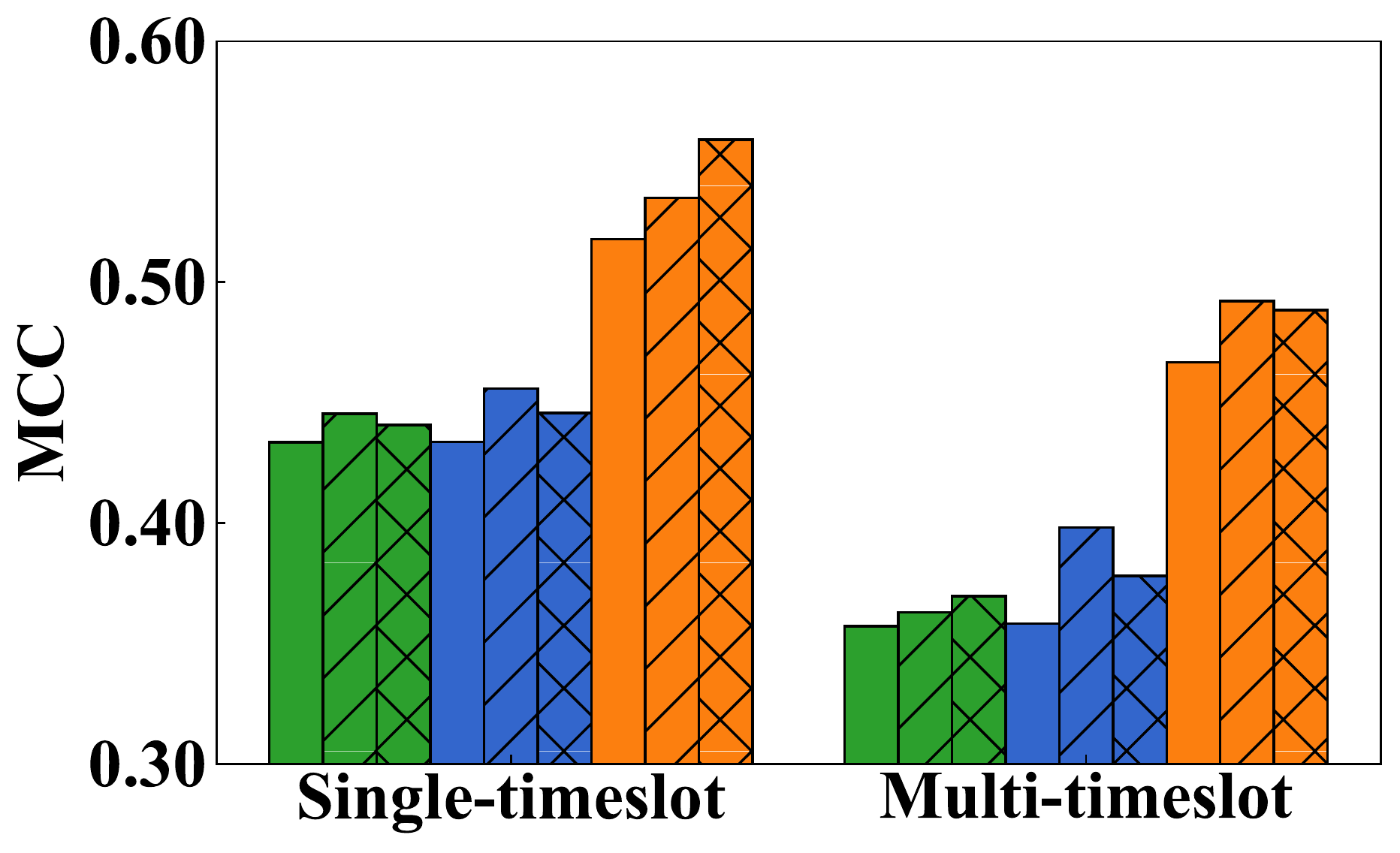}\label{rq2b}}
    \subfigure[On-off Attack]{\includegraphics[scale=0.26]{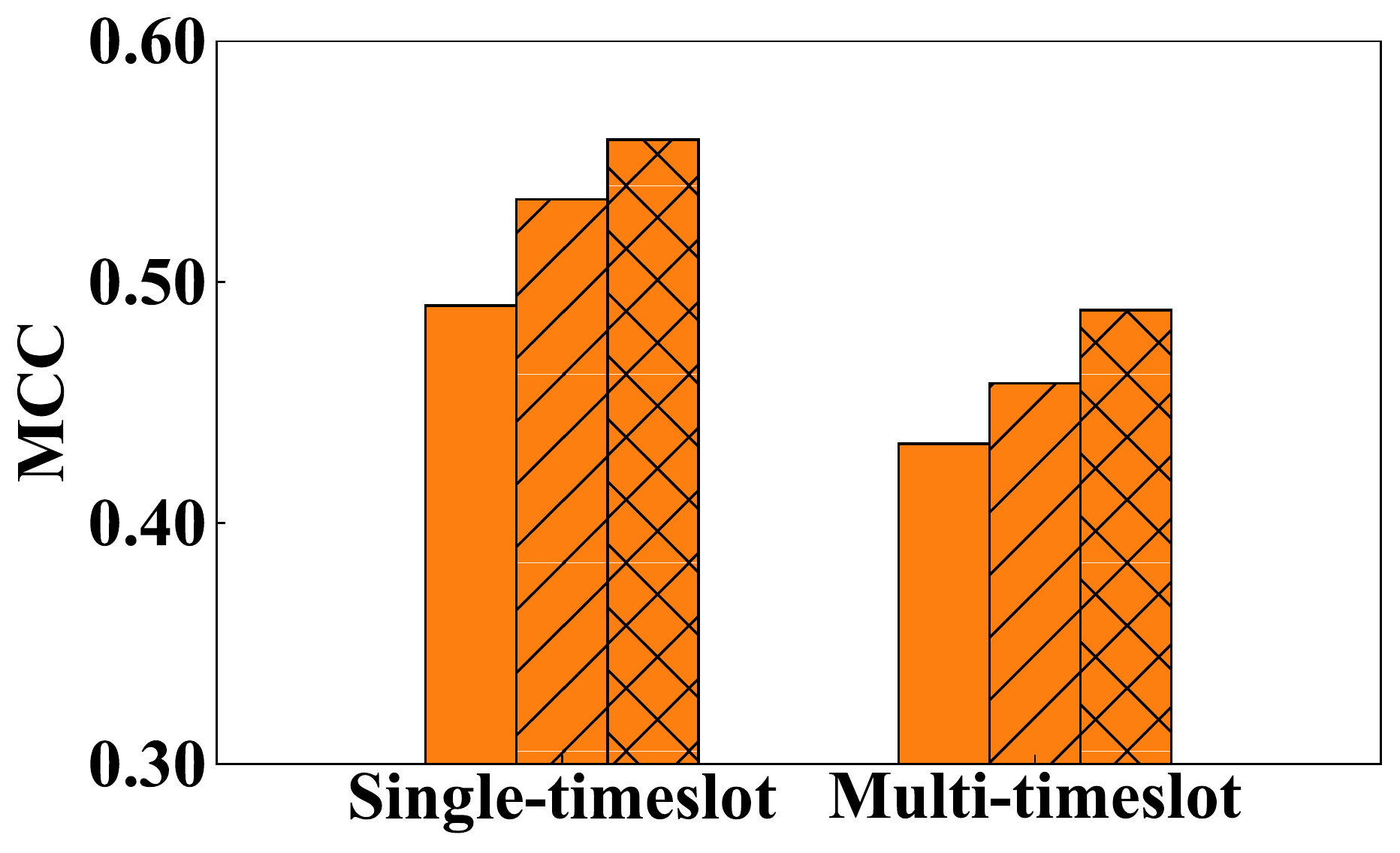}\label{rq2c}}
    
    \subfigure[Collaborative Bad-mouthing Attack]{\includegraphics[scale=0.26]{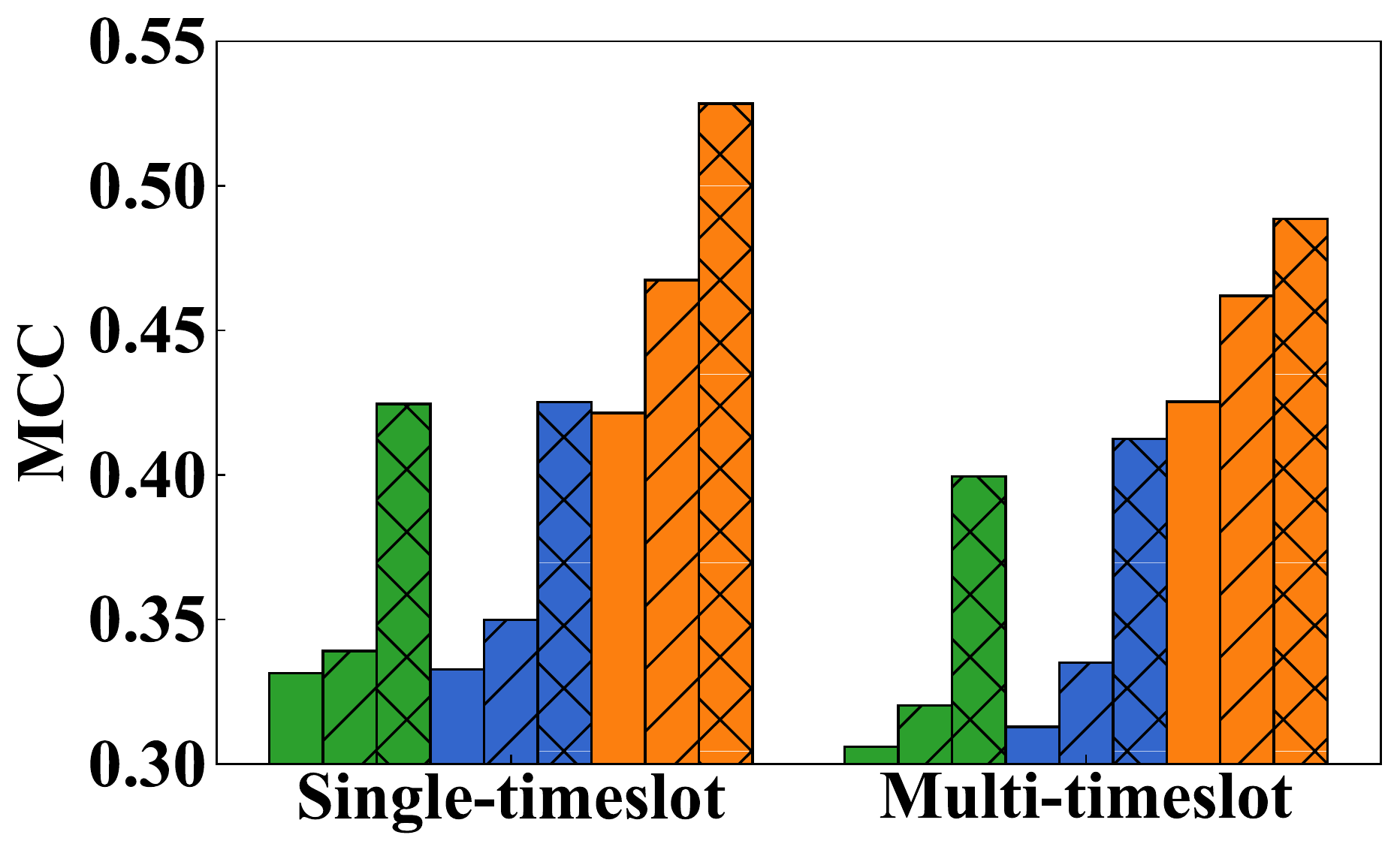}\label{rq2d}}
    \subfigure[Collaborative Good-mouthing Attack]{\includegraphics[scale=0.26]{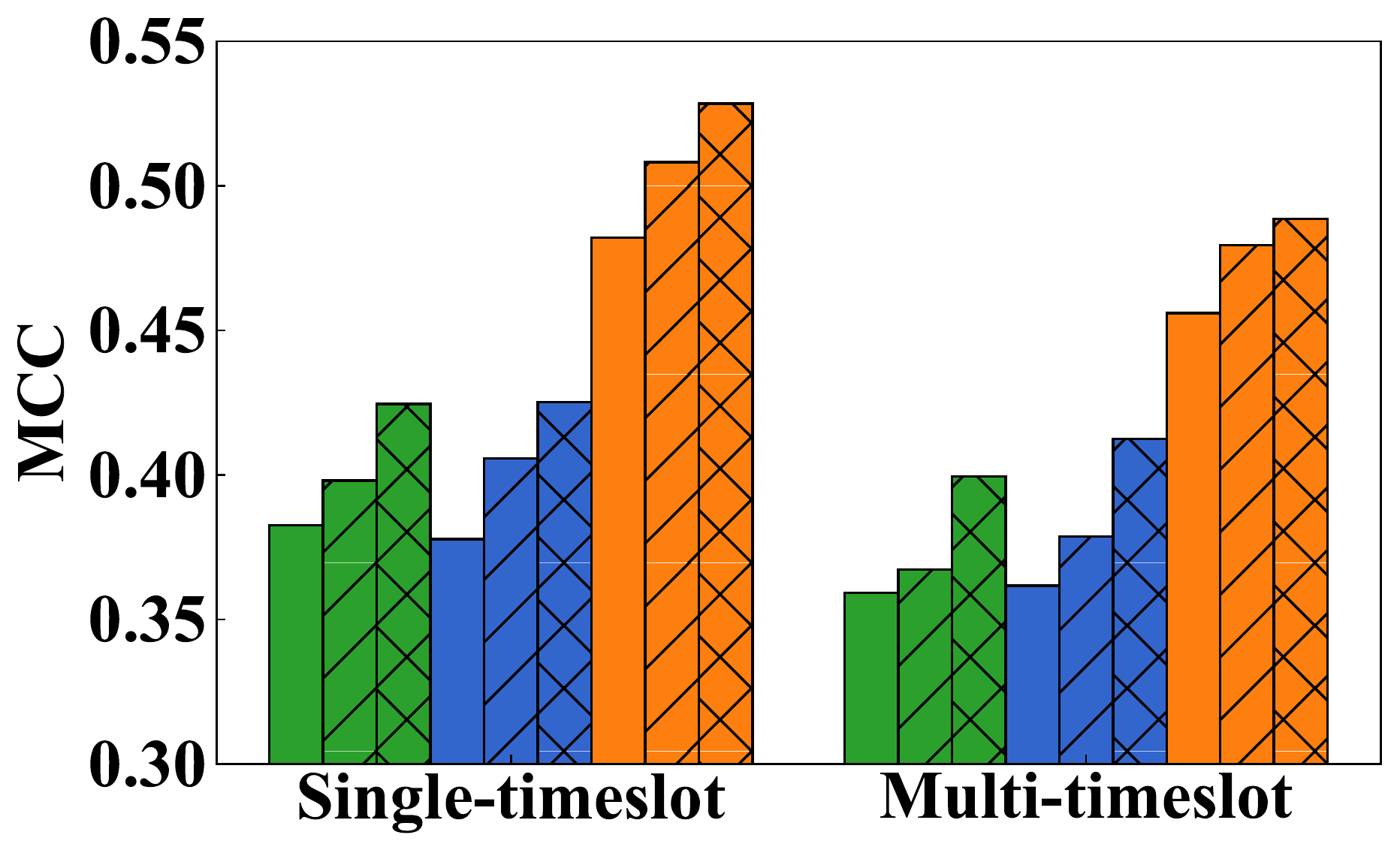}\label{rq2e}}
    \subfigure[On-off Attack]{\includegraphics[scale=0.26]{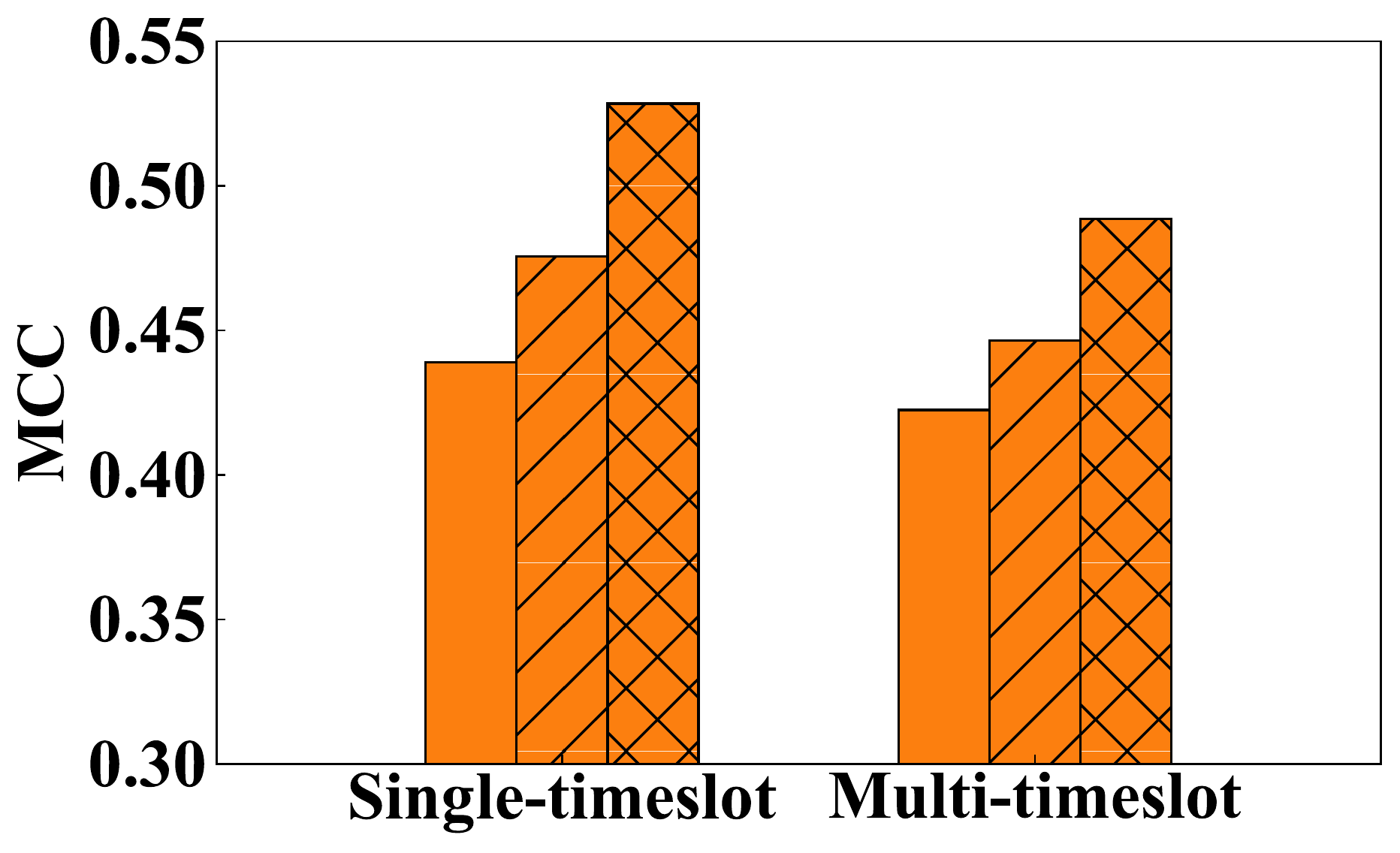}\label{rq2f}}

    \vspace{-2mm}
    \caption{Experimental results on the robustness of TrustGuard and baselines under different attacks. (a), (b), and (c) are the results based on Bitcoin-OTC. (d), (e), and (f) are the results based on Bitcoin-Alpha. De: Defense mechanism. TrustGuard is equipped with a mean aggregator in the ``attacks+w/o De" setting. On-off attack is only available in dynamic graph models, i.e., TrustGuard herein.}
    \label{with_attacks}
    \vspace{-2mm}
\end{figure*}

\subsection{Effectiveness of TrustGuard under Attacks (RQ2)} \label{rq2}
In this subsection, we investigate the robustness of TrustGuard and baselines against three common trust-related attacks, i.e., collaborative bad-mouthing, collaborative good-mouthing, and on-off attacks. As we have demonstrated the superiority of TrustGuard under different training ratios in the previous subsection, we maintain a fixed ratio of training data (i.e., the first 7 snapshots for training) to ensure the stability of results. For testing data, we use the 8th snapshot for testing in the single-timeslot tests, and use the last three snapshots for testing in the multi-timeslot tests. Additionally, we exclude experiments on the single-timeslot prediction on unobserved nodes since this task involves limited testing data and thus may introduce randomness.

We observe from Fig.~\ref{with_attacks} that: 
(i) The incorporation of our defense mechanism into all models offers significant improvements compared with the situation without it, which demonstrates the efficacy of the proposed defense mechanism against three typical attacks and its generality to any GNN-based models. For TrustGuard, the defense mechanism can improve the performance of TrustGuard without deploying it with 6.85\% on MCC against collaborative bad-mouthing attacks, 4.83\% on MCC against collaborative good-mouthing attacks, and 7.21\% on MCC against on-off attacks. 
(ii) Comparing Fig.~\ref{rq2a} with Fig.~\ref{rq2b}, and Fig.~\ref{rq2d} with Fig.~\ref{rq2e}, it is evident that collaborative bad-mouthing attacks are more disruptive and challenging to be defended against than collaborative good-mouthing attacks. The reason is that both datasets are extremely imbalanced, with a significant majority of trust edges, indicating that the nodes in these two graphs tend to trust each other. As a result, positive ratings are more common than negative ratings, reducing the impact of collaborative good-mouthing attacks when generating node embeddings. Furthermore, since trust is ``easy to lose, hard to gain"~\cite{yan2008trust,wang2020survey}, negative ratings have more impact on altering node embeddings than positive ones. 
(iii) Comparing TrustGuard without deploying the defense mechanism under attacks with other models without attacks, we find that TrustGuard in this situation still performs better than others. One possible reason is that the dynamic graph partitioning dilutes the impact of attacks. Specifically, when forming a node's embedding, the snapshot-based graph analysis in TrustGuard considers the node's behaviors at different timeslots with the help of an attention mechanism. Thus, an attack on that node at one or more timeslots may not significantly affect its final embedding. This evaluation manner can effectively counter on-off attacks (see observation (i)), where an on-off attacker behaves well and badly alternatively at different timeslots. 
(iv) Compared with other baseline models, TrustGuard generally exhibits a smaller decrease in performance when occurring attacks than the performance without suffering any attacks. Based on the observations (iii) and (iv), we can infer that the temporal attention layer in TrustGuard can also enhance its robustness.

\subsection{Explainability of TrustGuard (RQ3)}
In this subsection, we discuss the explainability of TrustGuard by visualizing the robust coefficients and attention scores learned by TrustGuard. 
Furthermore, we provide a toy example for visualizing trust evaluation results and present the results of a user study. 

\subsubsection{Robust Coefficients Visualization} 
\begin{figure}[t]
    \centering
    \subfigcapskip=-3pt
    \subfigure[Co-bad-mouthing Attack]{\includegraphics[scale=0.26]{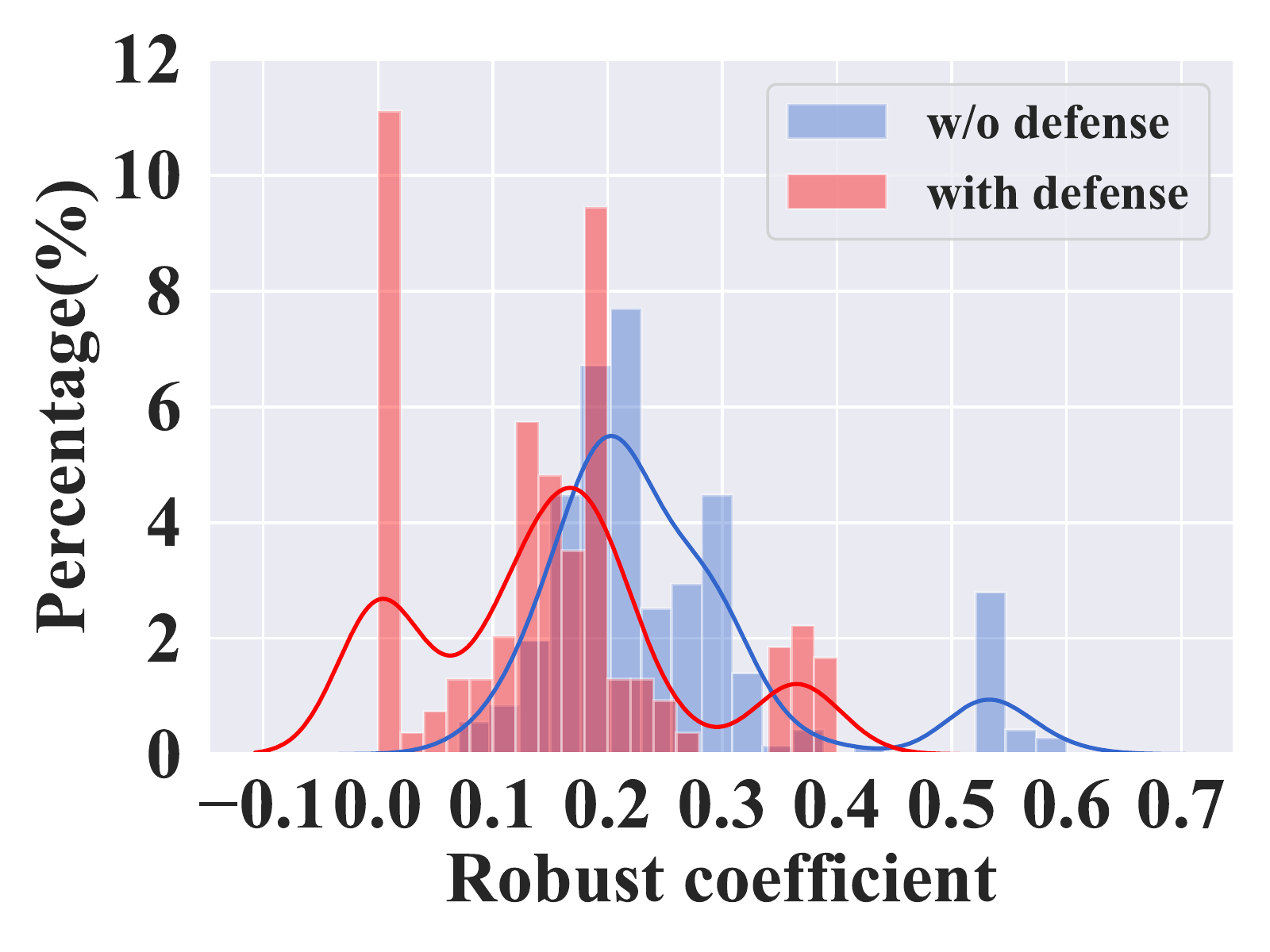}}
    \subfigure[Co-good-mouthing Attack]{\includegraphics[scale=0.26]{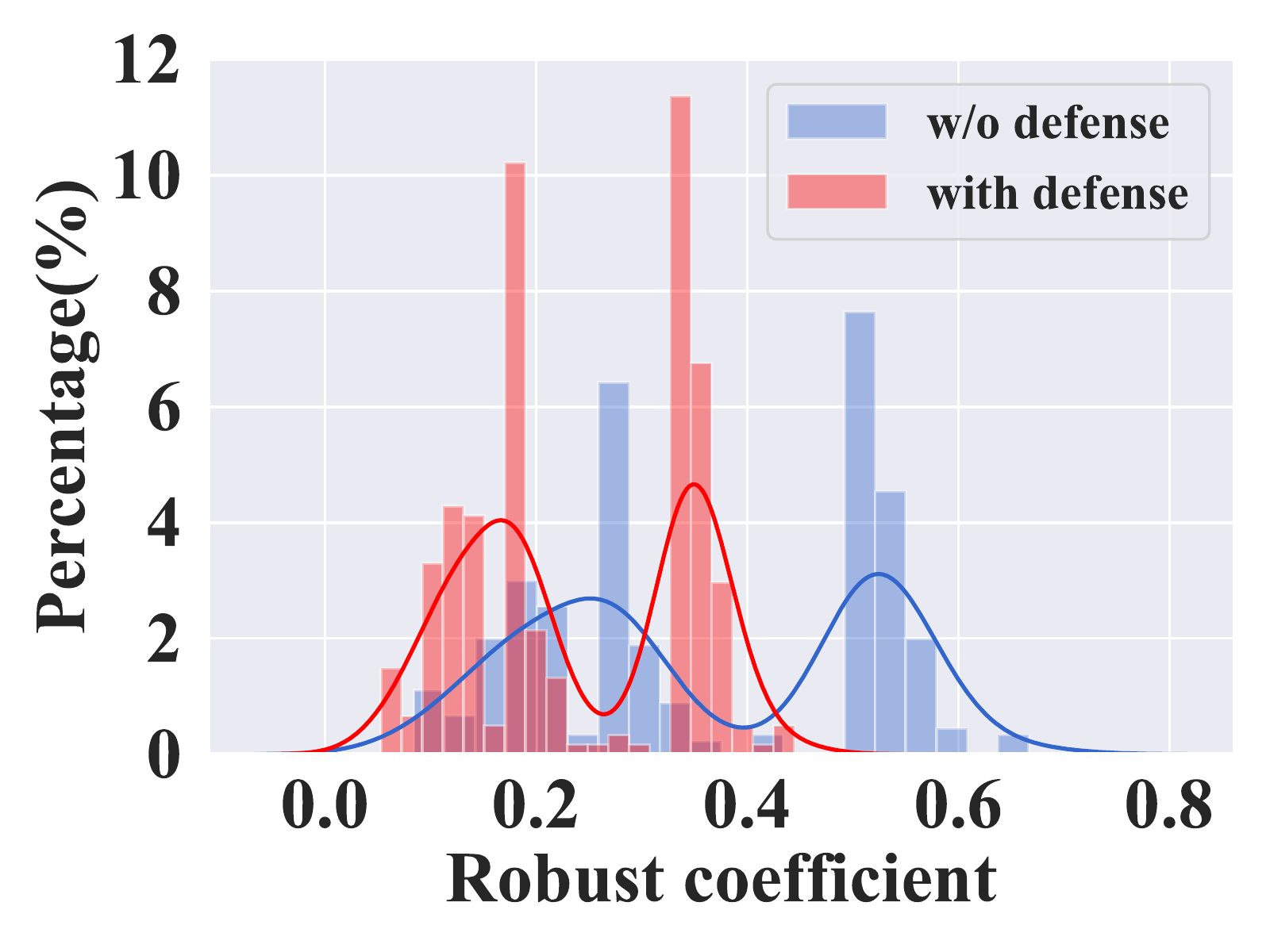}}

    \vspace{-2mm}
    \caption{Visualization of robust coefficients. The results are based on the single-timeslot prediction on observed nodes based on Bitcoin-OTC. Co-bad/Co-good-mouthing attack denotes a collaborative version herein.}
    \label{robust_coefficient}
    \vspace{-2mm}
\end{figure}

Robust coefficients indicate the importance of edges in trust propagation and aggregation. A robust trust evaluation model is expected to assign low coefficients to malicious edges. To assess whether TrustGuard does this, we plot the distributions of robust coefficients for malicious edges generated by collaborative bad-mouthing and collaborative good-mouthing attacks in Fig.~\ref{robust_coefficient}. We observe that malicious edges obtain high robust coefficients without the defense mechanism. This explains how these attacks disrupt the aggregation process of a GNN model. However, with the defense mechanism, we observe a significant leftward shift in the distributions of robust coefficients, indicating that lower coefficients are assigned to these edges. In addition, we calculate the mean values of robust coefficients of malicious edges generated by both attacks and find that they greatly decrease with the help of the defense mechanism. Specifically, the defense mechanism decreases the mean values by 40.96\% (0.249 $\to$ 0.147) when the collaborative bad-mouthing attack occurs and decreases the mean values by 33.43\% (0.362 $\to$ 0.241) when the collaborative good-mouthing attack happens. These observations suggest that the defense mechanism enhances the robustness of TrustGuard by effectively assigning low robust coefficients to malicious edges, which fits our expectations.

\subsubsection{Attention Scores Visualization} 
\begin{figure}[t]
    \centering
    \subfigcapskip=-3pt
    \subfigure{\includegraphics[scale=0.26]{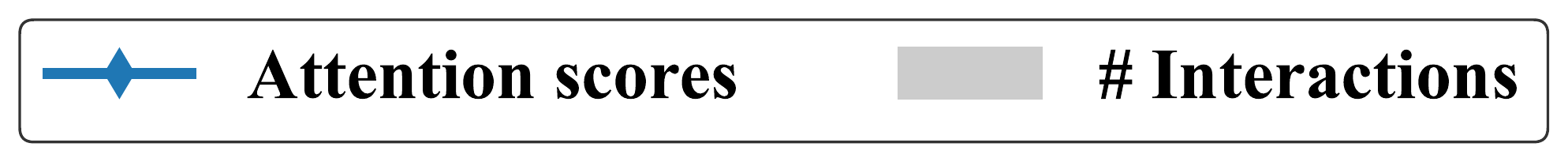}}
    \vspace{-3mm}

    \setcounter{subfigure}{0}
    \subfigure[Bitcoin-OTC]{\includegraphics[scale=0.25]{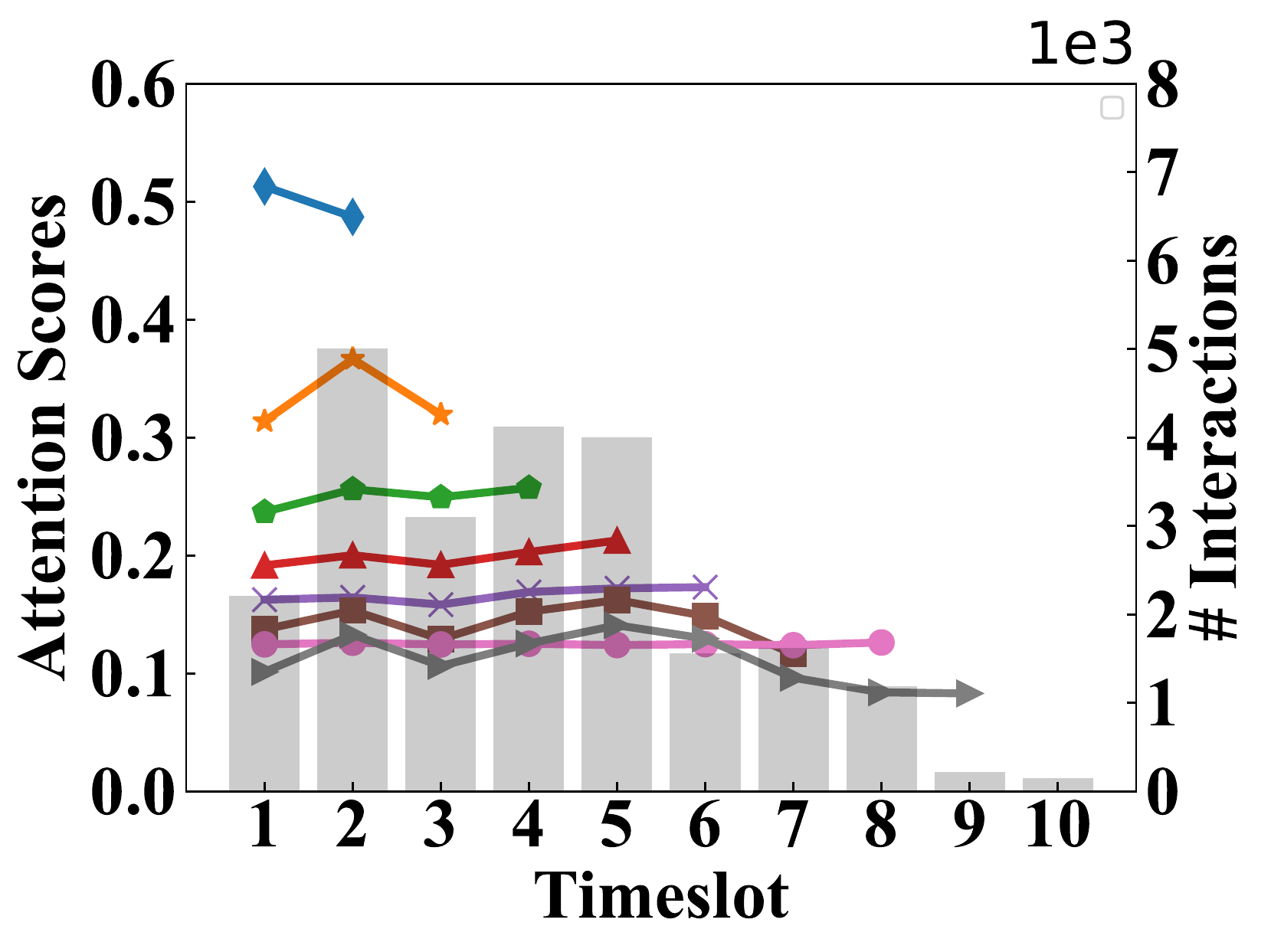}\label{attention_a}}
    \subfigure[Bitcoin-Alpha]{\includegraphics[scale=0.25]{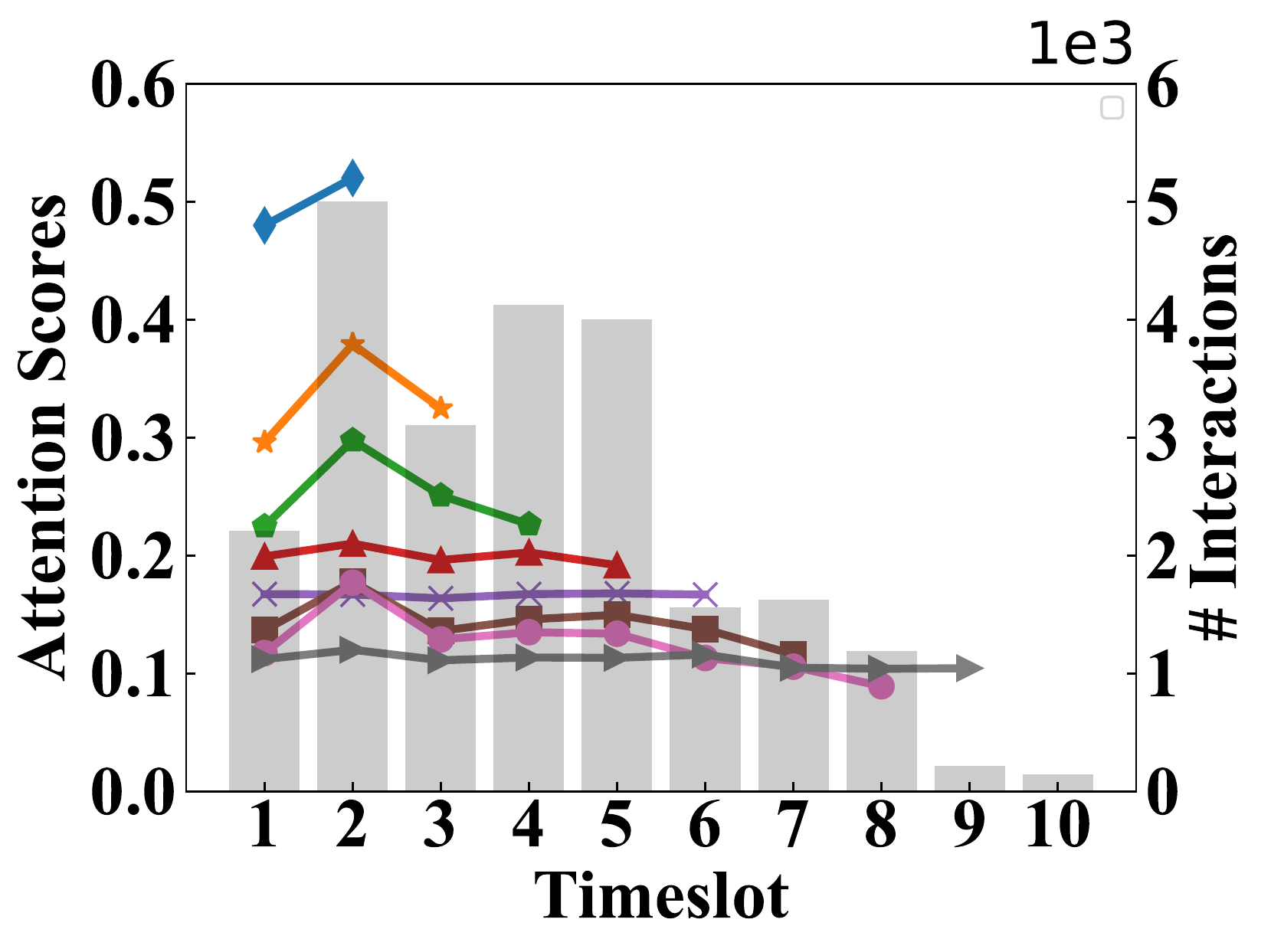}\label{attention_b}}

    \vspace{-2mm}
    \caption{Visualization of attention scores. The results are based on single-timeslot prediction on observed nodes.}
    \label{attention}
    \vspace{-2mm}
\end{figure}

As indicated in~\cite{min2021stgsn}, temporal patterns can be classified into five types, namely increasing trend, decreasing trend, seasonal, key points, and random. We visualize the distributions of attention scores along with trust interactions regarding both datasets, as 
shown in Fig.~\ref{attention}. Note that both Fig.~\ref{attention_a} and Fig.~\ref{attention_b} consist of 8 lines, representing the 8 subtasks in the single-timeslot prediction on observed nodes. 
From Fig.~\ref{attention}, we can see that the trend of attention scores generally aligns with the trend of the number of interactions. This implies that the temporal patterns learned by TrustGuard fall into the key point category, where certain snapshots exhibit greater influence than others. This observation is also in line with the nature of graph evolution in both datasets, where rating behaviors tend to be bursty and correlated with events. As a consequence, we conclude that TrustGuard is able to learn temporal patterns that accurately capture the graph evolution in real-world scenarios.

\begin{figure}[t]
    \centering
    \subfigcapskip=-3pt

    \setcounter{subfigure}{0}
    \subfigure[Toy Example]{\includegraphics[scale=0.38]{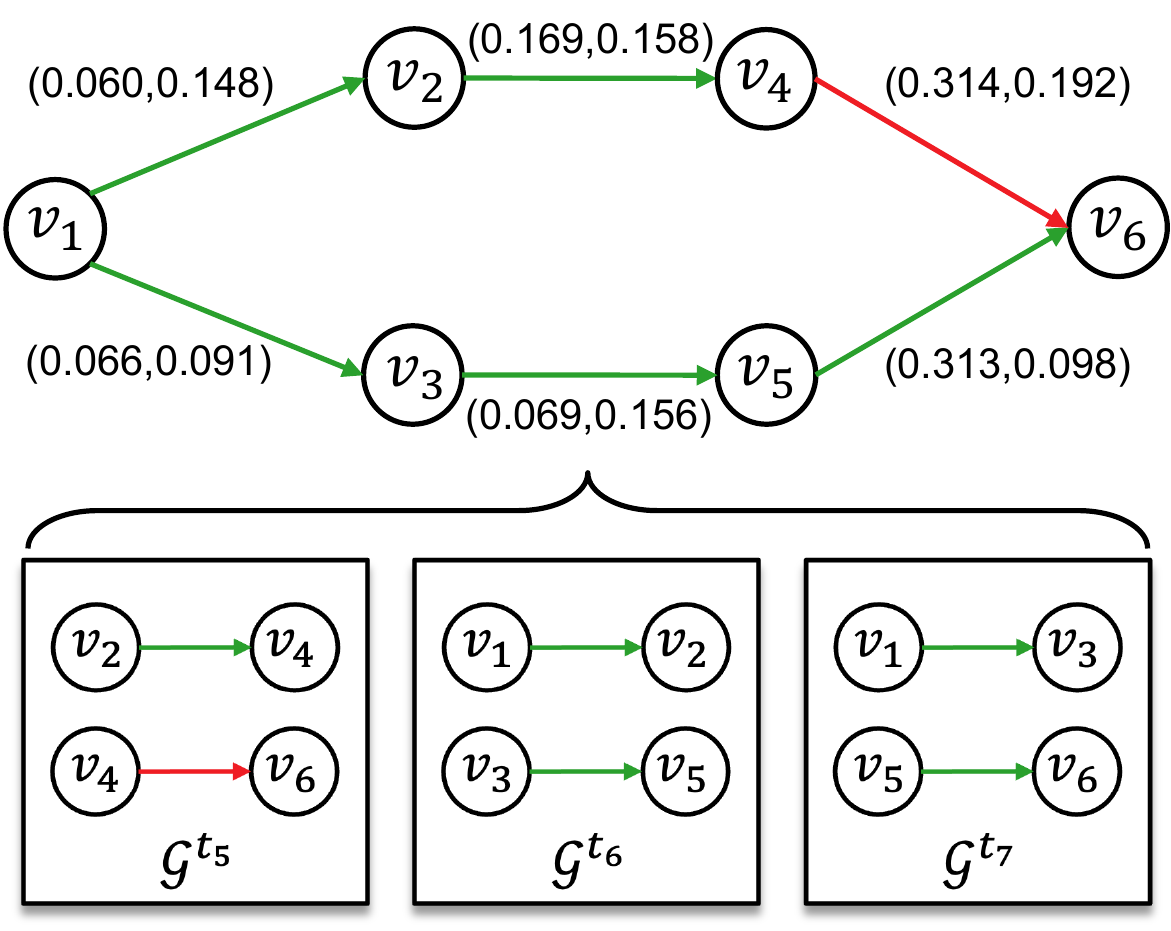}\label{user_a}}
    \subfigure[User Study]{\includegraphics[scale=0.25]{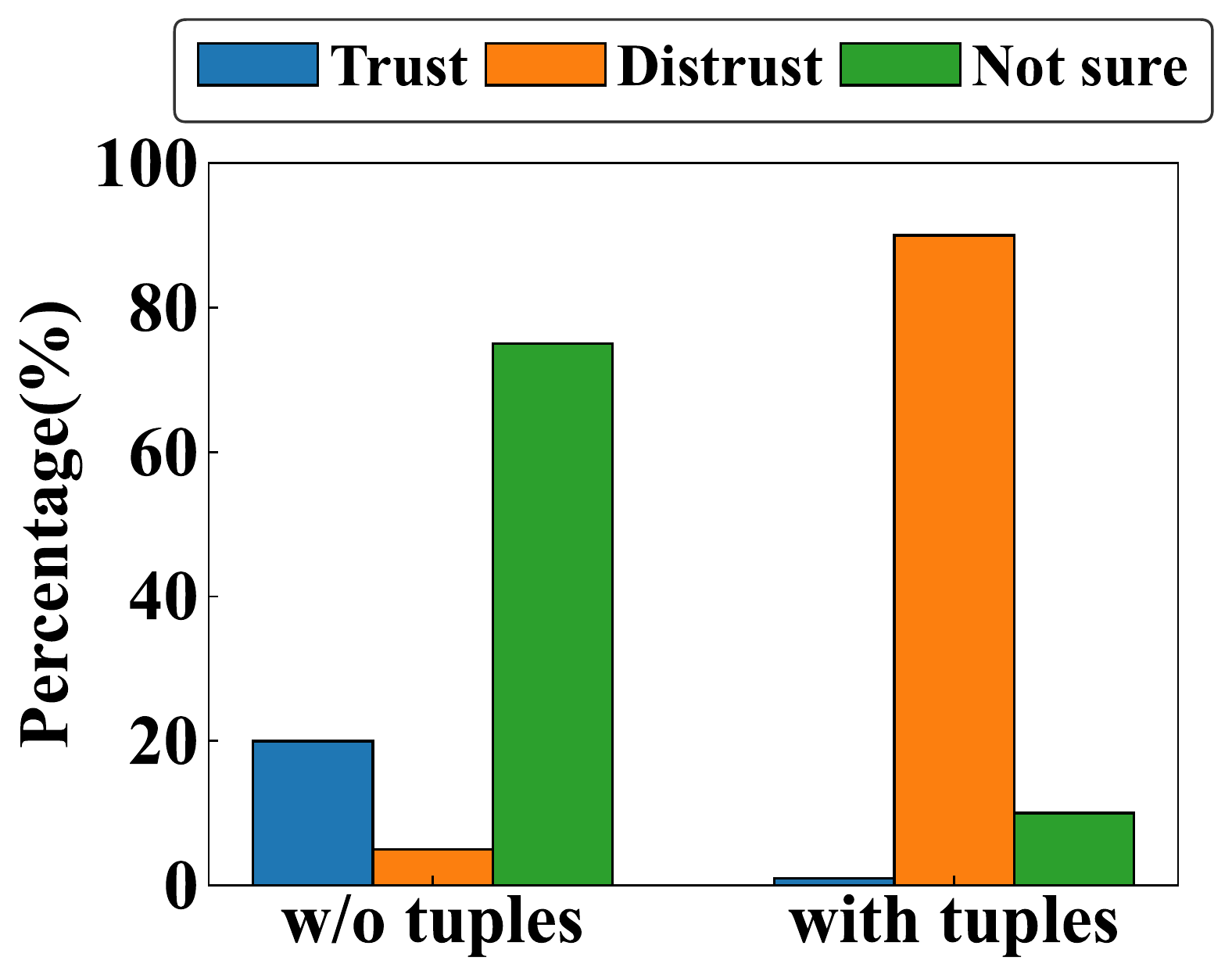}\label{user_b}}

    \vspace{-2mm}
    \caption{Toy example of $v_1$ distrusts $v_6$ at timeslot $t_8$ (a) and user study (b). In (a), green lines represent trusted relationships, while red lines represent distrusted relationships. Each tuple in (a) contains a robust coefficient representing spatial importance and an attention score representing temporal importance.}
    \label{user_study}
    \vspace{-2mm}
\end{figure}

\subsubsection{User Study} We present a toy example and conduct a user study to investigate whether the evaluation results derived from TrustGuard align with user expectations, as shown in Fig.~\ref{user_study}. To predict the trust relationship from a trustor to a trustee at timeslot $t_8$, it is natural to consider their interactions from $t_1$ to $t_7$ with the trust propagation length being 3, as specified in Section~\ref{section_rq5}. As a result, we can obtain a subgraph for each trustor-trustee pair, which allows us to observe the evolution of how the trustor establishes connections with the trustee, and know the importance of each interaction from corresponding tuples that contain robust coefficients and attention scores, such as the toy example in Fig.~\ref{user_a}. To examine whether the visualization of robust coefficients and attention scores can improve the explainability of TrustGuard from a human perspective, we conduct a user study. We first present Fig.~\ref{user_a} to 20 participants (see Appendix~\ref{background} for information on their background) in two cases, with and without tuples. Then, we ask them to determine whether $v_1$ trusts $v_6$ at timeslot $t_8$ in both cases. 
We find that 75\% participants are not sure about the trust relationship between $v_1$ and $v_6$ when the tuples are not presented. They argue that it is possible to infer that $v_1$ partially trusts $v_4$ and $v_5$ based on the conditionally transferability of trust. However, since $v_4$ holds a conflicting opinion with $v_5$ towards $v_6$, it becomes challenging to assess the trust relationship between $v_1$ and $v_6$. This situation changes when the tuples are presented. Specifically, 90\% participants argue that $v_1$ distrusts $v_6$ at timeslot $t_8$, which is the correct answer. 
Their reason is that the tuple values in $v_1 \to v_2 \to v_4 \to v_6$ are generally larger than those in $v_1 \to v_3 \to v_5 \to v_6$. Hence, $v_1$ is more likely to distrust $v_6$ at timeslot $t_8$. 

We further ask participants to rate their acceptance on the visualization with a scale from 1 to 5, i.e., from totally dissatisfied to totally satisfied. The average of these ratings is 3.83, indicating that the participants hold positive attitudes towards this visualization. All these findings suggest that the evaluation results of TrustGuard are user-acceptable and can be explained by considering the interactions and their spatial and temporal importance within a trustor-trustee subgraph. However, it is worth noting that the granularity of explainability should be further improved by providing additional details about the inner mechanism of TrustGuard through a specific explanation method, such as GNNExplainer~\cite{ying2019gnnexplainer}, which we leave as our future work.

\subsection{Ablation Study (RQ4)}
In this subsection, we investigate the contribution of each component in the spatial and temporal aggregation layers of TrustGuard.

\begin{table*}[tbp]
	\centering 
	\footnotesize
	\caption{Ablation study on trustor \& trustee information representation and learning temporal patterns.}
        \label{ablation_study}
    {\vspace{-2mm}In each column, the best result is highlighted in \textbf{bold}. Improvement is calculated relative to the best baseline. \ding{172} denotes the single-timeslot prediction on observed nodes. \ding{173} denotes the multi-timeslot prediction on observed nodes.}
    \\[2mm] 

	\begin{tabular}{@{}c|c|cccc|cccc@{}}  
		\toprule[1.5pt]
    \multirow{2.5}{*}{\makebox[0.05\textwidth][c]{Task}} &\multirow{2.5}{*}{Setting} &\multicolumn{4}{c|}{Bitcoin-OTC} &\multicolumn{4}{c}{Bitcoin-Alpha}\\
    \cmidrule{3-10}
		& &MCC &AUC &BA &F1-macro &MCC &AUC &BA &F1-macro \\
   
   \midrule
		  \multirow{6}{*}{\ding{172}} & TrustorGuard &0.265$\pm$0.006 &0.720$\pm$0.006 &  0.668$\pm$0.007 & 0.620$\pm$0.003 &0.241$\pm$0.006 &0.708$\pm$0.004& 0.661$\pm$0.004 & 0.590$\pm$0.006 \\
		& TrusteeGuard &0.374$\pm$0.013 &0.731$\pm$0.008 &0.661$\pm$0.008 &0.674$\pm$0.004 &0.341$\pm$0.016 &0.690$\pm$0.011& 0.649$\pm$0.009 &0.655$\pm$0.005 \\
  \cmidrule{2-10}
	& GuardMean &0.357$\pm$0.010 &0.745$\pm$0.004 & 0.674$\pm$0.005 & 0.673$\pm$0.004 &0.341$\pm$0.017 &0.707$\pm$0.008& 0.668$\pm$0.006 & 0.654$\pm$0.005 \\
		& GuardDecay &0.370$\pm$0.006 &0.753$\pm$0.005 &0.679$\pm$0.003 &0.675$\pm$0.004 &0.346$\pm$0.009 &0.719$\pm$0.006& 0.670$\pm$0.005 &0.656$\pm$0.007 \\	
  \cmidrule{2-10}
        & \textbf{TrustGuard} & \textbf{0.389$\pm$0.007} &\textbf{0.765$\pm$0.008} & \textbf{0.693$\pm$0.008} &\textbf{0.687$\pm$0.007} &\textbf{0.362$\pm$0.004}&\textbf{0.756$\pm$0.009}& \textbf{0.692$\pm$0.004} &\textbf{0.669$\pm$0.002} \\
    \midrule

		  \multirow{6}{*}{\ding{173}} & TrustorGuard &0.221$\pm$0.005 &0.683$\pm$0.003 &  0.629$\pm$0.003 & 0.604$\pm$0.003 &0.197$\pm$0.013 &0.660$\pm$0.003 & 0.618$\pm$0.008 & 0.578$\pm$0.009 \\
		& TrusteeGuard &0.325$\pm$0.012 &0.694$\pm$0.005 &0.626$\pm$0.003 &0.650$\pm$0.002 &0.274$\pm$0.011 &0.672$\pm$0.015 &0.605$\pm$0.005 &0.622$\pm$0.002 \\
  \cmidrule{2-10}
            & GuardMean &0.298$\pm$0.002 &0.701$\pm$0.003 &  0.635$\pm$0.005 & 0.645$\pm$0.001 &0.252$\pm$0.010 &0.662$\pm$0.010& 0.612$\pm$0.009 & 0.618$\pm$0.004 \\
		& GuardDecay &0.302$\pm$0.003 &0.708$\pm$0.003 &0.638$\pm$0.004 &0.647$\pm$0.001 &0.254$\pm$0.010 &0.667$\pm$0.007& 0.614$\pm$0.005 &0.620$\pm$0.003 \\
  \cmidrule{2-10}
		& \textbf{TrustGuard} & \textbf{0.330$\pm$0.005} &\textbf{0.725$\pm$0.004} & \textbf{0.642$\pm$0.003} &\textbf{0.658$\pm$0.003} &\textbf{0.288$\pm$0.002}&\textbf{0.692$\pm$0.003}& \textbf{0.632$\pm$0.006} &\textbf{0.639$\pm$0.001} \\

	\bottomrule[1.5pt]
	\end{tabular}
	\vspace{-2mm}
\end{table*}

\subsubsection{Trustor \& Trustee Information Representation} To analyze the necessity of covering both trustor and trustee roles in node embeddings, we construct two variants of TrustGuard, i.e., TrustorGuard and TrusteeGuard. In the TrustorGuard variant, spatial aggregation is performed by only taking the out-degree neighbors of a node into account, which corresponds to its role as a trustor. In the TrusteeGuard variant, spatial aggregation is carried out by only taking the in-degree neighbors of a node into account, which corresponds to its role as a trustee. Based on the results with standard deviations in Table~\ref{ablation_study}, we can conclude that TrustGuard consistently outperforms those variants that only consider one role of a node. The reason is that incorporating both roles provides holistic information, which helps in generating informative and representative node embeddings. Furthermore, trust is influenced by many factors, such as trustors' properties and trustees' properties, which can be learned by modeling both roles. We also observe that the TrustorGuard variant performs significantly poorly. This implies that opinions from others are more conducive in shaping one's embedding than its own. Overall, these results support the significance of incorporating both trustor and trustee roles when forming a node's spatial embedding.

\subsubsection{Learning Temporal Patterns} To analyze the necessity of learning temporal patterns, we construct two variants of TrustGuard, including GuardMean and GuardDecay. In the GuardMean variant, each timeslot is given the same weight in the temporal aggregation layer. In the GuardDecay variant, a time decay factor $e^{(t_i - t_n)/ \tau}$ is introduced to implement a temporal pattern that exhibits the increasing trend. $t_i$ denotes a current timeslot, $t_n$ denotes the target timeslot, and $\tau$ is an adjusting factor. The results with standard deviations presented in Table~\ref{ablation_study} show that TrustGuard is superior to these two variants. This suggests that the attention mechanism adopted in TrustGuard has a notable advantage over manually configured temporal patterns. We also observe a slight advantage of the GuardDecay variant over the GuardMean variant, indicating the importance of capturing temporal dynamics. To summarize, the attention mechanism exhibits a huge potential on generality, as it can automatically extract latent temporal patterns from a sequence of time-ordered graphs.

\subsection{Parameter Sensitivity Analysis (RQ5)} \label{section_rq5}
In this subsection, we explore the sensitivity of TrustGuard to trust propagation length, number of attention heads, and pruning threshold.

\subsubsection{Trust Propagation Length} 
\begin{figure}[t]
    \centering
    \subfigcapskip=-3pt
    \subfigure{\includegraphics[scale=0.17]{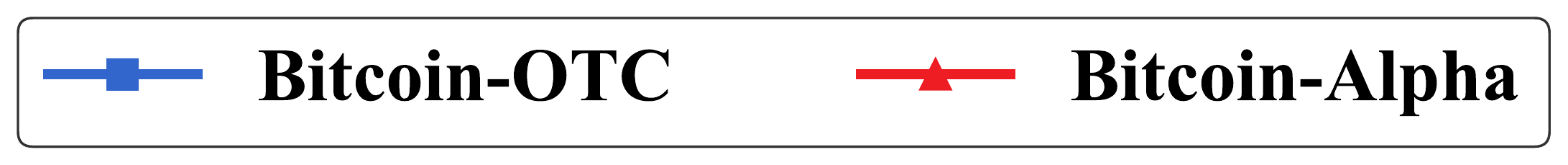}}
    \vspace{-3mm}

    \setcounter{subfigure}{0}
    \subfigure[]{\includegraphics[scale=0.17]{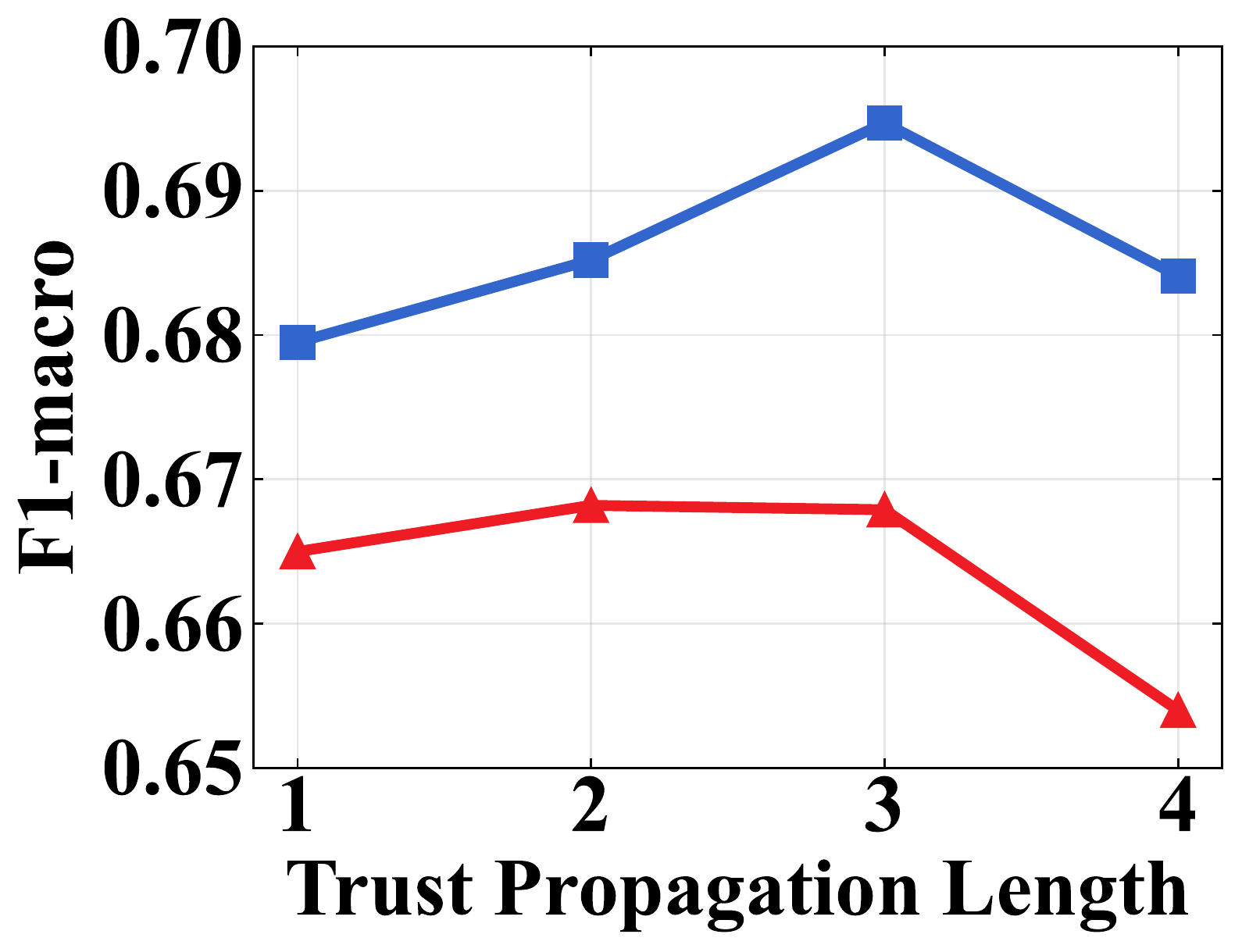}\label{parameter_a}}
    \subfigure[]{\includegraphics[scale=0.17]{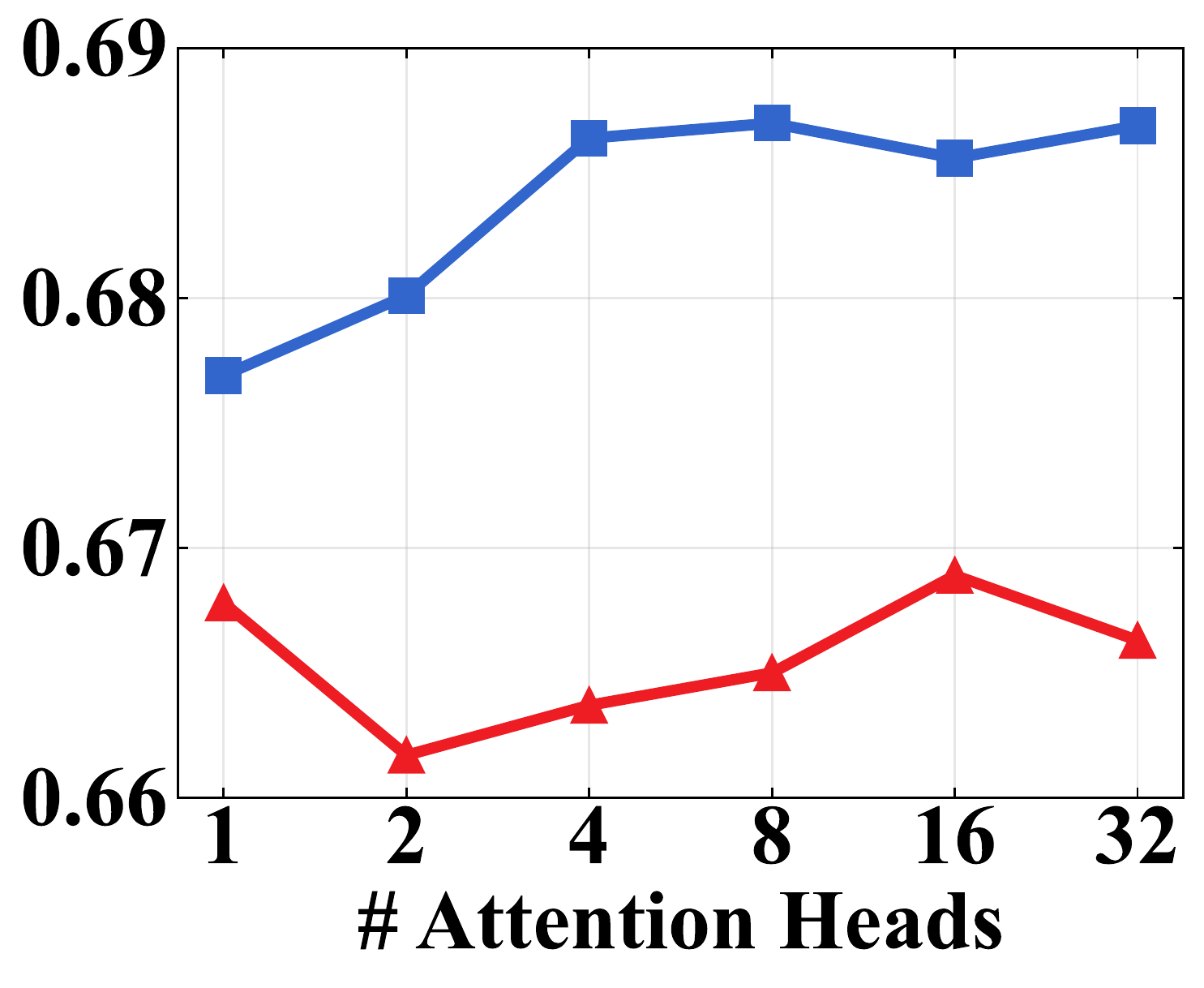}\label{parameter_b}}
    \subfigure[]{\includegraphics[scale=0.17]{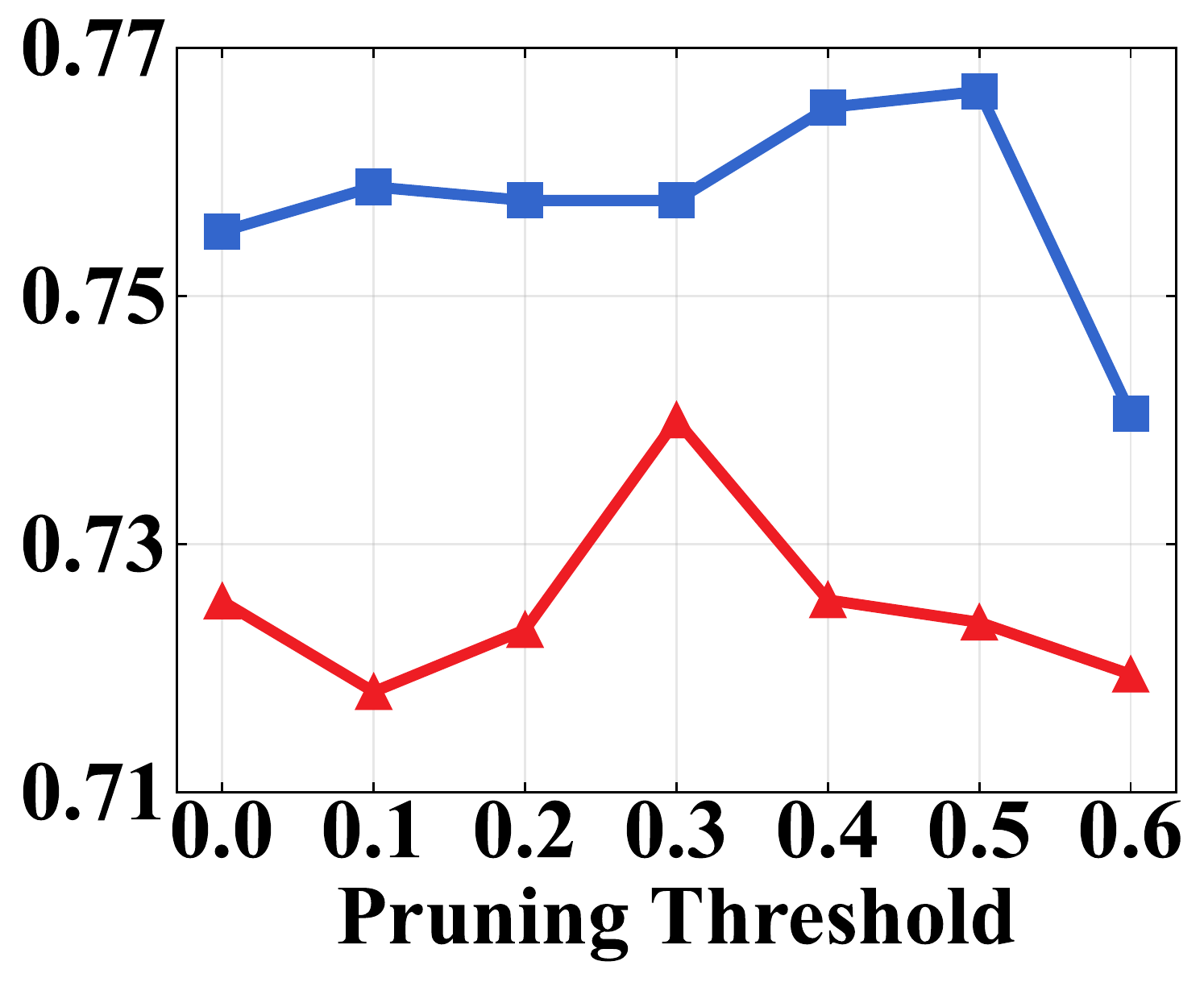}\label{parameter_c}}

    \vspace{-2mm}
    \caption{Sensitivity analysis for trust propagation length, number of attention heads, and pruning threshold. The results are based on the single-timeslot prediction on observed nodes.}
    \label{parameter}
    \vspace{-2mm}
\end{figure}

Trust propagation length refers to the number of orders of neighbors that need to be considered when generating a node's embedding. As shown in Fig.~\ref{parameter_a}, the performance of TrustGuard improves as the trust propagation length increases. However, excessively increasing the trust propagation length leads to performance decline. This observation is reasonable because the conditionally transferable nature of trust cannot be adequately captured if the trust propagation length is too short, while if the length is too long, excessive noise and redundancy could be introduced since trust may decrease during propagation. Based on the results and analysis, we set the default trust propagation length as 3.

\subsubsection{Number of Attention Heads} Based on the results in Fig.~\ref{parameter_b}, we can infer that increasing the number of attention heads generally leads to performance improvement. This could be attributed to the ability of multiple attention heads to learn temporal patterns from different aspects. In addition, we find that TrustGuard with the single-head configuration performs quite well on the Bitcoin-Alpha dataset. This may be caused by the small size of this dataset, where a single head is sufficient to learn the temporal patterns. Additionally, too many attention heads may lead to over-fitting, which could also explain the performance decline of configuring 32 attention heads on the Bitcoin-Alpha dataset. Based on the experimental results, we set 8 and 16 attention heads for Bitcoin-OTC and Bitcoin-Alpha, respectively.

\subsubsection{Pruning Threshold} As can be seen from Fig.~\ref{parameter_c}, there exists an optimal pruning threshold for each dataset. If the pruning threshold is set too small, it would have a negligible impact on the accuracy of TrustGuard, as only a limited number of malicious edges are filtered out. Conversely, if the pruning threshold is set too large, it would adversely affect the performance of TrustGuard. The reason is that not only the malicious edges are pruned, but also some normal edges are affected. Based on the experimental results, we set the threshold values as 0.5 and 0.3 for Bitcoin-OTC and Bitcoin-Alpha, respectively.

\section{Further Discussion} \label{section6}
\subsection{Scalability Analysis}
In this subsection, we discuss the scalability of TrustGuard by analyzing its time complexity, investigating the impact of the number of historical snapshots on TrustGuard’s running time, and examining its applicability across various network types.

\subsubsection{Time Complexity}
We analyze the time complexity of the spatial aggregation layer and the temporal aggregation layer since they are dominant in TrustGuard. Let $\left | \mathcal{V} \right |$ and $\left | \mathcal{E} \right |$ denote the number of nodes and edges, respectively, in the dynamic graph $\mathcal{G}(T)$, $\left | \mathcal{V}^{t_i} \right |$ the number of nodes in the snapshot $\mathcal{G}^{t_i}$, $\left | \mathcal{W} \right |$ the number of trust levels, $d_e$ the dimension of node embeddings, $L$ the number of trust propagation layers, and $n$ the number of snapshots. The time complexity of the spatial aggregation layer is $O(L \left | \mathcal{E} \right | \left | \mathcal{W} \right | d_e + L (\sum_{i=1}^{n} \left | \mathcal{V}^{t_i} \right |) d_{e}^{2})$, with $\sum_{i=1}^{n} \left | \mathcal{V}^{t_i} \right | < n \left | \mathcal{V} \right |$. The temporal aggregation layer has a time complexity of $O(n \left | \mathcal{V} \right | d_{e}^{2})$, indicating a linear increase in operational efficiency as $n$ grows. Although the impact of $n$ on the spatial aggregation layer is indirect, it is not negligible.
In summary, the overall time complexity of TrustGuard can be expressed as $O(L \left | \mathcal{E} \right | \left | \mathcal{W} \right | d_e + L (\sum_{i=1}^{n} \left | \mathcal{V}^{t_i} \right |) d_{e}^{2} + n \left | \mathcal{V} \right | d_{e}^{2})$.

\subsubsection{Impact of Historical Snapshot Quantity on Running Time}
\begin{figure}[t]
    \centering
    \subfigcapskip=-3pt
    \hspace{4mm}
    \subfigure{\includegraphics[scale=0.24]{legend.pdf}}
    \vspace{-3mm}

    \setcounter{subfigure}{0}
    \subfigure[Bitcoin-OTC]{\includegraphics[scale=0.24]{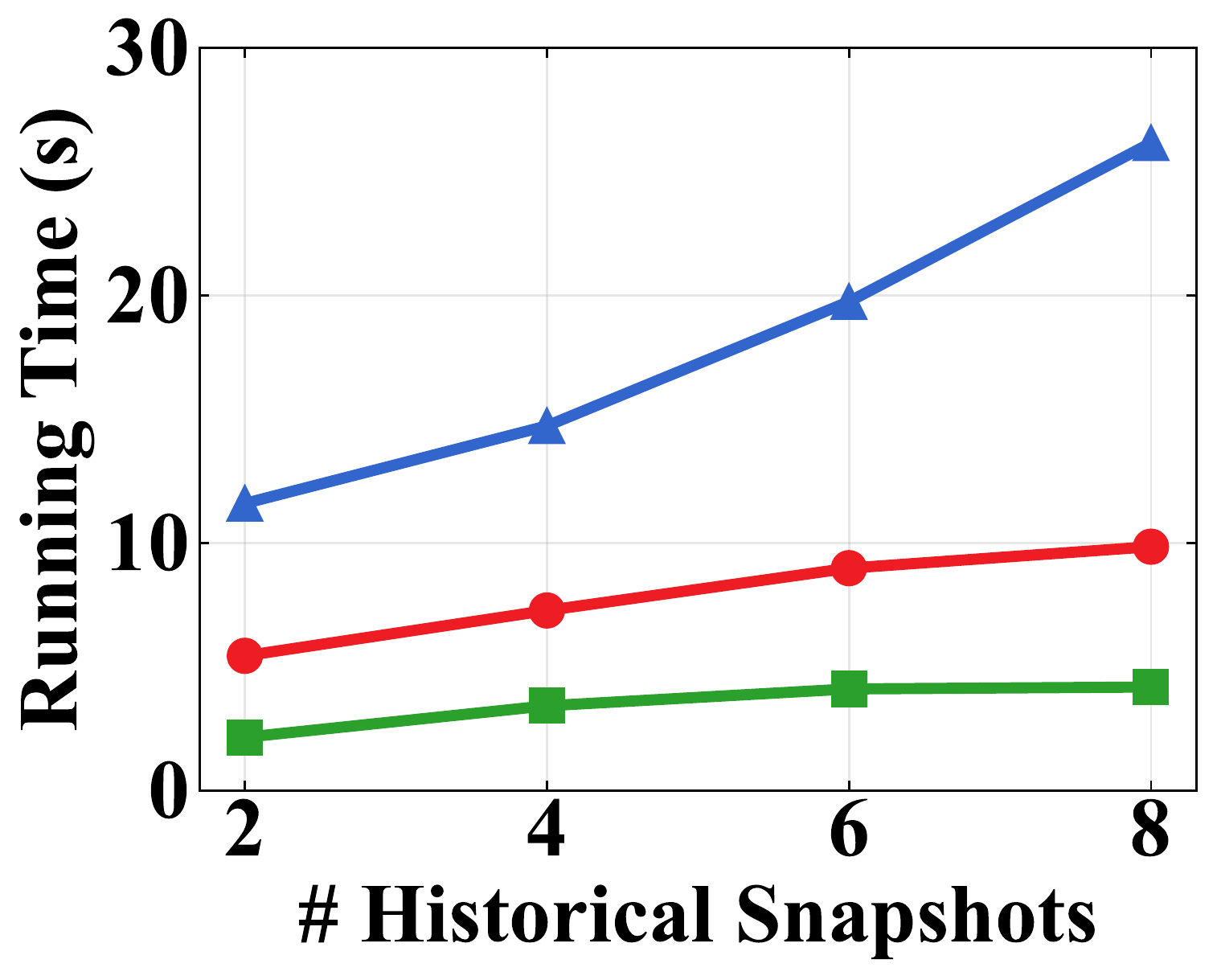}}
    \subfigure[Bitcoin-Alpha]{\includegraphics[scale=0.24]{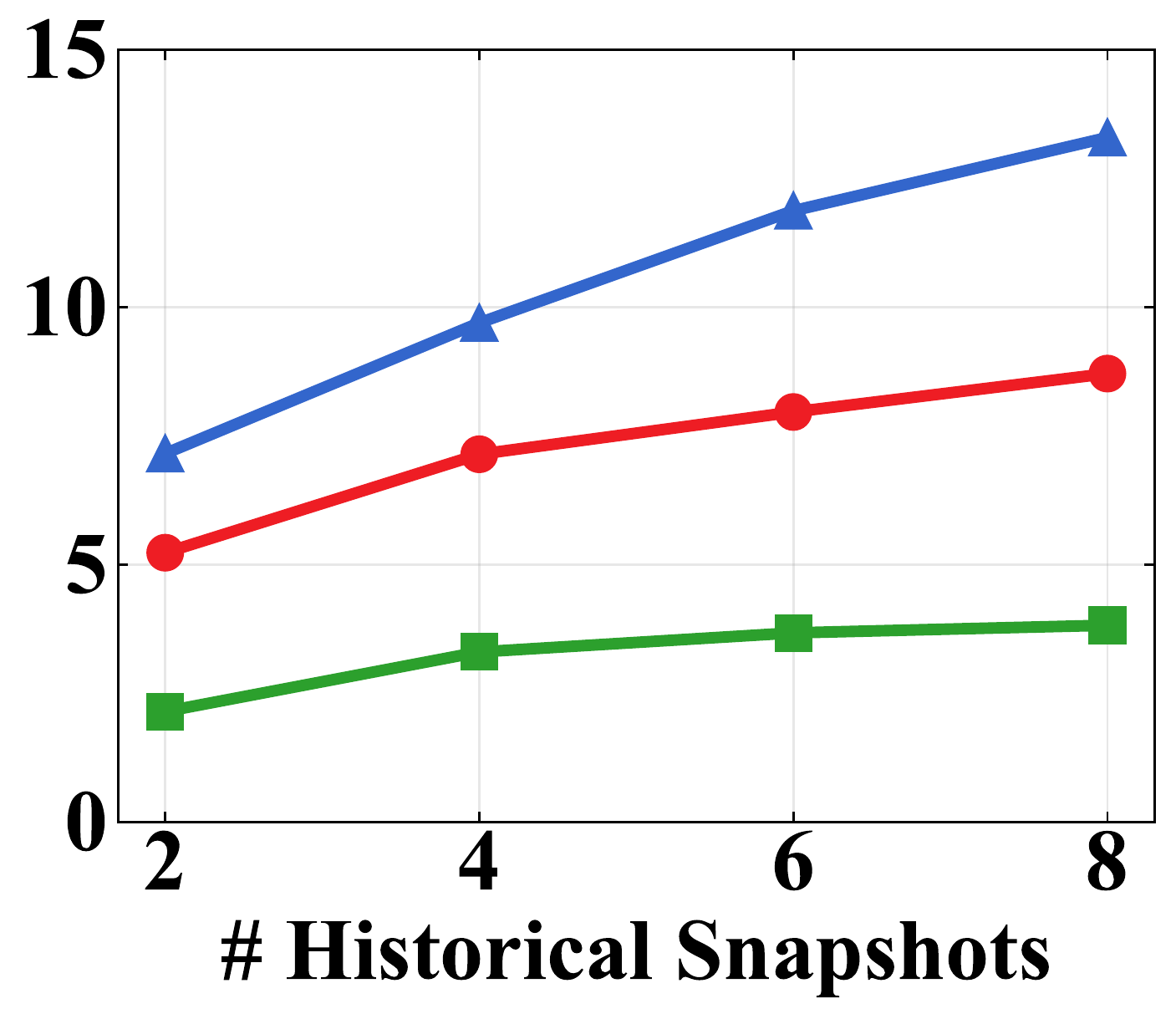}}

    \vspace{-2mm}
    \caption{Running time of TrustGuard and baselines on Bitcoin-OTC and Bitcoin-Alpha datasets.}
    \label{running time}
    \vspace{-2mm}
\end{figure}

We evaluate the running time of TrustGuard and compare it with the baselines by varying the number of historical snapshots utilized for training. Fig.~\ref{running time} shows that as the number of historical snapshots increases, the running time of all models also increases due to the growth of graph size. Guardian is the most efficient model owing to its usage of a straightforward average operation for message aggregation, in contrast to GATrust and TrustGuard, which employ more complex attention mechanisms. Notably, TrustGuard runs faster than GATrust, primarily because it only calculates one type of attention score, whereas GATrust calculates three distinct types. Furthermore, unlike the other two models, TrustGuard is dynamic in nature. With the increase of historical snapshots, its running time shows a modest increment, suggesting that TrustGuard is scalable to large dynamic datasets.

\subsubsection{Applicability to Various Network Types}
\begin{figure}[t]
    \centering
    \subfigcapskip=-3pt
    \hspace{5mm}
    \subfigure{\includegraphics[scale=0.24]{legend.pdf}}
    \vspace{-3mm}

    \setcounter{subfigure}{0}
    \subfigure[Advogato]{\includegraphics[scale=0.24]{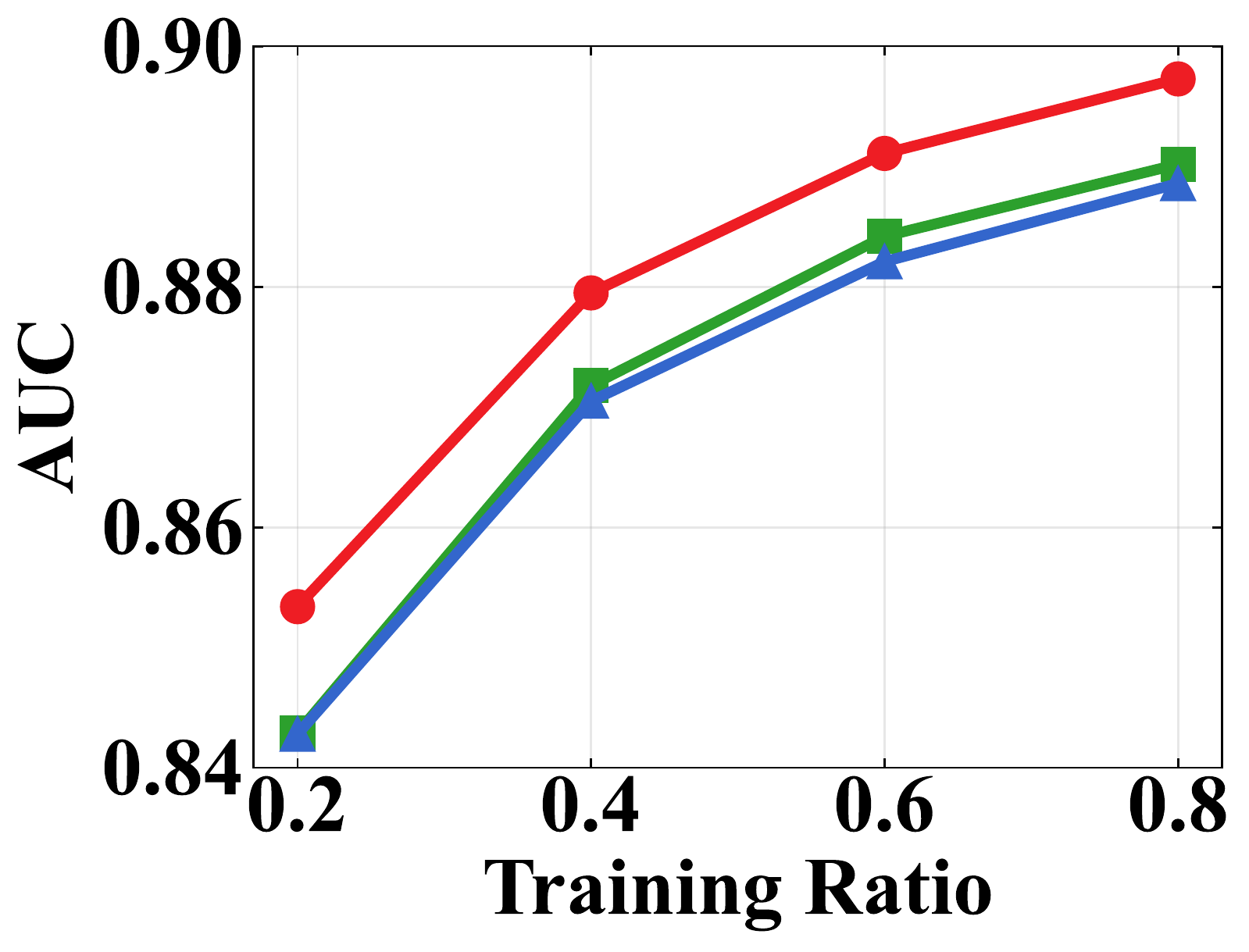}}
    \subfigure[PGP]{\includegraphics[scale=0.24]{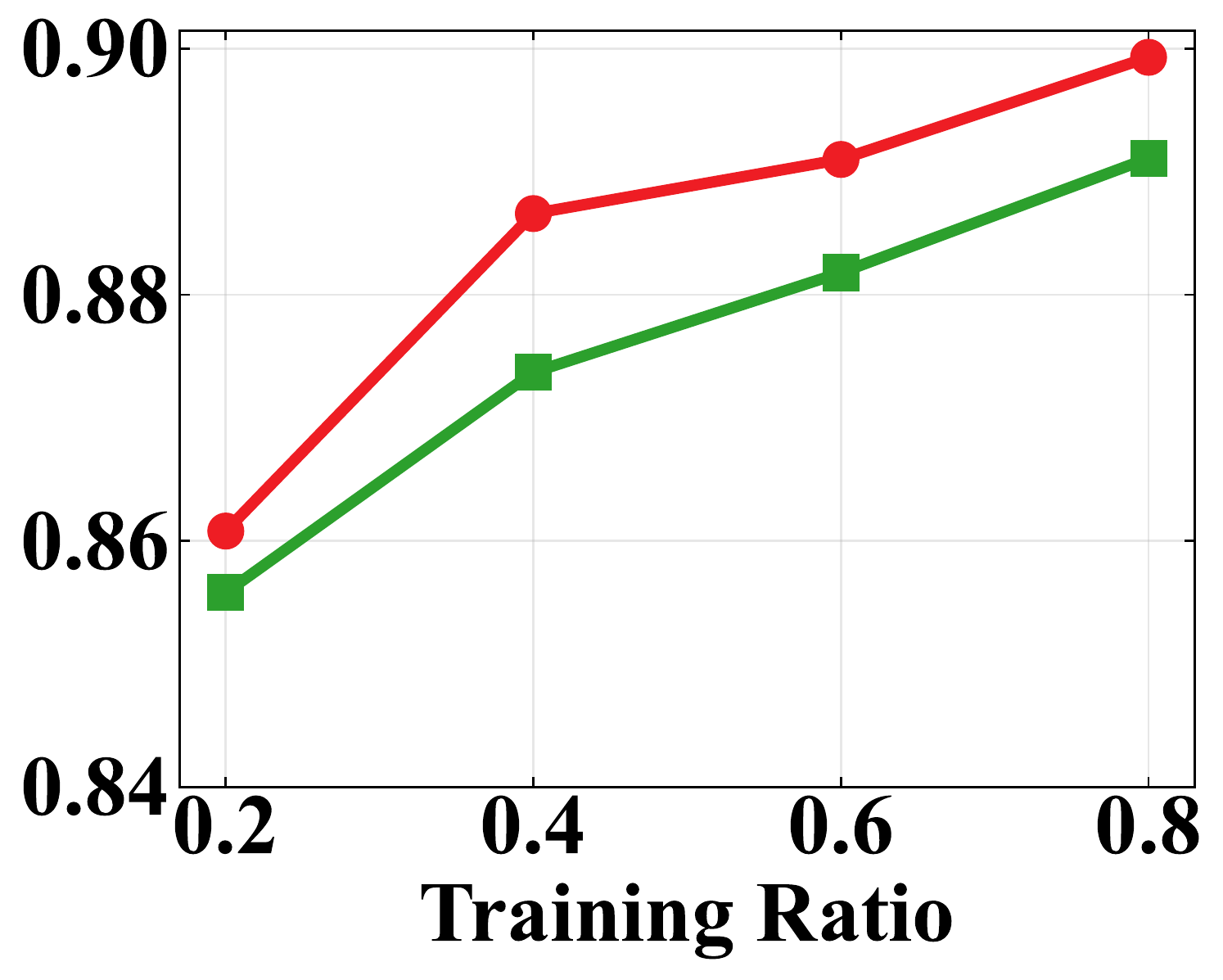}}

    \vspace{-2mm}
    \caption{Experimental results of TrustGuard and baselines on Advogato and PGP datasets.}
    \label{scalability}
    \vspace{-2mm}
\end{figure}

The datasets employed in this study, Bitcoin-OTC and Bitcoin-Alpha, are derived from bitcoin exchange networks. They are the only datasets we can find in the public that contain dynamic trust relationships with labels. To examine whether TrustGuard is applicable to other types of networks, we conduct additional testing based on two large static datasets: Advogato~\cite{massa2009bowling} and PGP~\cite{nr}. They are collected from social networks and public key certification networks, respectively, and contain four trust levels. Detailed information regarding these two datasets is provided in Appendix~\ref{additional_datasets}. Although these static datasets restrict the efficacy of the attention mechanism, our experimental results in Fig.~\ref{scalability} show that TrustGuard can still achieve an average AUC improvement of over 1\% on both datasets with varying training ratios, compared to the best baseline. We do not report the performance of GATrust on PGP, as it runs out of the memory on our machine. These results affirm the scalability of TrustGuard across diverse network types.

\subsection{Basic Trust Properties Support}
TrustGuard can support basic trust properties in terms of subjectivity, dynamicity, context-awareness, asymmetry, and conditionally transferability. By considering these properties, TrustGuard is particularly applicable to trust evaluation. (i) TrustGuard captures the subjective and objective properties of nodes as trustors or trustees through the messages constructed in Eq.~\ref{eq2} and \ref{eq5}. The aggregation of a set of messages associated with a node models its possibility to provide positive or negative ratings to others, which reflects its subjective opinion or behavior. (ii) TrustGuard highly supports dynamicity by splitting a dynamic graph into several snapshots. This allows it to learn meaningful temporal patterns of trust to improve evaluation accuracy. (iii) TrustGuard models asymmetry by considering a node's in-degree and out-degree neighbors, which corresponds to its role as a trustee and a trustor, respectively. This enables TrustGuard to capture different trust behaviors of different roles. (iv) TrustGuard allows conditional transfer of trust from one node to another by stacking several trust propagation layers. In addition, it takes into account the embedding (which can be understood as trustworthiness) of each node within propagation paths. (v) TrustGuard also supports context-awareness. Although we do not include it in TrustGuard since the existing benchmark datasets do not provide contextual information, it is natural and straightforward to encode contextual information into node embeddings. One possible approach is to extend the message construction in Eq.~\ref{eq2} and \ref{eq5} by concatenating the contextual information with other information, as shown below:
\begin{equation}
    \setlength{\abovedisplayskip}{4pt}
    \setlength{\belowdisplayskip}{4pt}
    Msg_{u \gets v,q}=h_v \otimes \omega_{u \gets v,q} \otimes C_q,
    \nonumber
\end{equation}
where $\omega_{u \gets v,q}$ denotes the trust relationship from $v$ to $u$ in a specific context $q$, and $C_q$ denotes the embedding of the context $q$. Since there may be several contexts, it is also promising to perform trust evaluation under a specific context by granting different attentions to different contexts for aggregation.


\subsection{Node \& Community Trust Evaluation}
TrustGuard has demonstrated superiority with respect to the evaluation of trust relationships between any two nodes, which belongs to edge classification. Except for this, TrustGuard has the potential to be extended for evaluating node trust and community trust, which relate to node classification and graph classification, respectively. This is owing to the fact that node embeddings learned by TrustGuard can be used for a variety of downstream tasks. Specifically, for node trust evaluation, the prediction layer in TrustGuard takes the embedding of each node as an input. For community trust evaluation, it regards the average embeddings of nodes in a community as an input. However, due to the lack of node/community trust values in existing benchmark datasets, 
we cannot demonstrate the effectiveness of TrustGuard on these two application scenarios through experiments.

\subsection{Fine-grained Trust Evaluation Support}

Fine-grained trust evaluation refers to the ability to numerically quantify trust in a detailed manner. It has several advantages: (i) It makes possible to calculate different numeric values of trust, thereby offering great flexibility for managing trust relationships. (ii) It allows entities or systems to make precise judgments based on trust values, enabling effective and informative decision-making. (iii) It supports personalized or customized trust evaluation, improving user satisfaction on evaluation.
The results shown in Fig.~\ref{scalability} demonstrate that TrustGuard can well support fine-grained trust evaluation.


\subsection{Limitations of TrustGuard}
First, the proposed defense mechanism is heuristic, based on the network theory of homophily~\cite{mcpherson2001birds}, insights about trust relationships~\cite{chen2014trust,wang2021c}, and analyses on adversarial attacks~\cite{wu2019adversarial,jin2020graph}. While its key idea has been proven effective and widely accepted in relevant fields, its generality should be further enhanced. For example, it is unclear whether TrustGuard is equally effective in dealing with graphs characterized by heterophily. Additionally, it is imperative to investigate a new defense mechanism with theoretical guarantees. Second, we focus on typical attacks in the context of trust evaluation in this paper. However, it is important to note that GNN itself could suffer from some vulnerabilities, e.g., being prone to adversarial attacks. In such attacks, there exists a singular adversary with full or partial knowledge of TrustGuard’s architecture and the entire graph (i.e., dataset). This adversary strategically manipulates the graph structure to bypass the specifics of TrustGuard by using a specific algorithm or formulating an optimization problem~\cite{sharma2023temporal,chen2023tkde}. Thus, this type of attack emphasizes the adversarial relationship between the adversary and the GNN model from a global view, whereas our focus is the anomalous behavior of nodes from a local view. How to attack TrustGuard by a strategic adversary remains further exploration, which can help us uncover the vulnerability of GNN models and facilitate the development of trustworthy GNNs~\cite{zhang2022trustworthy}.


\section{Conclusion} \label{section7}
In this paper, we proposed TrustGuard, a GNN-based trust evaluation model that supports trust dynamicity, is robust against typical attacks, and provides convincing explanations. TrustGuard was designed with a layered architecture, where the proposed defense mechanism in the spatial aggregation layer can facilitate the robust aggregation of local trust relationships, while the position-aware attention mechanism adopted in the temporal aggregation layer can help in capturing temporal patterns from a sequence of snapshots. Extensive experiments demonstrated that TrustGuard has superiority over state-of-the-art GNN-based trust evaluation models in terms of trust prediction across single-timeslot and multi-timeslot, no matter whether there are attacks or not. To explain the evaluation results derived from TrustGuard, we visualized robust coefficients and attention scores. A user study validated their effectiveness in enhancing the explainability of TrustGuard. Ablation studies further confirmed the necessity and rationality of the key designs of TrustGuard.

\section*{Acknowledgments}
This work is supported in part by the National Natural Science Foundation of China under Grant U23A20300 and 62072351; in part by the Key Research Project of Shaanxi Natural Science Foundation under Grant 2023-JC-ZD-35; and in part by the 111 Project under Grant B16037.

\bibliographystyle{IEEEtran}
\bibliography{reference}

\begin{IEEEbiography}[{\includegraphics[width=1in,height=1.25in,clip,keepaspectratio]{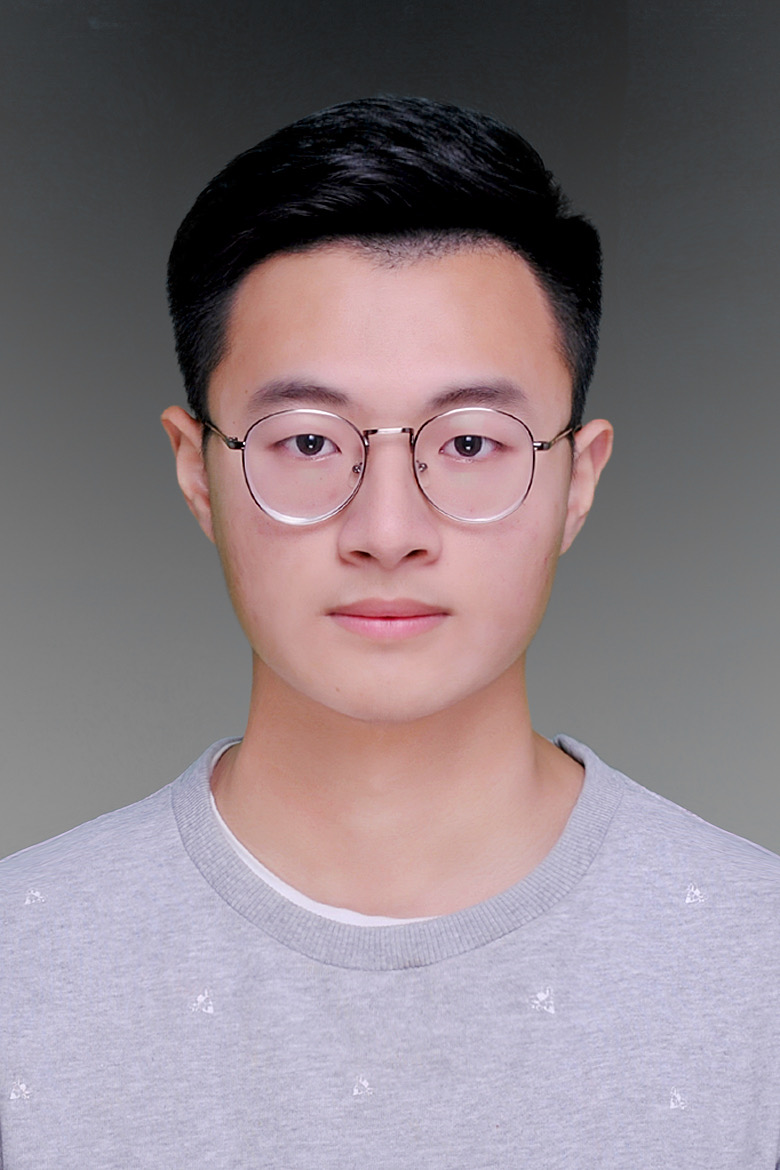}}]{Jie Wang} received the B.S. degree in network engineering from Xidian University in 2020, where he is currently pursuing the Ph.D. degree with the School of Cyber Engineering. His research interests include graph neural networks, trust evaluation, and explainability.
\end{IEEEbiography}

\begin{IEEEbiography}[{\includegraphics[width=1in,height=1.25in,clip,keepaspectratio]{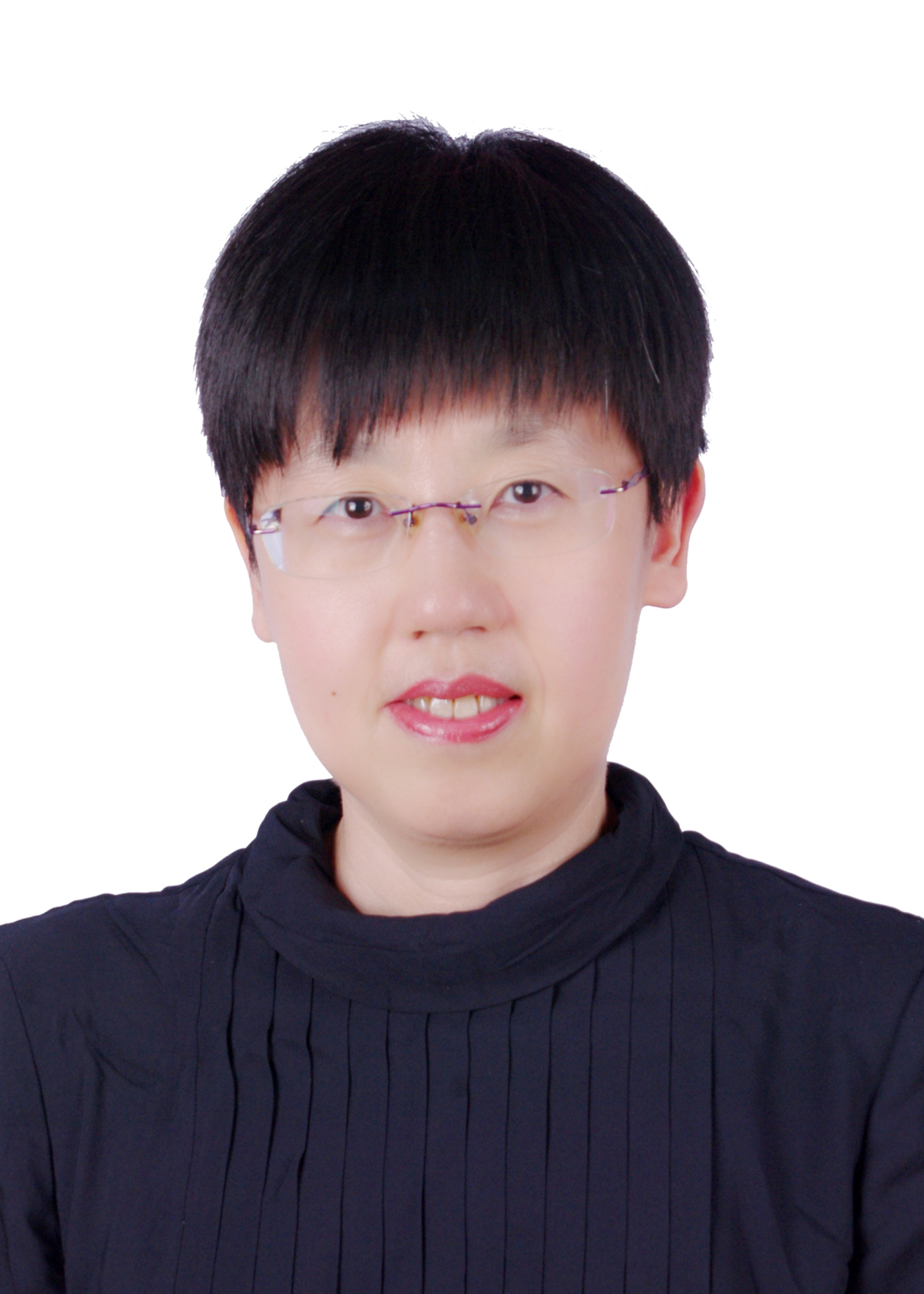}}]{Zheng Yan} is currently a Huashan distinguished professor at the Xidian University, China. She joined the Nokia Research Center, Helsinki in 2000, working as a senior researcher until 2011. She worked as a visiting professor and a Finnish Academy Fellow at the Aalto University, Finland for over seven years. She earned her PhD degree from the Helsinki University of Technology, Finland. Her research interests are in cyber trust, security, privacy, and data analytics. She has authored 370+ publications, with 260+ first and corresponding authorships, featured prominently in top-tier venues. She is the sole author of two books on trust management, utilized in teaching for a decade. 130+ patents invented by her have been adopted by industry, a few of them have been incorporated into international standards and widely used in practice. She has delivered about 30 invited keynote speeches and talks at international conferences and world-leading companies. Dr. Yan served and is serving as an area/associate/guest editor for 60+ reputable journals, a steering committee member of several conferences, a general/program chair for 30+ international conferences. She is a founding steering committee co-chair of IEEE International Conference on Blockchain. Dr. Yan is recognized as a World Top 2\% Scientist and an Elsevier Highly Cited Chinese Researcher. She has earned many accolades, including Distinguished Inventor of Nokia for significant technology contributions, N²Women Star in Computer Networking and Communications, the IEEE TCSC Award for Excellence in Scalable Computing, the ELEC Impact Award for patent contributions to Finnish society, the Best Journal Paper Award issued by IEEE ComSoc and several other Best Paper awards, 18 times IEEE Distinguished/Outstanding Leadership/Service awards, three EU awards, etc. She is a fellow of IEEE, IET and AAIA. Her excellence has been covered by the media on numerous occasions.
\end{IEEEbiography}

\begin{IEEEbiography}[{\includegraphics[width=1in,height=1.25in,clip,keepaspectratio]{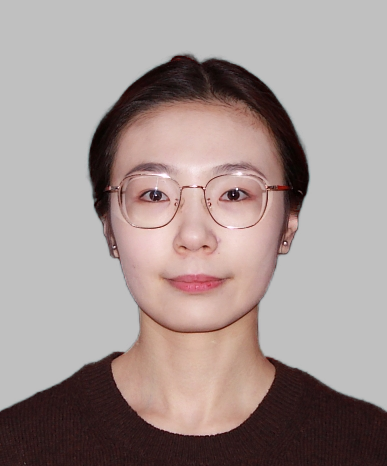}}]{Jiahe Lan} received the B.S. degree in Network Engineering from Xidian University, Xi’an, China, in 2020, where she is currently pursuing the Ph.D. degree with the School of Cyber Engineering. Her research interests include adversarial machine learning, voice processing systems, and human–computer interaction.
\end{IEEEbiography}

\begin{IEEEbiography}[{\includegraphics[width=1in,height=1.25in,clip,keepaspectratio]{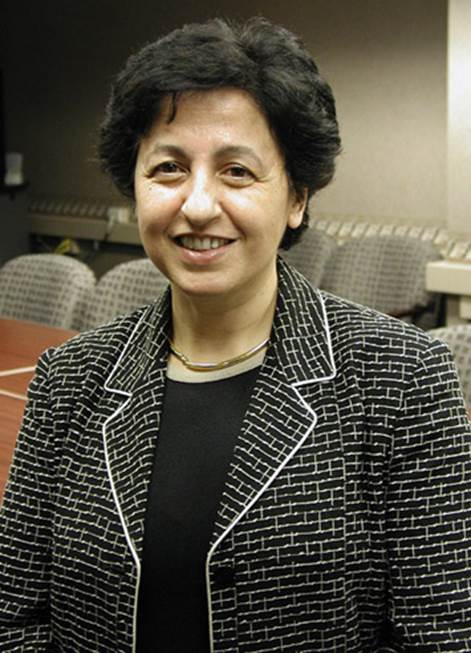}}]{Elisa Bertino} is Samuel Conte professor of Computer Science at Purdue University. Prior to joining Purdue, she was a professor and department head at the Department of Computer Science and Communication of the University of Milan. She has been a visiting researcher at the IBM Research Laboratory in San Jose (now Almaden), at Rutgers University, at Telcordia Technologies. She has also held visiting professor positions at the Singapore National University and the Singapore Management University.  Her recent research focuses on security and privacy of cellular networks and IoT systems, and on edge analytics for cybersecurity.  Elisa Bertino is a Fellow member of IEEE, ACM, and AAAS. She received the 2002 IEEE Computer Society Technical Achievement Award for “For outstanding contributions to database systems and database security and advanced data management systems”, the 2005 IEEE Computer Society Tsutomu Kanai Award for “Pioneering and innovative research contributions to secure distributed systems”, the 2019-2020 ACM Athena Lecturer Award, and the 2021 IEEE 2021 Innovation in Societal Infrastructure Award. She received an Honorary Doctorate from Aalborg University in 2021 and a Research Doctorate in Computer Science from the University of Salerno in 2023. She has served as EiC on IEEE Transactions on Dependable and Secure Computing and as Chair of the ACM Special Interest Group on Security, Audit, and Control (SIGSAC). She is currently serving as ACM Vice-president.
\end{IEEEbiography}

\begin{IEEEbiography}[{\includegraphics[width=1in,height=1.25in,clip,keepaspectratio]{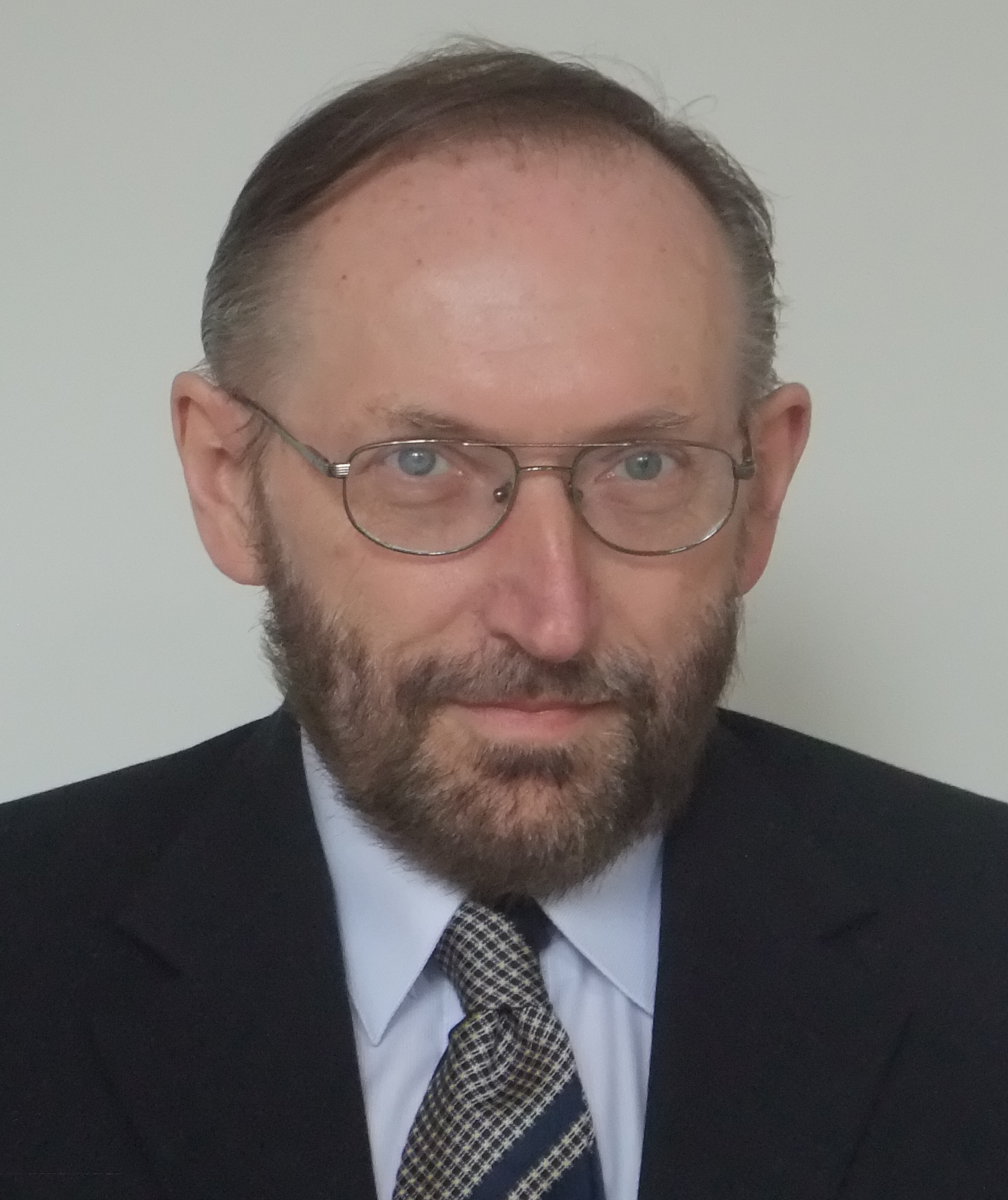}}]{Witold Pedrycz} is currently a Professor and the Canada Research Chair in computational intelligence with the Department of Electrical and Computer Engineering, University of Alberta, Edmonton, AB, Canada. He is also with the Systems Research Institute, Polish Academy of Sciences, Warsaw, Poland. He is a Foreign Member of the Polish Academy of Sciences. His research interests include computational intelligence, data mining, and software engineering. He was a fellow of the Royal Society of Canada, Ottawa, ON, Canada, in 2012. He received a prestigious Norbert Wiener Award from the IEEE Systems, Man, and Cybernetics Society in 2007, the IEEE Canada Computer Engineering Medal, the Cajastur Prize for Soft Computing from the European Center for Soft Computing, the Killam Prize, and the Fuzzy Pioneer Award from the IEEE Computational Intelligence Society. He serves on the Advisory Board of the IEEE TRANSACTIONS ON FUZZY SYSTEMS. He is the Editor-in-Chief of Information Sciences, WIREs Data Mining and Knowledge Discovery, and International Journal of Granular Computing.
\end{IEEEbiography}

\clearpage
\setcounter{page}{1}

\begin{figure*}[!htb]
    \centering
    \subfigcapskip=-3pt
    \hspace{4mm}
    \subfigure{\includegraphics[scale=0.24]{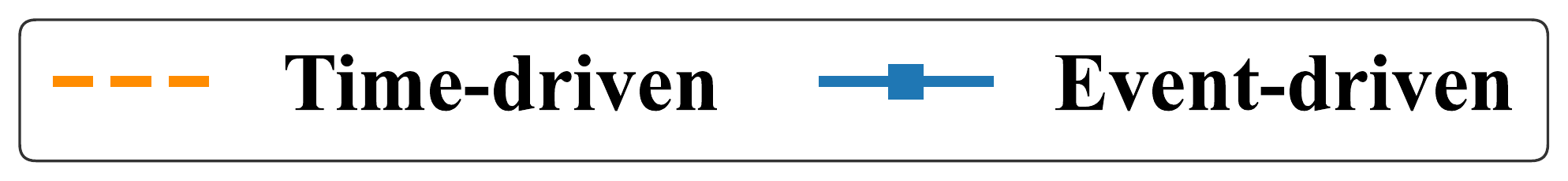}}
    \vspace{-3mm}

    \setcounter{subfigure}{0}
    \subfigure[\ding{172}+Bitcoin-OTC]{\includegraphics[scale=0.26]{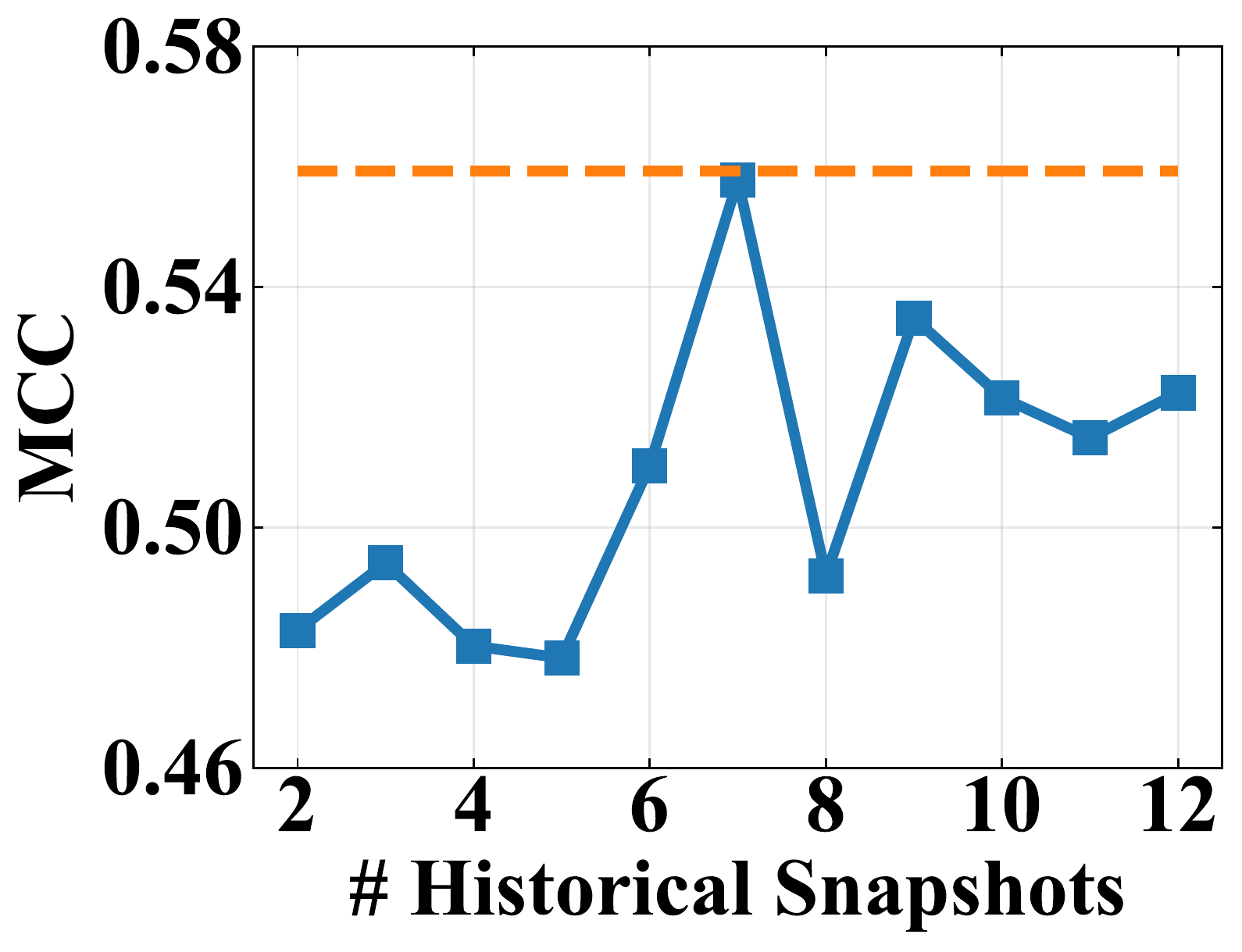}}
    \subfigure[\ding{173}+Bitcoin-OTC]{\includegraphics[scale=0.26]{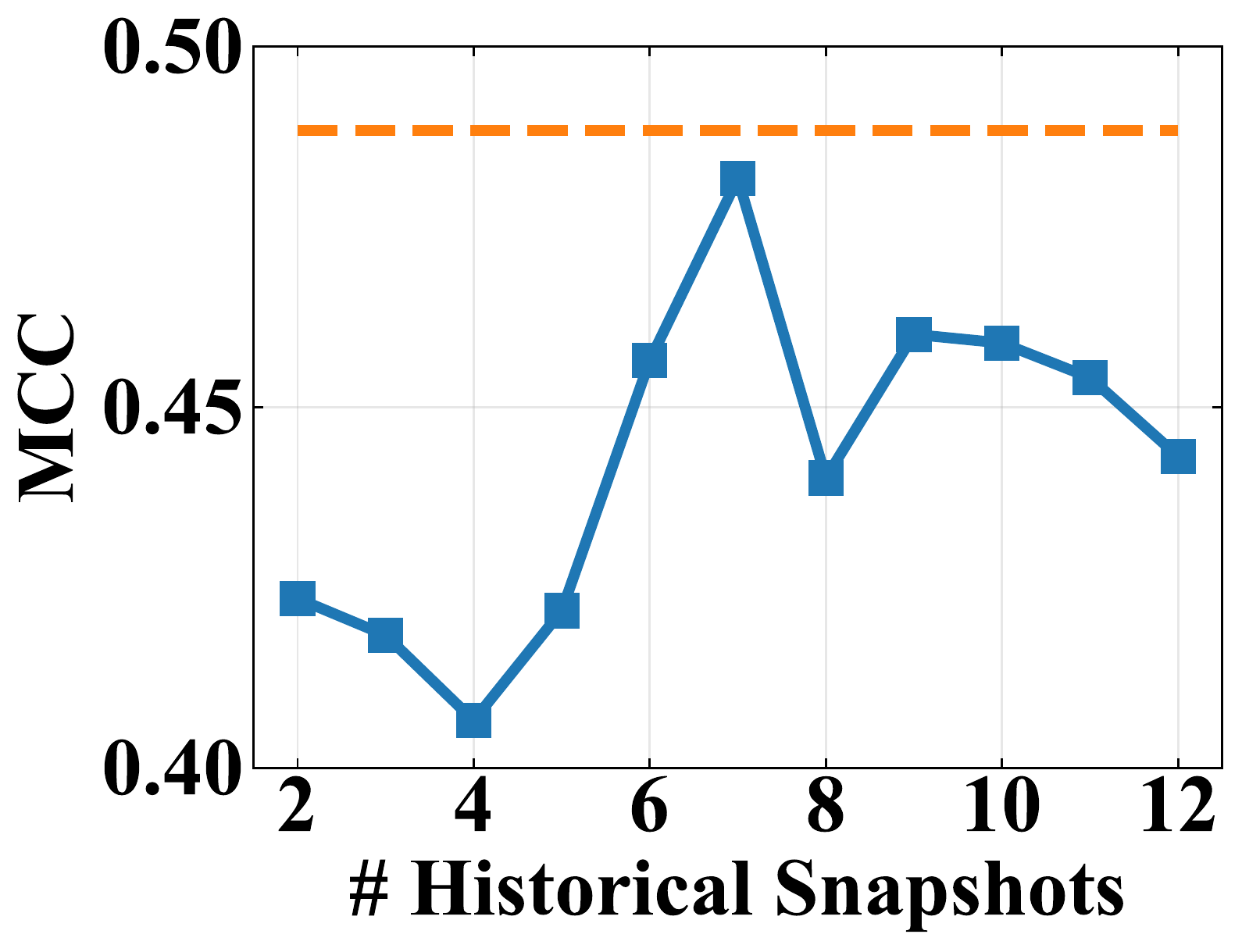}}
    \subfigure[\ding{172}+Bitcoin-Alpha]{\includegraphics[scale=0.26]{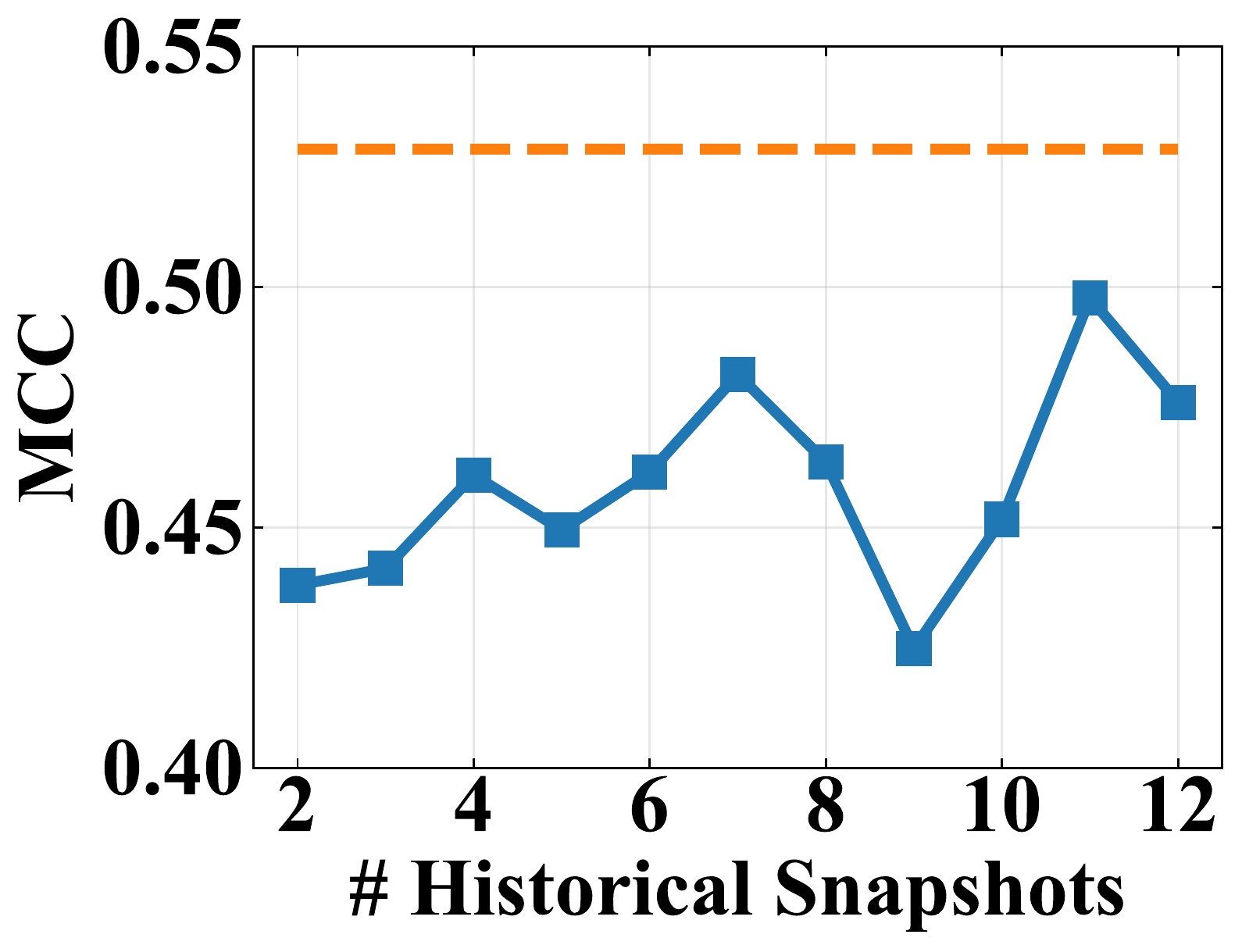}}
    \subfigure[\ding{173}+Bitcoin-Alpha]{\includegraphics[scale=0.26]{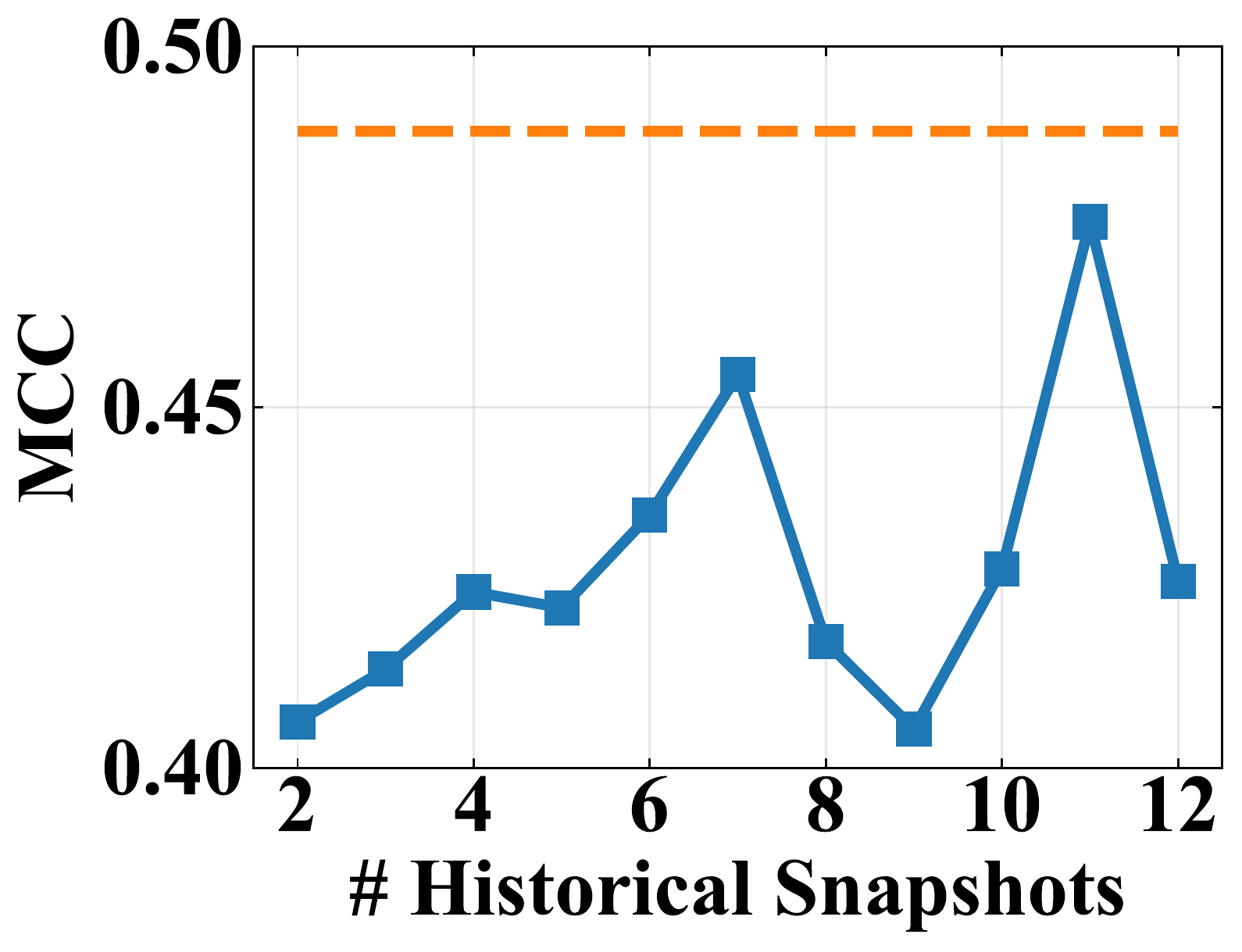}}

    \vspace{-2mm}
    \caption{Experimental results of TrustGuard based on time-driven and event-driven strategies. \ding{172} denotes the single-timeslot prediction on observed nodes. \ding{173} denotes the multi-timeslot prediction on observed nodes.}
    \label{event_driven}
\end{figure*}

\begin{figure*}[tbp]
    \centering
    \subfigcapskip=-3pt
    \hspace{7mm}
    \subfigure{\includegraphics[scale=0.26]{legend.pdf}}
    \vspace{-1mm}

    \setcounter{subfigure}{0}
    \hspace{-6mm}
    \subfigure[\ding{172}+Bitcoin-OTC]{
        \begin{minipage}[b]{0.2\textwidth}
        \includegraphics[scale=0.26]{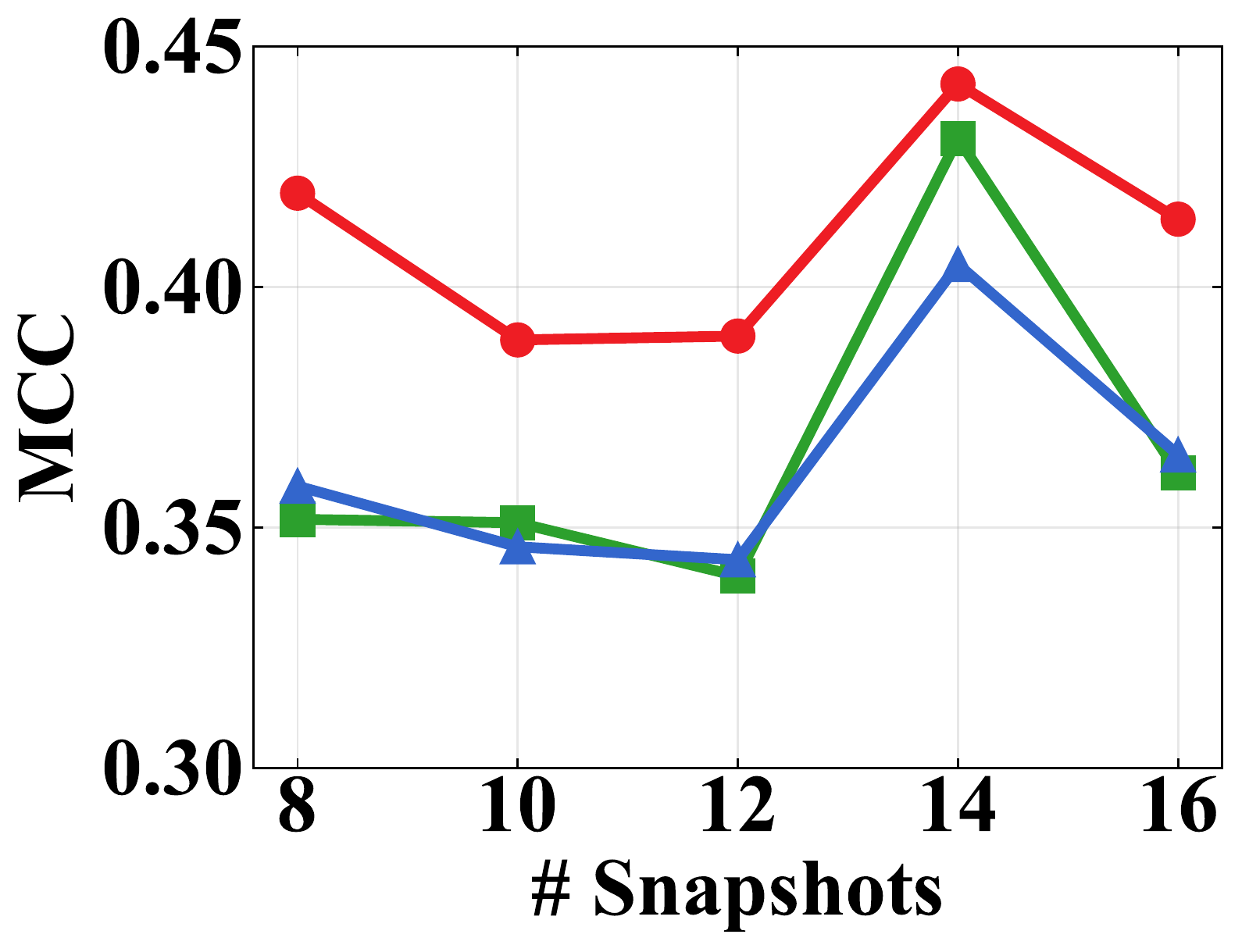}
        \end{minipage}
        \hspace{3mm}
        \begin{minipage}[b]{0.2\textwidth}
        \includegraphics[scale=0.26]{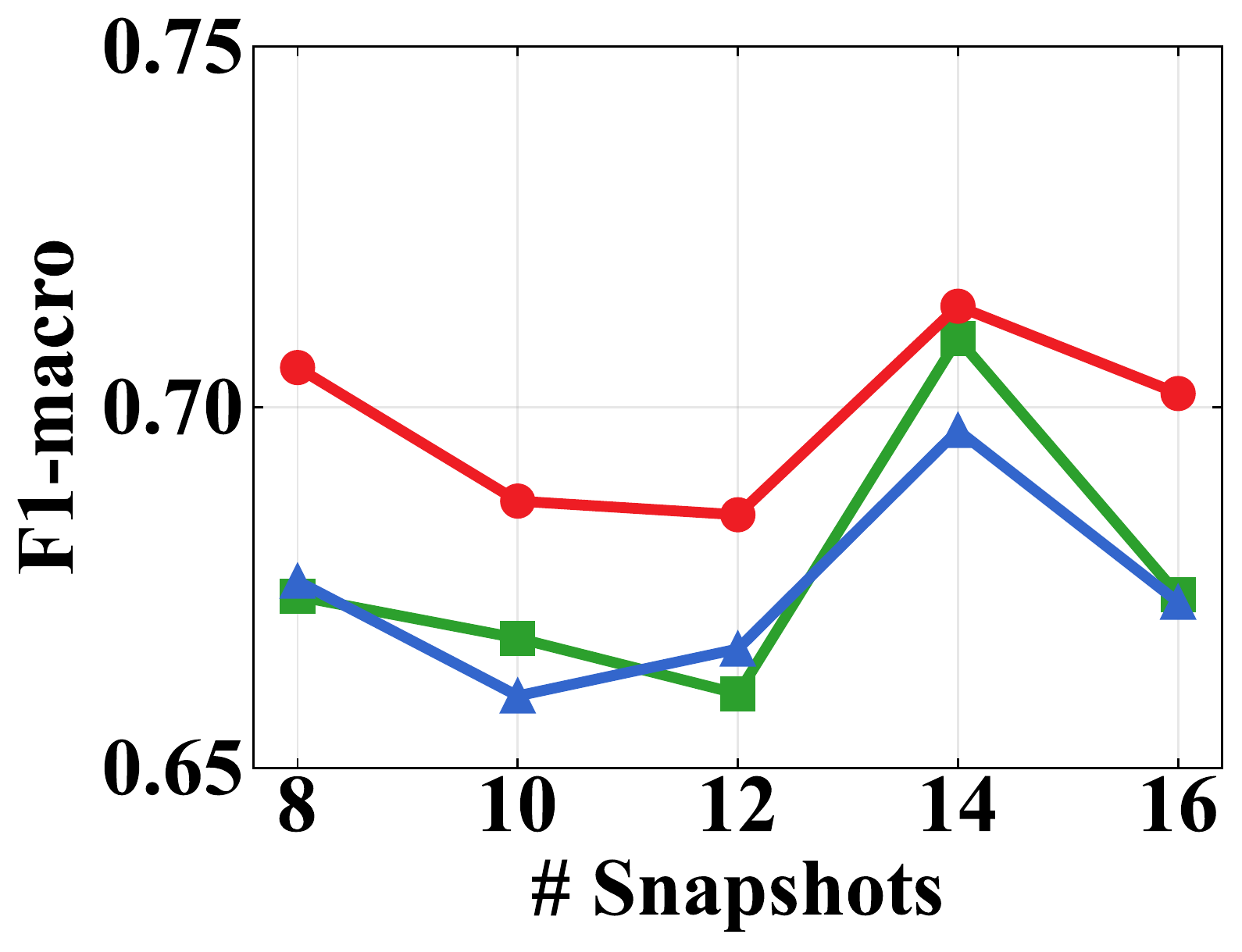}
        \end{minipage}
    }
    \hspace{2mm}
    \subfigure[\ding{172}+Bitcoin-Alpha]{
        \begin{minipage}[b]{0.2\textwidth}
        \includegraphics[scale=0.26]{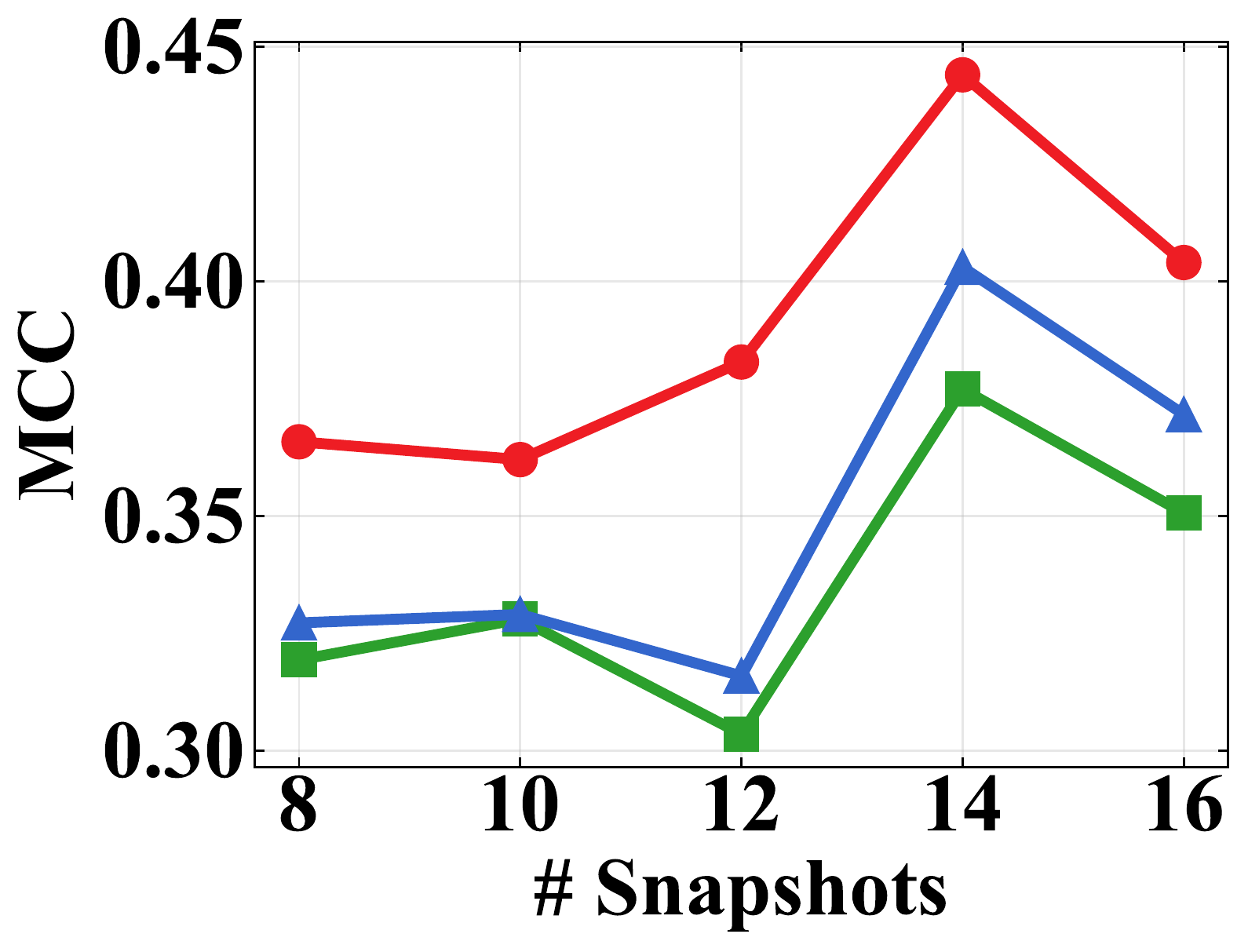}
        \end{minipage}
        \hspace{3mm}
        \begin{minipage}[b]{0.2\textwidth}
        \includegraphics[scale=0.26]{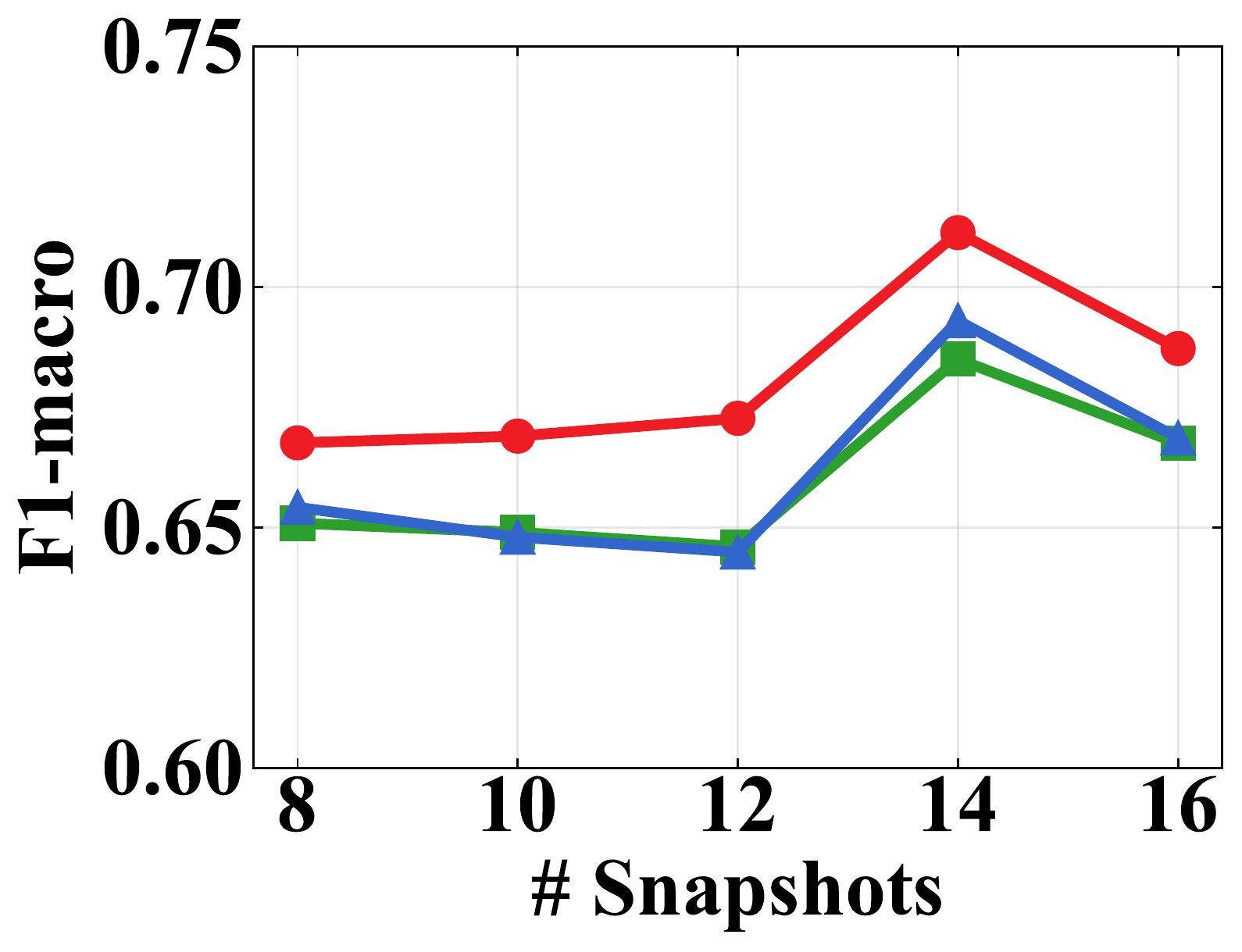}
        \end{minipage}
    }

    \hspace{-6mm}
    \subfigure[\ding{173}+Bitcoin-OTC]{
        \begin{minipage}[b]{0.2\textwidth}
        \includegraphics[scale=0.26]{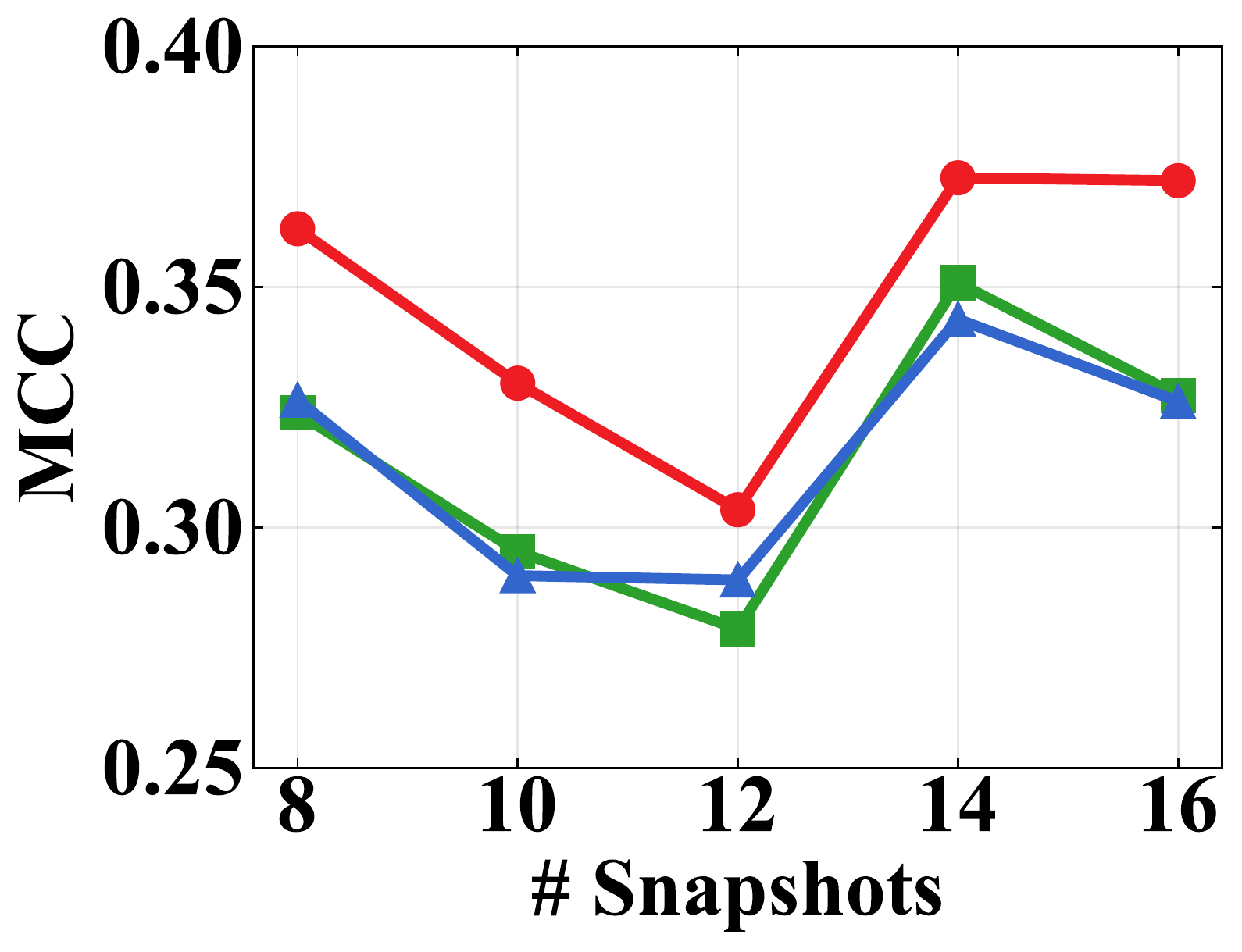}
        \end{minipage}
        \hspace{3mm}
        \begin{minipage}[b]{0.2\textwidth}
        \includegraphics[scale=0.26]{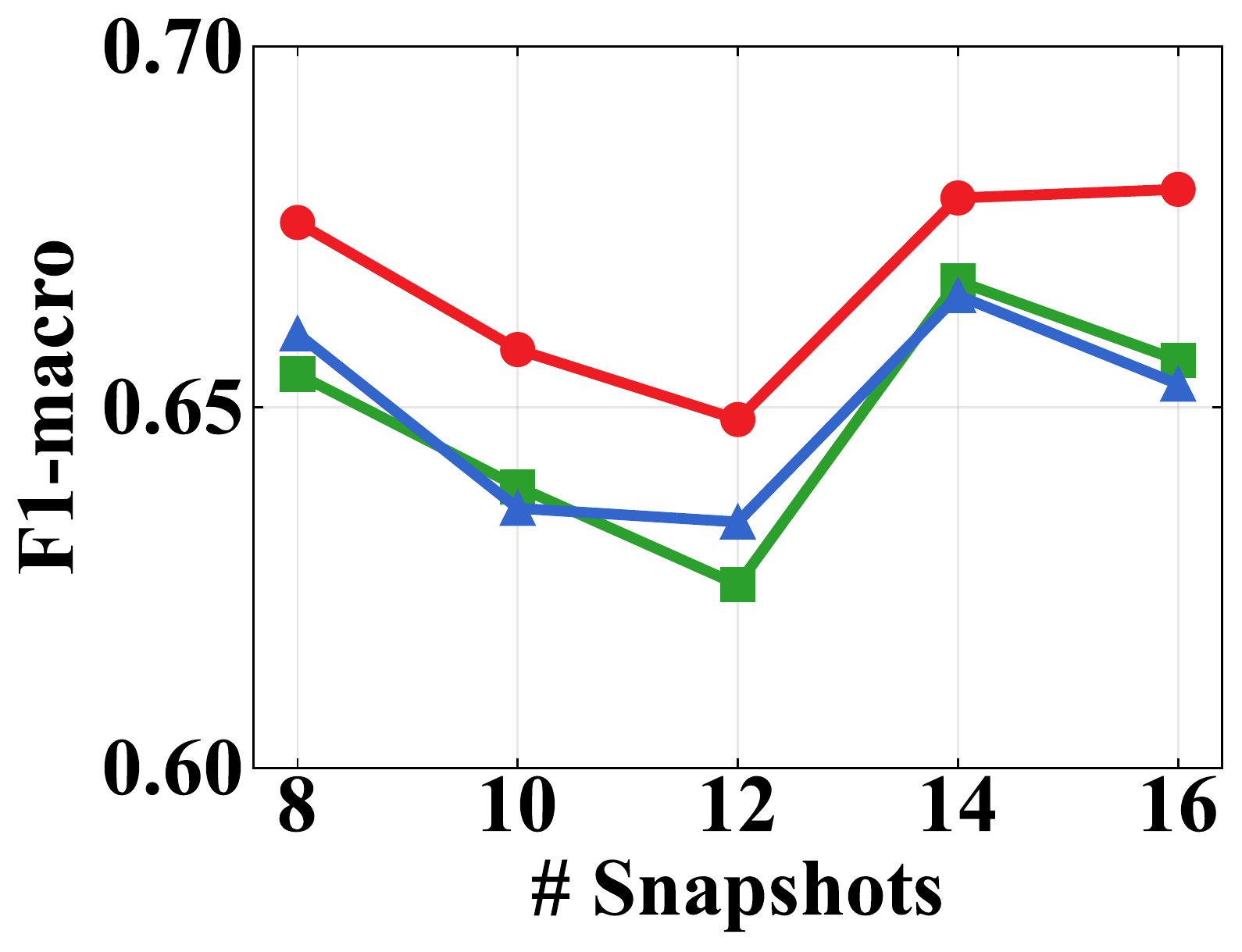}
        \end{minipage}
    }
    \hspace{2mm}
    \subfigure[\ding{173}+Bitcoin-Alpha]{
        \begin{minipage}[b]{0.2\textwidth}
        \includegraphics[scale=0.26]{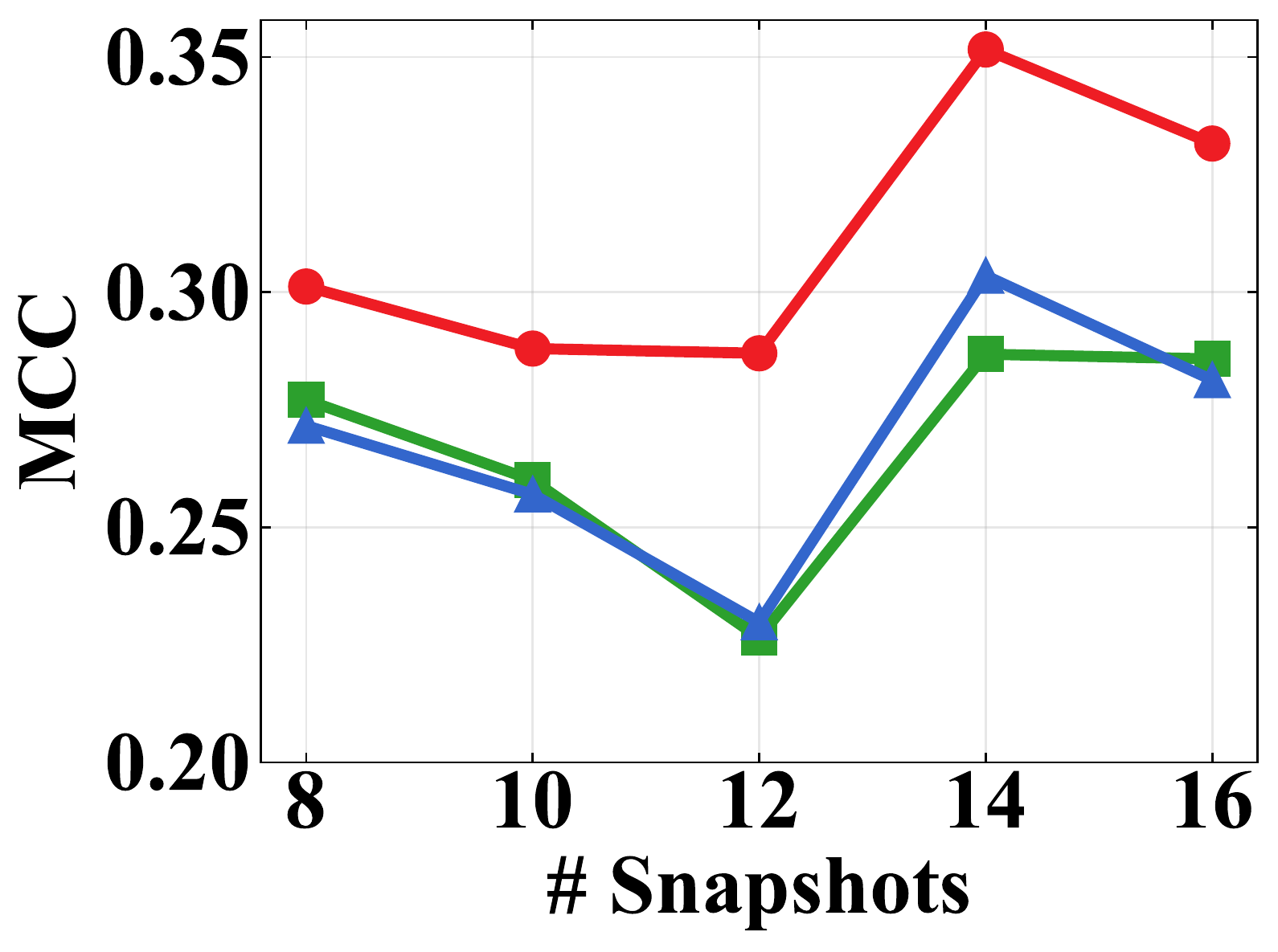}
        \end{minipage}
        \hspace{3mm}
        \begin{minipage}[b]{0.2\textwidth}
        \includegraphics[scale=0.26]{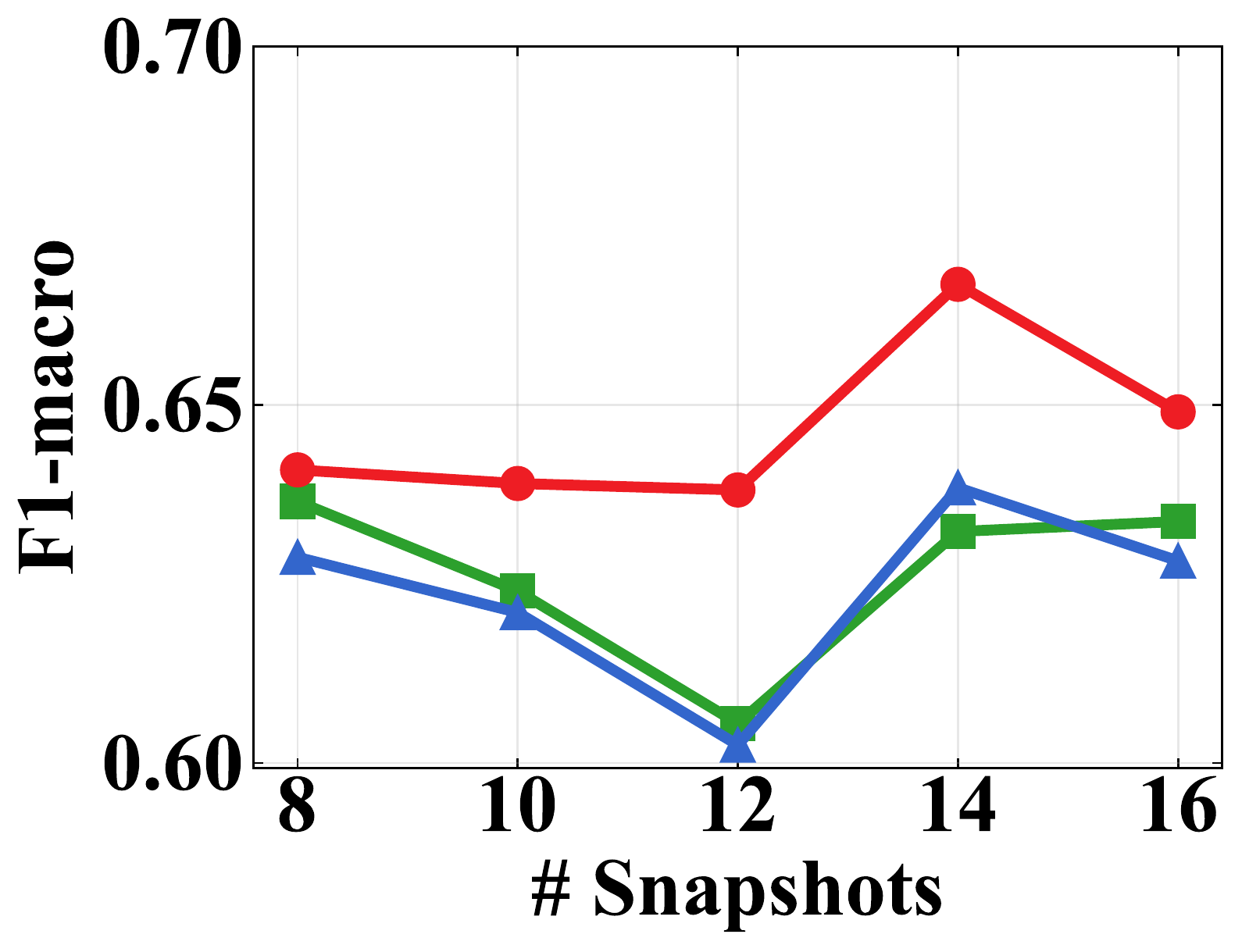}
        \end{minipage}
    }

    \hspace{-6mm}
    \subfigure[\ding{174}+Bitcoin-OTC]{
        \begin{minipage}[b]{0.2\textwidth}
        \includegraphics[scale=0.26]{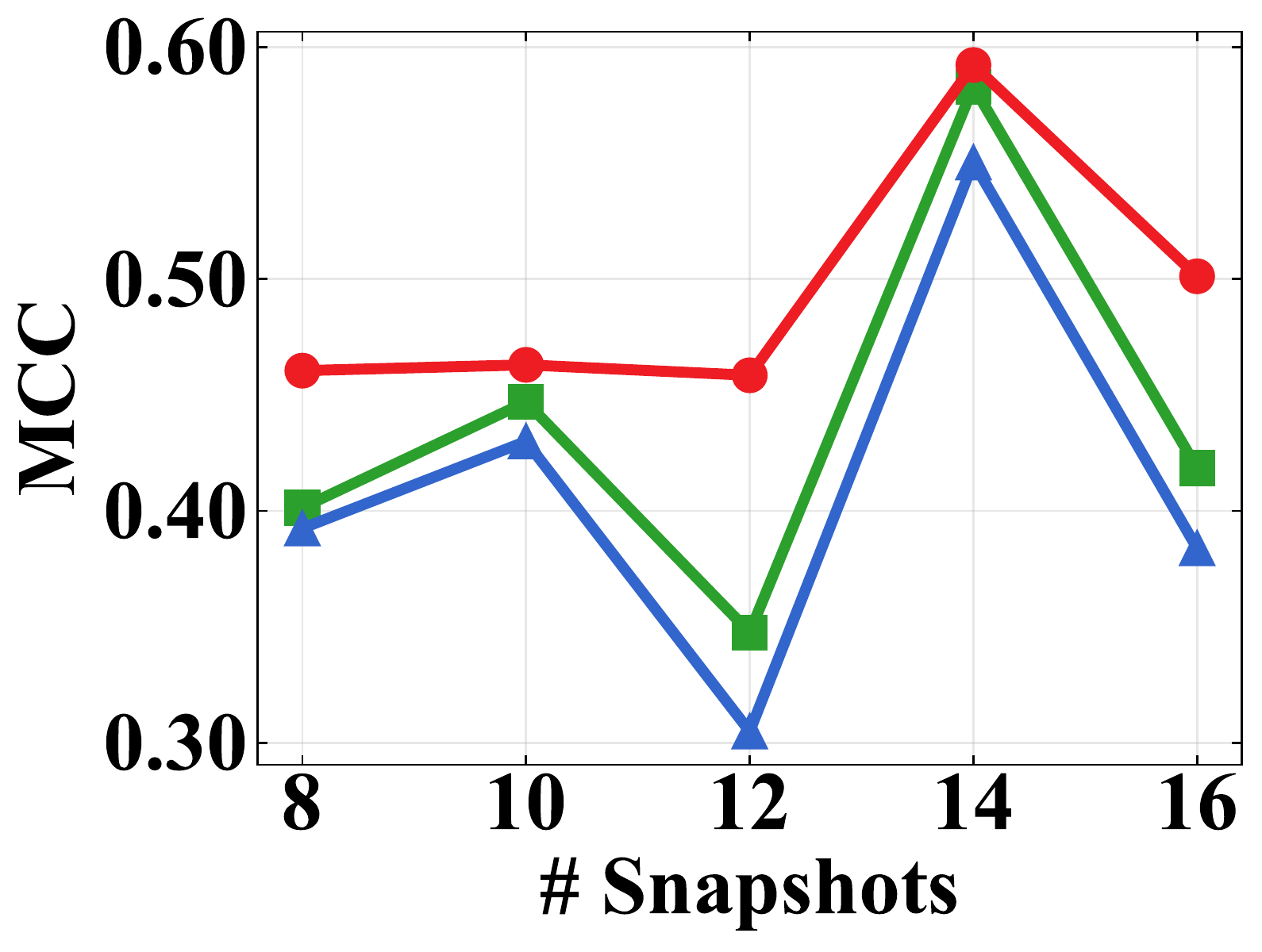}
        \end{minipage}
        \hspace{3mm}
        \begin{minipage}[b]{0.2\textwidth}
        \includegraphics[scale=0.26]{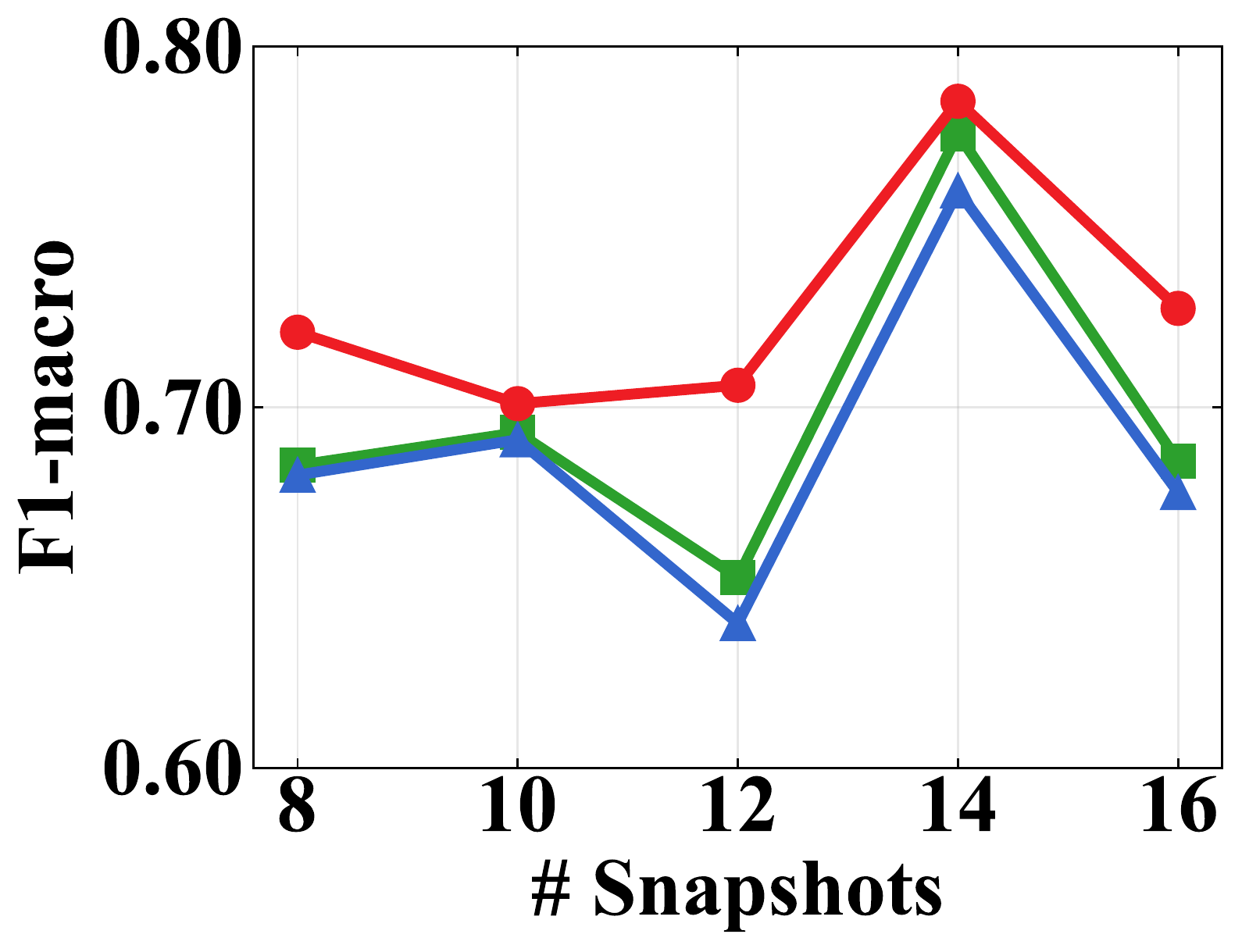}
        \end{minipage}
    }
    \hspace{2mm}
    \subfigure[\ding{174}+Bitcoin-Alpha]{
        \begin{minipage}[b]{0.2\textwidth}
        \includegraphics[scale=0.26]{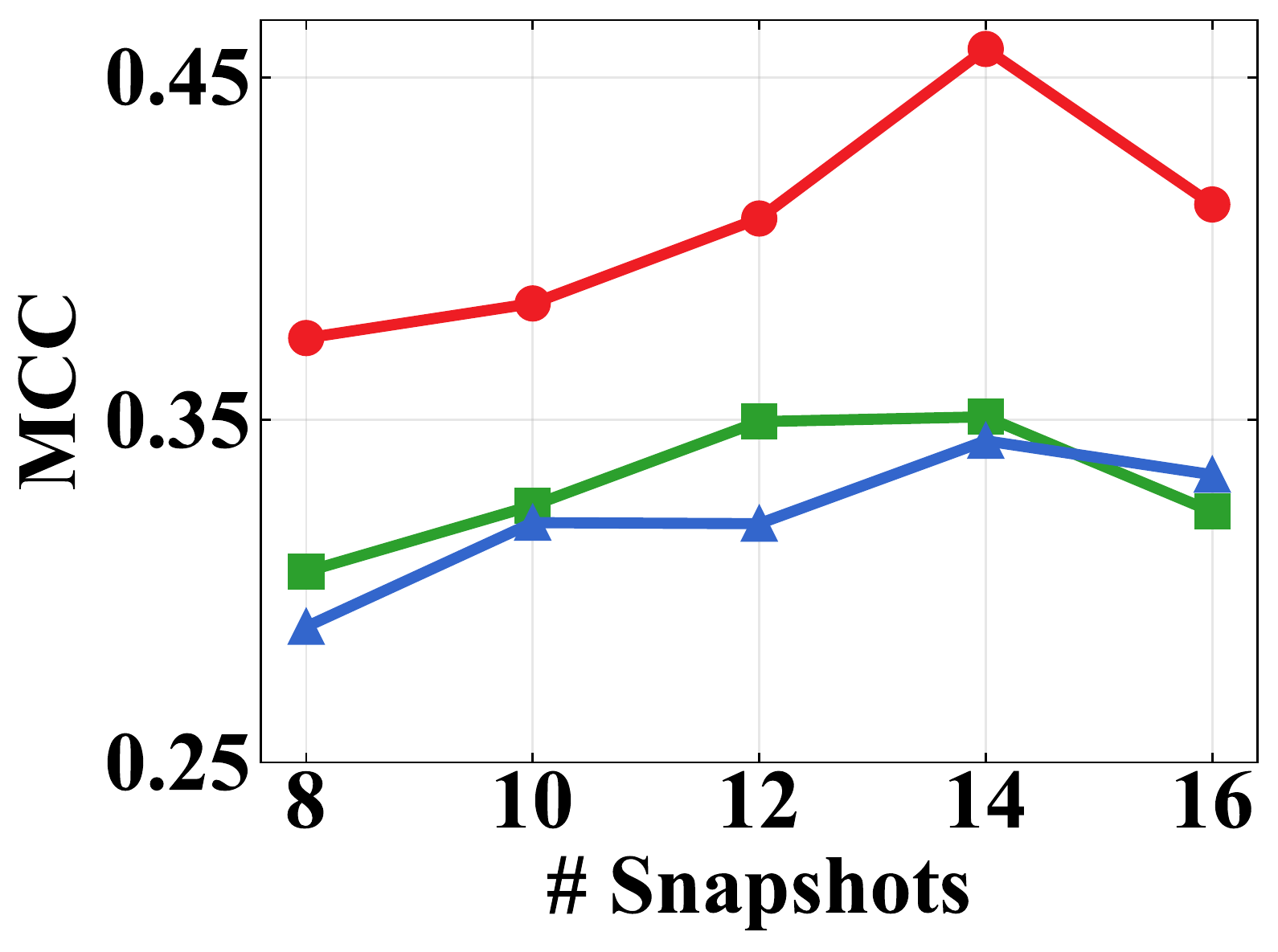}
        \end{minipage}
        \hspace{3mm}
        \begin{minipage}[b]{0.2\textwidth}
        \includegraphics[scale=0.26]{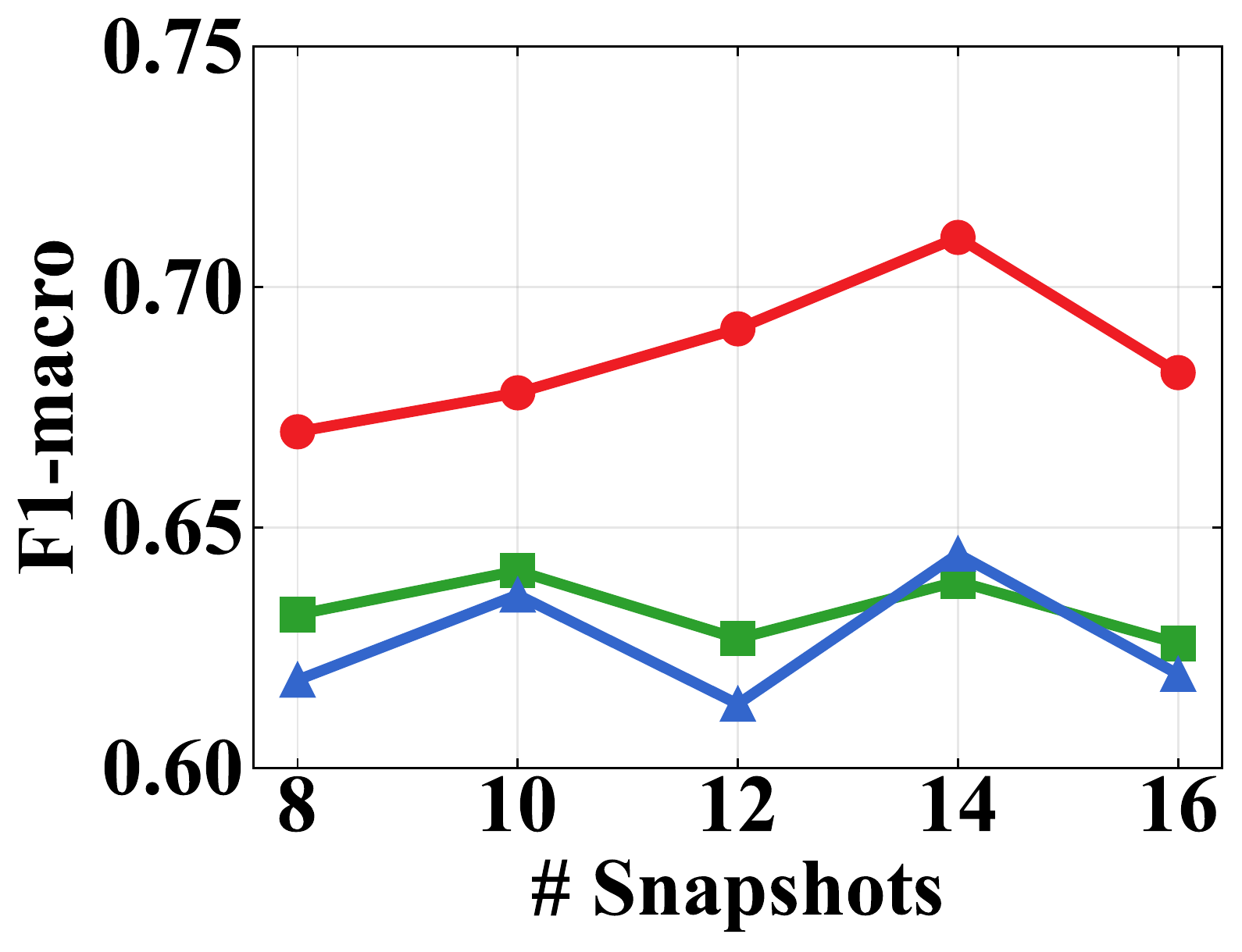}
        \end{minipage}
    }

    \vspace{-2mm}
    \caption{Impact of observation frequency on TrustGuard and baselines in the absence of attacks. \ding{172} denotes the single-timeslot prediction on observed nodes. \ding{173} denotes the multi-timeslot prediction on observed nodes. \ding{174} denotes the single-timeslot prediction on unobserved nodes.}
    \label{number of snapshots}
\end{figure*}

\appendices
\section{Comparison of Time-Driven and Event-Driven Strategies} \label{appdendix_events}
In this appendix, we first thoroughly discuss the pros and cons of time-driven and event-driven strategies by considering their impact on GNN models. Then, we experimentally investigate how these two strategies affect the performance of TrustGuard.

\subsection{Discussion}
The event-driven strategy ensures that each snapshot is equally rich in events, which focuses on how the events affect the graph’s structure, i.e., the significance of events. However, this could lead to irregular time intervals between snapshots, posing a challenge for GNN models in learning effective temporal patterns. Moreover, these irregular intervals can obscure the actual temporal relationships between events. For instance, rapidly consecutive events might be important, but if they are dispersed across different snapshots, their contextual relationship could be lost. The time-driven strategy mitigates this issue by offering a consistent temporal frame for GNN models to effectively learn temporal patterns, such as seasonality and increasing trends~\cite{min2021stgsn}. This strategy is simple to implement and can be applied to handle bursty events. However, it may miss some important temporal information if its observation frequency is not appropriately set~\cite{zhu2022learnable}.

\subsection{Experimental Evaluation}
To fairly compare the time-driven and event-driven strategies, it is imperative to maintain a fixed training-testing ratio for both. Thus, we follow the setting in Section~\ref{rq2} and focus on single-timeslot prediction on observed nodes and multi-timeslot prediction on observed nodes. We alter the number of events per snapshot by adjusting the number of historical snapshots utilized for training. A large number of historical snapshots corresponds to a small number of events per snapshot, and vice versa.

Fig.~\ref{event_driven} presents the comparison results. The MCC of the time-driven strategy is constant since it uses the first seven snapshots for training. For the event-driven strategy, we use the same training data, dividing it into different numbers of snapshots. It can be observed from Fig.~\ref{event_driven} that the time-driven strategy outperforms the event-driven strategy in both trust prediction tasks across both datasets. One possible reason is that the event-driven strategy causes irregular time intervals between snapshots, which severely restricts the capability of the attention mechanism that is used in the temporal aggregation layer in learning temporal patterns. Additionally, both real-world datasets about bitcoin exchange are characterized by bursty rating behaviors (i.e., events)~\cite{sankar2020dysat}, which may be misrepresented by the event-driven strategy.

\section{Impact of the Frequency of Observation on TrustGuard} \label{appdendix_snapshots}
The effectiveness of snapshot-based dynamic graph analysis in TrustGuard depends on the frequency of observation. If the observation frequency is not appropriately set, it may result in a loss of critical temporal information. Consequently, we investigate how the observation frequency, which directly affects the total number of snapshots, impacts TrustGuard.

As shown in Fig.~\ref{number of snapshots}, we observe that TrustGuard achieves significant improvements over Guardian and GATrust in all three types of trust prediction tasks, regardless of the observation frequency setting. This further emphasizes the necessity and advantages of dynamic graph analysis. We also notice that Guardian and GATrust have similar prediction performance as both of them are static models that do not make full use of the temporal information from both datasets to learn potential temporal patterns of trust. This greatly affects their learning accuracy. Additionally, we find that the observation frequency has a certain influence on the effectiveness of TrustGuard. For example, in the Bitcoin-Alpha dataset, TrustGuard with the division of 14 snapshots achieves the best performance, whereas its performance deteriorates with the division of 8 or 16 snapshots. This implies that a low observation frequency may pose challenges for TrustGuard in learning temporal patterns due to a long interval covered in a snapshot, whereas a high observation frequency may result in crucial temporal patterns being missed. Thus, how to set an appropriate observation frequency, particularly in an intelligent manner, is worth further efforts. In addition, it is also promising to explore the effectiveness of hybrid strategies that can complement the limitations of both time-driven and event-driven strategies, offering more flexibility in practice.

\section{Background Information of Participants}\label{background}
Table~\ref{participants} presents the background information of all participants.

\section{Advogato and PGP Datasets} \label{additional_datasets}
Advogato~\cite{massa2009bowling} is an online social network for open-source software developers. In this network, a trust edge from $u$ to $v$ represents that $u$ trusts $v$'s software development ability. The trust relationships can be divided into four types, i.e., $\{Observer, Apprentice, Journeyer, Master\}$. Pretty-Good-Privacy (PGP)~\cite{nr} is a public certification network, where a trust edge from $u$ to $v$ represents that $u$ attests to $v$’s trust. There are also four levels of trust in this dataset. The statistics of the two datasets can be found in Table~\ref{additional_dataset}.

\begin{table}[H]
\footnotesize
\centering
\caption{Background information of participants.}
\label{participants}
\begin{tabular}{c|c|c|c|c}
\toprule[1.5pt]
User       & Gender & Age & Nationality & \makecell[c]{ML\\ Background}      \\ \midrule
1 & Female     & 24          & Chinese          & \ding{51} \\
2 & Female     & 21          & Chinese          & \ding{51} \\
3 & Male     & 25          & Chinese          & \ding{51} \\
4 & Male     & 24          & Chinese          & \ding{51} \\
5 & Male     & 26          & Chinese          & \ding{51} \\
6 & Male     & 24          & Chinese          & \ding{55} \\
7 & Male     & 27          & Chinese          & \ding{51} \\
8 & Male     & 25          & Chinese          & \ding{51} \\
9 & Female     & 24          & Chinese          & \ding{51} \\
10 & Male     & 54          & Chinese          & \ding{55} \\
11 & Female     & 53          & Chinese          & \ding{55} \\
12 & Male     & 33          & Chinese          & \ding{55} \\
13 & Female     & 30          & Chinese          & \ding{55} \\
14 & Male     & 25          & Finnish          & \ding{55} \\
15 & Male     & 26          & Finnish          & \ding{51} \\
16 & Female     & 27          & Finnish          & \ding{51} \\
17 & Male     & 24          & American          & \ding{51} \\
18 & Male     & 23          & Australian          & \ding{51} \\
19 & Male     & 20          & Canadian          & \ding{55} \\
20 & Female     & 30          & British          & \ding{55} \\

\bottomrule[1.5pt]
\end{tabular}
\end{table}

\begin{table}[H]
\footnotesize
\centering
\caption{Statistics of Advogato and PGP datasets.}
\label{additional_dataset}
\begin{tabular}{c|c|c|c|c}
\toprule[1.5pt]
Dataset       & \# Nodes & \# Edges & Avg. Degree & Trust Level      \\ \midrule
\multirow{2}{*}{Advogato} & \multirow{2}{*}{6541} & \multirow{2}{*}{51127} & \multirow{2}{*}{19.2} & \{Observer, Apprentice, \\
                   &                    &          &      &   Journeyer, Master\}   \\
\multirow{2}{*}{PGP} & \multirow{2}{*}{38546} & \multirow{2}{*}{317979} & \multirow{2}{*}{16.5} & \multirow{2}{*}{\{1,2,3,4\}} \\
                   &                    &          &      &  \\
\bottomrule[1.5pt]
\end{tabular}
\end{table}

\section{Supporting Evidence Related to Adversarial Attacks}
\label{appendix_adversarial}
\begin{figure}
	\centering
	\includegraphics[scale=0.35]{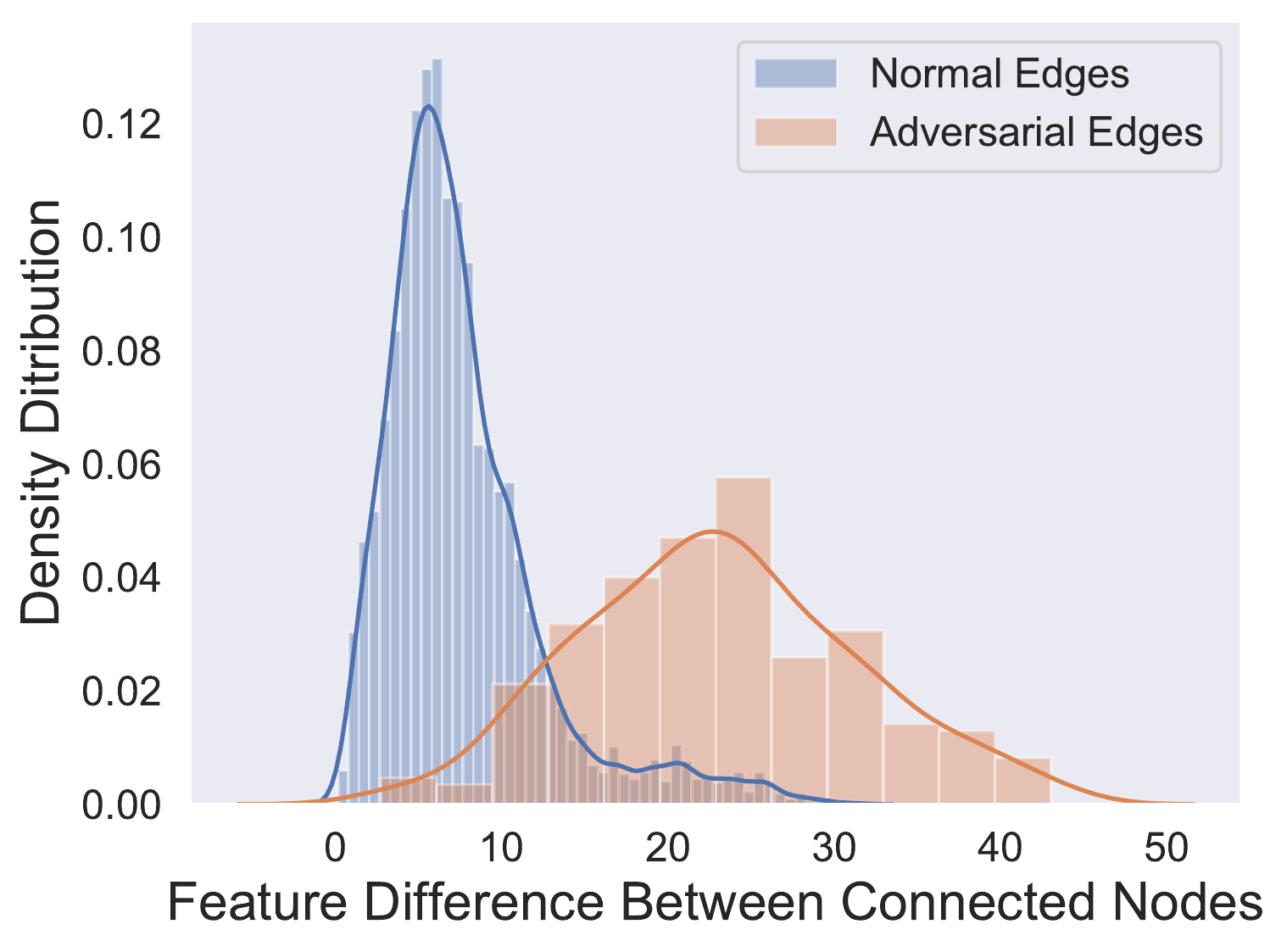}
	\caption{Histograms of the feature difference between connected nodes before and after adversarial attacks.}
	\label{adversarial_attack}
\end{figure}

To support the discussion on adversarial attacks in Section~\ref{section_robustness}, we refer to Fig.~\ref{adversarial_attack}, which is sourced from Jin \textit{et al.}~\cite{jin2020graph}. This figure clearly illustrates that adversarial attacks tend to connect nodes with large feature difference.

\end{document}